%% file: paper.tex
\documentclass{article}
\newcommand{\Venue}{iclr}

\newcommand{\NipsVenueAlias}{nips}
\ifx\Venue\NipsVenueAlias
  \renewcommand{\Venue}{neurips}
\fi
 
\input{venues/iclr/preamble.tex}

\begin{document}
\input{venues/iclr/title.tex}
\input{common/paper.tex}

\input{venues/iclr/bibliography.tex}
\input{venues/iclr/appendix.tex}

\end{document}

%% file: venues/iclr/preamble.tex
\usepackage{venues/iclr/configuration/iclr2027_conference}
\usepackage{times}
\usepackage[T1]{fontenc}
\usepackage[utf8]{inputenc}
\usepackage{url}
\input{common/packages}

\input{common/metadata}

\iclrfinalcopy

%% file: common/packages.tex
\usepackage{graphicx}
\usepackage{booktabs}
\usepackage{longtable}
\usepackage{multirow}
\usepackage{tabularx}
\usepackage{comment}
\usepackage{enumitem}
\usepackage{xcolor}
\usepackage{colortbl}
\usepackage[most]{tcolorbox}
\usepackage{pifont}
\usepackage{microtype}
\usepackage{wrapfig}
\usepackage{inconsolata}
\usepackage{xspace}
\usepackage{common/support/algorithm}
\usepackage{common/support/algorithmicx}
\usepackage[noend]{common/support/algpseudocode}
\usepackage{hyperref}
\input{common/macros/lins_macros}
\input{common/macros/lins_algo}

\graphicspath{{common/}{common/figures/}{./}}

\theoremstyle{definition}

\definecolor{darkgray}{gray}{0.25}
\definecolor{taggreen}{RGB}{0,128,0}
\definecolor{tagmagenta}{RGB}{170,0,120}
\definecolor{tagteal}{RGB}{0,110,130}

\definecolor{readonlyorange}{RGB}{230,85,0}
\definecolor{writableblue}{RGB}{0,100,200}
\newcommand{\ro}[1]{\textcolor{readonlyorange}{\texttt{#1}}}  
\newcommand{\rw}[1]{\textcolor{writableblue}{\texttt{#1}}}  


%% file: common/macros/lins_macros.tex
\usepackage{amsmath,amsthm,amssymb}

\newenvironment{talign*}
{\let\displaystyle\textstyle\csname align*\endcsname}
{\endalign}

%% file: common/macros/lins_algo.tex
\usepackage{algorithm}
\usepackage[noend]{algpseudocode}

\algnewcommand\algorithmicforeach{\textbf{for each}}
\algdef{S}[FOR]{ForEach}[1]{\algorithmicforeach\ #1\ \algorithmicdo}

\errorcontextlines\maxdimen

\makeatletter
\newcommand*{\algrule}[1][\algorithmicindent]{\makebox[#1][l]{\hspace*{.5em}\thealgruleextra\vrule height \thealgruleheight depth \thealgruledepth}}%
\newcommand*{\thealgruleextra}{}
\newcommand*{\thealgruleheight}{.75\baselineskip}
\newcommand*{\thealgruledepth}{.25\baselineskip}

\newcount\ALG@printindent@tempcnta
\def\ALG@printindent{%
	\ifnum \theALG@nested>0
	\ifx\ALG@text\ALG@x@notext
	\else
		\unskip
		\addvspace{-1pt}
		\ALG@printindent@tempcnta=1
		\loop
		\algrule[\csname ALG@ind@\the\ALG@printindent@tempcnta\endcsname]%
		\advance \ALG@printindent@tempcnta 1
		\ifnum \ALG@printindent@tempcnta<\numexpr\theALG@nested+1\relax
			\repeat
		\fi
	\fi
}%
\usepackage{etoolbox}
\patchcmd{\ALG@doentity}{\noindent\hskip\ALG@tlm}{\ALG@printindent}{}{\errmessage{failed to patch}}
\makeatother

\newbox\statebox
\newcommand{\myState}[1]{%
	\setbox\statebox=\vbox{#1}%
	\edef\thealgruleheight{\dimexpr \the\ht\statebox+1pt\relax}%
	\edef\thealgruledepth{\dimexpr \the\dp\statebox+1pt\relax}%
	\ifdim\thealgruleheight<.75\baselineskip
		\def\thealgruleheight{\dimexpr .75\baselineskip+1pt\relax}%
	\fi
	\ifdim\thealgruledepth<.25\baselineskip
		\def\thealgruledepth{\dimexpr .25\baselineskip+1pt\relax}%
	\fi
	\State #1%
	\def\thealgruleheight{\dimexpr .75\baselineskip+1pt\relax}%
	\def\thealgruledepth{\dimexpr .25\baselineskip+1pt\relax}%
}

%% file: common/metadata.tex
\newcommand{\PaperTitle}{MS-Eval: Benchmarking Sub-Agent Delegation and Orchestration in MCP-Intensive Working Scenarios}

\makeatletter
\let\PaperKernelMakefntext\@makefntext
\newcommand{\PaperIclrAuthorBlock}{
    Bowen Ding\textsuperscript{\rm 1,2,*}\,,
    Qixin Wang\textsuperscript{\rm 2,*,\dag}\,,
    Lexing Yuan\textsuperscript{\rm 2,\dag}\,,
    Weiyi Lu\textsuperscript{\rm 2,\dag}\,,
    Qiwei Ge\textsuperscript{\rm 4}\,,
    Ruishan Fang\textsuperscript{\rm 1,2}\,,\\
    \bf Tao Lin\textsuperscript{\rm 2,3,\ddag}
  \\ [1mm]
    \textsuperscript{1} Zhejiang University~
    \textsuperscript{2} School of Engineering, Westlake University \\
    \textsuperscript{3} Institute of Advanced Technology, Westlake Institute for Advanced Study \\
    \textsuperscript{4} Independent Researcher \\
    $^{2}$\texttt{\{dingbowen, lintao\}@westlake.edu.cn}%
    \gdef\@makefntext##1{##1}%
    \gdef\@thanks{%
      \footnotetext{\raggedright
        \textsuperscript{*}~Co-first author.\quad
        \textsuperscript{\dag}~Work done at Westlake University as an intern.\quad
        \textsuperscript{\ddag}~Corresponding authors.}%
      \global\let\@makefntext\PaperKernelMakefntext
    }%
    \\
}
\makeatother


%% file: venues/iclr/title.tex
\title{\PaperTitle}
\author{\PaperIclrAuthorBlock}
\maketitle

%% file: common/paper.tex

\begin{abstract}
    LLM agents are increasingly integrated into real-world works, yet they must handle diverse domain-specific tools and complex multi-sub-task problems that sequential tool invocation resolves inefficiently and with severe context bloat. The \textit{Main-Agent Delegation and Sub-Agent Execution} (MAD-SAE) paradigm addresses these issues by delegating sub-tasks to specialized sub-agents for parallel execution in isolated contexts, but its delegation and orchestration quality remains unquantified. We introduce MS-Eval, a unified evaluation kit comprising an MCP-intensive benchmark, a controlled execution harness, tailored delegation and orchestration metrics, and the MSQ index for comprehensive assessment. Evaluating four MAD-SAE variants and eight frontier models reveals that {(1)  sub-agent specialization (Spec) is the best practical practice}; {(2) open-source models are largely competitive with few exceptions}; and (3) MAD-SAE generally raises cost with contingent payoff, benefiting most open-source models in both performance and efficiency while leaving closed-source models with higher cost and degraded performance. Our case study further traces these gains to better single-agent solutions, effective context offloading, and genuine sub-task parallelism, whereas unnecessary or redundant delegation turns the same cost into pure overhead. Codes:~\url{https://github.com/LINs-lab/MS-Eval}
    \end{abstract}

\section{Introduction}
\input{common/figures_tex/msq_rank.tex}
\vspace{-0.2cm}
Driven by the scaling of parameters and data~\citep{kaplan2020scalinglawsneurallanguage, retu}, LLMs have evolved into tool-empowered autonomous agents~\citep{glm5,ds_v41_flash} that are increasingly integrated into daily works to augment human productivity~\citep{xu2026theagentcompany}.

However, acting as a useful working assistant poses immense challenges. Real-world working tasks demand diverse and domain-specific tools inherently. For instance, researchers necessitates scholarly retrieval tools for literature reviews, educational administrators relies on Canvas for course management~\citep{toolathlon}. Additionally, problems in working scenarios are often complicated, involving multiple sub-tasks~\citep{toolathlon-gym}, as the example illustrated in Figure~\ref{fig:main_picture}(a). Traditionally, agents resolve these sub-tasks by rigidly invoking tools in a sequential manner~\citep{mcp_atlas,toolathlon}. This lack of parallelization severely limits task-solving efficiency~\citep{kimiteam2026kimik25visualagentic}. More critically, exploratory sub-tasks often retrieve information-intensive chunks that quickly bloat the context window, leading to severe decision distortion in subsequent steps~\citep{lost_in_the_middle,mitigating_context_interference}.

To address these challenges, mainstream harness frameworks (e.g., Claude Code~\citep{claude-code}, Codex~\citep{codex}, Cursor~\citep{cursor}) integrate Model Context Protocol (MCP) servers~\citep{anthropic2024mcp} to fulfill specialized tool demands, and support the autonomous multi-agents runtime paradigm, {Main-Agent Delegation and Sub-Agent Execution } (MAD-SAE, details in \S~\ref{sec:mad_sae}). In the paradigm, the main-agent dynamically delegates sub-tasks to specialized sub-agents for parallel execution in isolated contexts, ultimately returning concise summaries. Consequently, it is expected to alleviate the context burden of single-agent, and accelerate task-solving via parallel sub-agent execution. While widely adopted, the sub-agent delegation and orachestration quality of MAD-SAE has yet to be quantified. This leaves a critical open problem: 
\begin{tcolorbox}[
    colback=red!5!white,
    colframe=red!45!black,
    boxrule=0.6pt,
    arc=2mm,
    left=1mm,
    right=1mm,
    top=0.8mm,
    bottom=0.8mm,
    before skip=5pt,
    after skip=5pt
]
    \centering
    \emph{Can frontier LLMs in MAD-SAE {optimally route sub-tasks at the right time} for {highly efficient and cost-effective} task completion?}
\end{tcolorbox}

To facilitate this evaluation, we propose the \textbf{M}AD-\textbf{S}AE \textbf{Eval}uation (MS-Eval) kit,  comprising three components: (1) a curated MCP-intensive  benchmark covering eight working scenarios, featuring tasks annotated with optimal delegation priors (\S~\ref{sec:benchmark}); (2) A controlled harness standardizing the execution principle and tool accessibility for MAD-SAE and the Single-Agent (SA) baseline (\S~\ref{sec:harness}); (3) tailored metrics evaluating MAD-SAE on Delegation Appropriateness (e.g., ODR, UDR, and Sub-Agent Hit Scores) and Orchestration Utilities (performance, efficiency, and cost gains over SA). Additionally, we propose the MAD-SAE Quality (MSQ) index, a unified metric integrating both perspectives for a comprehensive assessment (\S~\ref{sec:eval}).

After evaluating on four implementation variants of MAD-SAE (e.g.,  Gen, Spec, Plan-1st and Oracle, details in \S~\ref{sec:mad-sae-variants}), and eight frontier models under the best MAD-SAE practice, we reveal several critical insights:
\vspace{-0.2cm}
\begin{itemize}[leftmargin=12pt]
    \item \textbf{Sub-agent specialization (Spec) is the best MAD-SAE practice.} Spec avoids the overly conservative delegation of Gen, and the aggressive over-delegation of Plan-1st, achieving the highest MSQ with 20.8\% among practical variants in Table~\ref{tab:variants_models_comparison}(a).
    \vspace{-0.1cm}
    \item \textbf{Open-source models lead delegation and orchestration quality.} Excluding Oracle, the top six MSQ settings (Fig.~\ref{fig:msq_rank}) are all open-source models, with GLM-5.3 (Spec) attaining the highest MSQ of 24.6\% and Kimi-K3 (Spec) the only setting improving performance, efficiency, and cost simultaneously as shown in Table~\ref{tab:variants_models_comparison}(b).
    \vspace{-0.1cm}
    \item \textbf{MAD-SAE generally incurs higher cost, but its payoff is contingent on delegation quality.} Across the full benchmark $\mathcal{D}$ (\S~\ref{sec:mad-sae-analysis}), MAD-SAE improves both performance and efficiency for most open-source models at added cost, whereas closed-source models see cost rise alongside degraded performance and efficiency. Our case study traces these gains to more efficient single-agent solutions, effective context offloading, and genuine sub-task parallelism, while unnecessary, unverified, or redundant delegation turns the same cost into pure overhead.    
\end{itemize}

\section{Related Work}
\label{sec:related_work}

\noindent \textbf{MCP-Intensive Evaluation.} To empower LLMs as capable working assistants, modern harness frameworks~\citep{codex, claude-code,cursor} increasingly integrate Model Context Protocol (MCP) servers~\citep{anthropic2024mcp}, motivating the growing demand of MCP-intensive evaluation. Existing MCP evaluations have progressively evolved along three directions. Early efforts established large-scale evaluation protocols focusing on tool retrieval, execution correctness, and automated judging mechanisms \citep{mcpeval, livemcpbench, mcp_radar}. To improve reliability and reduce evaluation subjectivity, subsequent works introduced deterministic execution-based verification and broader real-world MCP environments \citep{mcpuniverse, livemcp101, mcpagentbench}, with MCPWorld further incorporating semantic internal-state verification through application-level instrumentation \citep{mcpworld}. Recent benchmarks have pushed toward more realistic working scenarios, emphasizing long-horizon CRUD workflows by MCPMark~\citep{mcpmark}, cross-application orchestration by MCP-Atlas~\citep{mcp_atlas}, and ambiguous user instructions across diverse applications by Toolathlon~\citep{toolathlon}. Toolathlon-Gym~\citep{toolathlon-gym} inherits Toolathlon's broad scenario coverage and provides a locally deployed MCP server and database environment, on which we build MS-Eval.

\noindent \textbf{Multi-Agent Orchestration.} 
Multi-Agent Systems (MAS) are designed to overcome the limitations of single-agent systems in context capacity~\citep{when_do_mas_help}, insufficient exploration~\citep{dora_explorer}, and inefficient serial execution~\citep{masterman2024landscapeemergingaiagent}, through specialized collaboration~\citep{llm_based_multi_agents}. Early multi-agent studies explored structured collaboration via role specialization, communication protocols, and iterative interaction among agents~\citep{camel,du2023improvingfactualityreasoninglanguage,societyofmind}. Recent efforts have shifted toward autonomous orchestration, enabling LLM agents to dynamically plan workflows, assign responsibilities, and coordinate heterogeneous agents or tools through harness frameworks~\citep{claude-code,codex,cursor}. More recent systems further scale this paradigm toward adaptive, large-scale delegation, e.g., agent swarms of~\citet{kimiteam2026kimik25visualagentic}. Despite this progress, existing MAS evaluations remain limited: MultiAgentBench~\citep{multiagentbench} assesses task completion, delegation, and coordination quality under plan-first runtime constraints, but does not compare efficiency or cost against single-agent execution; \citet{fu2026agentshelpcontrolledprotocolaligned} conduct a controlled comparison between single-agent and MAS in terms of accuracy and cost, yet focus on short-horizon reasoning, coding, and instruction following, lacking standardized evaluation of delegation quality; ClawArena-Team~\citep{clawarenateam} considers sub-agent orchestration but primarily studies a main agent creating its own sub-agent team (e.g., tool granting), without assessing the rationality of sub-agent delegation or collaborative gains. Moreover, none of these works consider MCP-intensive working scenarios.

To our knowledge, we are the first to systematically evaluate sub-agent delegation quality and orchestration versus single-agent execution in performance, efficiency, and cost under MCP-intensive working scenarios.

\section{Main-Agent Delegation and Sub-Agent Execution}
\label{sec:mad_sae}

\subsection{Definition and Property}
\label{sec:def_and_property}
Main-Agent Delegation and Sub-Agent Execution (MAD-SAE) serves as an autonomous muti-agent orchestration paradigm, being supported by mainstream harness frameworks, such as Claude Code~\citep{claude-code}, Codex~\citep{codex}, Cursor~\citep{cursor}, etc. The paradigm transforms the classical Single-Agent (SA) task-solving runtime into a collaborative, hierarchical orchestration, decoupling long-horizon contexts via two distinct roles:

\noindent \textbf{Main-Agent.} Governing global execution, the main-agent is uniquely empowered with \textit{Agent-Delegation (AD)} tools (i.e., \texttt{Agent, AgentSwarm} in Fig.~\ref{fig:main_picture}). With thses tool, the main-agent dynamically decomposes tasks into sub-tasks and delegates them to specialized sub-agents. Each concurrent execution of multiple specialized sub-agents is called a \textbf{\textit{wave}}. Depending on the delegation pattern, a runtime session may feature zero, one, or multiple sequential waves.

\noindent \textbf{Sub-Agent.} Triggered within a wave, a sub-agent is a specialized AI assistant responsible solely for its assigned sub-task. Each sub-agent has its own type, operates within an isolated context window, is constrained by a specialized prompt and a tightly restricted tool allowlist, and ultimately returns its results to the main-agent.

Intuitively, MAD-SAE is expected to have two advantages: \textbf{(1) Higher efficiency}: parallelizable sub-agents enable concurrent sub-task execution, potentially surpassing the serial SA baseline~\citep{kimiteam2026kimik25visualagentic}. \textbf{(2) Performance gains}: specialized sub-agents may solve their expert sub-tasks better and, via isolated contexts, offload intensive information (e.g., exploration findings) from the main-agent, preventing context bloat from degrading decisions~\citep{when_do_mas_help,cursor_subagents}.

\subsection{Variants}
\label{sec:mad-sae-variants}
We consider three practical MAD-SAE variants, General (Gen), Specialized (Spec), and Plan-1st (Plan), plus an Oracle baseline for golden study (see Appx.~\ref{appx:mad-sae-variants-illustration} for details). \textbf{Gen} lets the main-agent autonomously decide when and how to delegate to generally defined sub-agents: a read-only Planner for decomposition and delegation, a read-only Explorer for exploration, and a writable Coder for editing workflows. \textbf{Spec} retains Gen's Planner but replaces the general sub-agents with domain-specialized ones, splitting exploration into four read-only experts (Academic-Literature-Researcher, Enterprise-Data-Analyst, Financial-Market-Analyst, and Web-Domain-Researcher) and editing into two writable experts (External-Workflow-Operator and Office-Report-Builder). \textbf{Plan-1st} inherits Gen's typology but mandates calling the Planner for decomposition and delegation planning before execution. \textbf{Oracle} injects golden delegation priors into Gen's prompt to test orchestration utilities under golden delegation.

\section{MS-Eval Kit}
\label{sec:ms-eval-kit}
To comprehensively investigate the sub-agent delegation appropriateness of main-agent and the underlying orchestration utilities against the SA baseline, we introduce the \textbf{M}AD-\textbf{S}AE \textbf{Eval}uation Kit ({MS-Eval Kit}). This kit comprises three core components: a curated benchmark (\S~\ref{sec:benchmark}), a controlled harness (\S~\ref{sec:harness}), and tailored evaluation metrics (\S~\ref{sec:eval}).

\input{common/figures_tex/main_picture.tex}

\subsection{Benchmark}
\label{sec:benchmark}

We sample, refine\footnote{We cover all Toolathlon-Gym scenarios and align task descriptions with evaluation criteria to reduce false negatives.}, and annotate 166 tasks from Toolathlon-Gym~\citep{toolathlon-gym}, a large-scale and long-horizon environment covering eight real-world work scenarios: Learning Management System (LMS), Enterprise Data Warehouse (EDW), E-Commerce (EC), Stock Market (SM), Video Platform (VP), Railway System (RS), Academic Research (AR) and  Food-and-Catering (FC).

To facilitate the assessment on delegation appropriateness, we establish  Delegation Prior, a golden delegation plan, as the ground truth for each task. As an example in Fig.~\ref{fig:main_picture}, such prior specifies (1) \textit{when} to delegate a wave of sub-agents, (2) \textit{which types} of sub-agents to assign, and (3) \textit{which sub-tasks} they should handle to complete the task efficiently. The annotations are derived from each task's description, required tools, and expected deliverables, following two principles: (i) maximizing opportunities for parallel execution, and (ii) isolating exploratory sub-tasks within sub-agents to reduce context overload for main-agent. These principles echo expected advantages mentioned in \S~\ref{sec:mad_sae}. 

Based on the annotation, the entire dataset is divided into two splits: (1) $\mathcal{D}_{\text{no-need}}$ with 77 tasks, where the main-agent alone is already efficient enough, and (2) $\mathcal{D}_{\text{need}}$ with 89 tasks, where the main-agent should delegate sub-tasks to sub-agents for efficient and effective solutions. Refer to Appx.~\ref{appx:descriptive_statistics} for detailed sub-agent delegation statistics.

\subsection{Harness}
\label{sec:harness}
We develop MAD-SAE and single-agent (SA), two agent executation paradigms, on top of Kimi-Code~\citep{moonshot_kimi_code_agent_tools}. Kimi-Code is an open-source framework with native sub-agent support. To rigorously isolate the effects of execution paradigms under specified working scenarios, we disable Kimi-Code's built-in skills and strictly manage the executation principle and tool accessibility. Refer to Appx.~\ref{appx:harness} for detials.

\noindent \textbf{Executation Principle.} Agent behaviors are standardized through tailored principles in system prompts.
For SA, the basic principle is enough to govern its behavior, by enforcing efficient task execution, strict workspace boundary restrictions, and a mandatory completion protocol. For MAD-SAE, the main-agent inherits the basic principle of SA, and is further augmented with sub-agent delegation norms, available sub-agent profiles, and concrete few-shot delegation examples to guide appropriate delegation. Conversely, sub-agents operate under specialized constraints: they are strictly confined to their isolated workspaces, mandated to return concise result summaries to the main-agent, and explicitly prohibited from invoking the global \texttt{claim\_done} termination signal.

\noindent \textbf{Tool accessibility.} SA is granted with basic tools, which includes \texttt{TodoList} for progress tracking, MCP tools for specified task-solving, \texttt{claim\_done} for task termination and \texttt{python\_execute} for isolated python coding. Within MAD-SAE, the main-agent additionally accesses \texttt{Agent} and \texttt{AgentSwarm} besides basic tools, to delegate sub-agents for heterogeneous or homogeneous sub-tasks, while sub-agents receive role-specific tool allowlists and runtime exclusions.

\subsection{Evaluation Metrics}
\label{sec:eval}
Our evaluation considers two complementary perspectives: {Delegation Appropriateness (DA)} (\S~\ref{sec:da}) to assess the sub-agent delegation rationality, and {Orchestration Utilities (OU)} (\S~\ref{sec:ou}) to quantify performance, efficiency, and cost gains over the SA baseline. \S~\ref{sec:msq} aggregates these into {MAD-SAE Quality (MSQ)}, a unified index discounting the average utility by delegation appropriateness.

\subsubsection{Delegation Appropriateness (DA)} 
\label{sec:da}
Two questions are considered for evaluating DA: (1) \emph{whether} the main agent delegates when sub-agent assistance is warranted, and, (2) once it delegates, \emph{how well} the executed delegations match the expected ones. Accordingly, we introduce: {Under-Delegation Rate (UDR)}, {Over-Delegation Rate (ODR)}, and Sub-agent Hit Scores with $P_{\text{hit}}$, $R_{\text{hit}}$ and $\mathcal{F}_{\text{1 hit}}$.

\noindent \textbf{UDR and ODR} assess whether the main agent correctly determines when to delegate. Let $\mathcal{D}_{\mathrm{need}}$ and $\mathcal{D}_{\mathrm{no\text{-}need}}$ denote the sets of tasks annotated as requiring and not requiring sub-agent assistance, respectively, and let $\mathcal{D}_{\mathrm{del}}$ denote the tasks in which the main agent actually invokes at least one non-planning sub-agent. We define 
\begin{equation} 
    \small
    \label{eq:udr-odr}
    \mathrm{UDR} = \frac{ |\mathcal{D}_{\mathrm{need}} \setminus \mathcal{D}_{\mathrm{del}}| }{ |\mathcal{D}_{\mathrm{need}}| }, \qquad \mathrm{ODR} = \frac{ |\mathcal{D}_{\mathrm{no\text{-}need}} \cap \mathcal{D}_{\mathrm{del}}| }{ |\mathcal{D}_{\mathrm{no\text{-}need}}| }. 
\end{equation}
UDR measures missed delegation opportunities, whereas ODR measures unnecessary delegation.

\noindent \textbf{Sub-agent Hit Scores} evaluate fine-grained delegation quality by matching runtime-delegated sub-agents with annotated golden sub-agents. To match sub-agents defined under different conventions (e.g., \texttt{Coder} of Gen v.s. \texttt{Office-Report-Builder} of Spec), we functionally tag both golden and delegated sub-agents based on their sub-task semantics, using 24 labels (details in Appx.~\ref{appx:function_labeling_for_sub-agent_hit}) and DeepSeek-V4.1-Flash~\citep{ds_v41_flash}. Within a wave, tagged sub-agents are grouped into function units that never span waves, with all sub-agents solving the same sub-task collapsing into one unit.  Typically, each writable sub-agent forms its own unit, whereas read-only exploratory sub-agents form separate units only when handling different sub-tasks.

Treating function units as atomic delegation responsibilities, for task $t$ we obtain the golden unit group $G_t$ and delegated unit group $E_t$. Using maximum bipartite matching, we find $m_t$ one-to-one matches ($m_t \leq \min(|G_t|, |E_t|)$), determining which golden units are hit. We then compute:
\begin{equation}
    \small
    \label{eq:sub-agent-hit}
    \mathcal{P}_{\text{hit}}
    =
    \frac{\sum_t m_t}{\sum_t |E_t|},
    \qquad
    \mathcal{R}_{\text{hit}}
    =
    \frac{\sum_t m_t}{\sum_t |G_t|},
    \qquad
    \mathcal{F}_{\text{1 hit}}
    =
    \frac{2\mathcal{P}_{\text{hit}}\mathcal{R}_{\text{hit}}}
         {\mathcal{P}_{\text{hit}}+\mathcal{R}_{\text{hit}}}.
\end{equation}
where $\mathcal{P}_{\text{hit}}$ measures the precision of executed delegation units, $\mathcal{R}_{\text{hit}}$ measures the coverage of golden units, and $\mathcal{F}_{\text{1 hit}}$ balances the two.

\subsubsection{Orchestration Utilities (OU)}
\label{sec:ou}

We quantify MAD-SAE utilities over the SA baseline across three dimensions: \textbf{performance} ($p$), \textbf{efficiency} ($e$), and \textbf{cost} ($c$). Specifically, performance is measured by Success Rates (SR) in Eq.~\ref{eq:sr} on $\mathcal{D}_{\mathrm{del}}$ (defined in \S~\ref{sec:da}). Conversely, efficiency and cost are strictly evaluated via Critical Steps (Csteps) in Eq.~\ref{eq:csteps} and API billing estimated in Eq.~\ref{eq:api_billing}, on the successfully completed subset $\mathcal{D}_{\mathrm{del}}^{+} \subseteq \mathcal{D}_{\mathrm{del}}$, as resource savings are meaningful only for solved tasks. 

For each dimension $i \in \{p,e,c\}$, let $v_i^{\mathrm{mad}}$ and $v_i^{\mathrm{sa}}$ denote the measurement for MAD-SAE and SA of the specified dimension $i$. Because these metrics possess disparate scales and opposing optimization directions (i.e., higher-is-better for performance, but lower-is-better for efficiency and cost), we normalize them into a unified, higher-is-better $[0,1]$ index $U_{i}$. This standardization seamlessly facilitates the MAD-SAE quality (MSQ) aggregation in \S~\ref{sec:msq}:
\begin{equation}
    \small
    \label{eq:orchestration-utility}
    U_i =
    \begin{cases}
        \displaystyle
        \frac{v_i^{\mathrm{mad}}}
             {v_i^{\mathrm{mad}} + v_i^{\mathrm{sa}}},
        & i=p, \\[8pt]
        \displaystyle
        \frac{v_i^{\mathrm{sa}}}
             {v_i^{\mathrm{mad}} + v_i^{\mathrm{sa}}},
        & i\in\{e,c\}.
    \end{cases}
\end{equation}
Under this formulation, $U_i=50\%$ indicates parity between the two paradigms, $U_i>50\%$ reflects a net advantage for MAD-SAE, and $U_i<50\%$ reveals a negative orchestration utility.

\subsubsection{MAD-SAE Quality}
\label{sec:msq}
Building upon $\mathcal{F}_{\text{1 hit}}$ of Sub-Agent Hit Scores in \S~\ref{sec:da} and normalized utilities $U_{i}, i \in \{p,e,c\}$ in \S~\ref{sec:ou}, we aggregate these metrics for a overall assessment of the MAD-SAE quality (MSQ):
\begin{equation}
    \small
    \label{eq:msq}
    \mathrm{MSQ}
    =
    \mathcal{F}_{\text{1 hit}} \cdot
    \left(
        w_p U_p + w_e U_e + w_c U_c
    \right),
    \qquad
    w_p+w_e+w_c=1.
\end{equation}
Here, $w_p$, $w_e$, and $w_c$ dictate the relative emphasis on performance, efficiency, and cost. While these weights can be flexibly tailored to specific priorities, we assign equal weights ($1/3$) in our experiments to ensure a balanced evaluation. Crucially, $\mathcal{F}_{\text{1 hit}}$ acts as a multiplicative discount: high orchestration utility receives full credit only when driven by appropriate delegation.

\section{Experiment Setup}
\label{sec:experiment_setup}
Leveraging the MS-Eval Kit mentioned in \S~\ref{sec:ms-eval-kit}, our experiments are structured to progressively address three core research questions:
(1) \textbf{RQ1:} Among representative MAD-SAE variants, which performs best, and what is its gap to the oracle upper bound?
(2) \textbf{RQ2:} Under the optimal implementation, which frontier model achieves the highest MSQ, and how do their orchestration profiles differ?
(3) \textbf{RQ3:} Ultimately, does MAD-SAE actualize its expected utilities over the monolithic Single-Agent (SA) paradigm, and what is the inherent trade-off?

\subsection{Models and Configuration}
\label{sec:implement_details}

\noindent \textbf{Models.}
Multiple closed-source and open-source models are considered into our evaluation. The closed-source ones include GPT-5.6-Sol~\citep{openai2026gpt56} and Claude-Sonnet-5~\citep{anthropic2026sonnet5}, while the open-source ones cover models of varying sizes: Qwen3.8-Max~\citep{qwen38}, Kimi-K3~\citep{k3}, GLM-5.3~\citep{zai2026glm53}, MiniMax-M3~\citep{minimax2026m3}, GLM-5.3-Flash~\citep{zai2026glm53-falsh}, and Deepseek-V4-Flash-0731~\citep{deepseekai2026deepseekv4}. Considering the high cost-effectiveness of Deepseek-V4-Flash-0731, we use it for RQ1 experiments. See Appx.~\ref{appx:eval_config} for detialed model inference configurations.

\noindent \textbf{Runtime.} All experiments proceed as fully autonomous, single-turn, multi-step runs with zero human intervention.
To accommodate the complex, long-horizon episode, we provision every main-agent and sub-agent with a 128K context window, a 100-step interaction limit, and a 7,200-second running-time budget. Following default Kimi-Code settings, \texttt{AgentSwarm} is capped at 128 concurrent sub-agents. To prevent unintended system exploitation, built-in terminal tools are globally disabled, and the terminal access is granted strictly per task and safely supported via a local mock terminal MCP server. During execution, a session terminates normally if the agent invokes \texttt{claim\_done} or yields a pure-text response without any tool calls. Besides it, the execution is also terminated upon breaching the step/time budgets, entering a loop of six consecutive repetitive tool calls, or exceeding the maximum API retry limit of ten.

\input{common/tables/all_res_table.tex}
\section{Main Results}
\label{sec:main_results}

\noindent \textbf{MAS-SAE Variants Comparison.}
As Table~\ref{tab:variants_models_comparison} (a), Spec achieves the highest $\mathcal{F}_{\text{1 hit}}$ and average utilities among three practical variants, yielding a peak MSQ of 20.8\%. The result outperform Gen by 9.2 points, the counterpart with general sub-agent profiles. Gen's underperformance stems from overly conservative delegation: despite a leading $\mathcal{P_{\text{hit}}}$ (48.0\%), its low $\mathcal{R_{\text{hit}}}$ (18.8\%) and high UDR (52.8\%) leave over half of $\mathcal{D}_{\mathrm{need}}$ tasks without MAD-SAE benefits. Conversely, Plan-1st suffers from aggressive over-delegation, erroneously invoking sub-agents in 61.0\% of $\mathcal{D}_{\mathrm{no\text{-}need}}$ tasks. This fails to translate into gains over SA, with all utilities ($U_p, U_e, U_c$) falling below the 50\% parity line. Ultimately, sub-agent specialization variant of MAD-SAE, \textbf{Spec, emerges as the best practice}, navigating between the extremes of conservative and aggressive delegation.

Guided by delegation priors, Oracle provides near-perfect delegations with 92.5\% of $\mathcal{F}_{\text{1 hit}}$, which exceeds Gen (the prior-free baseline) by 65.4 points. On the OU metrics, Oracle attains 49.6\% of $U_p$ and 53.5\% of $U_e$, matching the SA baseline in performance while realizing the expected higher efficiency discussed in \S~\ref{sec:def_and_property}. Meanwhile, its $U_c$ of 23.6\%, well below 50\%, indicates substantial additional cost relative to SA, leaving Oracle with an MSQ of only 39.1\%. Under near-perfect sub-agent delegation, Oracle isolates the trade-off of MAD-SAE: extra cost buys efficiency, but not necessarily performance gains.

\noindent \textbf{Models Comparison.}
Confirming Spec's superiority, Table~\ref{tab:variants_models_comparison}(b) summarizes the Spec-based MAD-SAE results across eight frontier models. 

GPT-5.6-Sol exhibits the most aggressive delegation behavior, with a massive ODR of 93.5\%. Crucially, such over-delegation fails to translate into orchestration advantages, with all its utilities falling below the 50\% parity line. Claude-Sonnet-5 similarly suffers from negative orchestration returns across the board ($U_p$: 45.9\%, $U_e$: 46.8\%, $U_c$: 34.1\%). Consequently, both models languish near the bottom of the MSQ leaderboard in Fig~\ref{fig:msq_rank}, with GPT-5.6-Sol (15.6\%) and Claude-Sonnet-5 (13.2\%) ranking third-to-last and second-to-last, respectively.

GLM-5.3 tops the MSQ leaderboard at 24.6\% through active yet rational delegation: its UDR of 8.9\% covers 91.1\% of $\mathcal{D}_{\mathrm{need}}$ tasks, and its $\mathcal{F}_{\text{1 hit}}$ of 51.1\% is the highest overall. Meanwhile, it preserves task performance while ranking second in efficiency, with a $U_e$ of 52.7\%. GLM-5.3-Flash achieves the highest efficiency utility, reaching a $U_e$ of 54.0\%. Kimi-K3 is the only model to surpass SA on performance, efficiency, and cost alike, solving 11 more tasks with the highest $U_p$ of 56.4\%. Yet its conservative delegation, with an $\mathcal{F}_{\text{1 hit}}$ of only 34.1\%, drags its MSQ down to 17.9\% and places it fourth. verall, \textbf{open-source models dominate MSQ}, with more appropriate sub-agent delegation and thus greater orchestration gains in performance and efficiency.


\input{common/figures_tex/performance_cost_efficiency.tex}
\vspace{-0.5cm}
\section{Analysis}
\label{sec:mad-sae-analysis} 
\vspace{-0.2cm}
The OU analysis in \S~\ref{sec:main_results} was confined to $\mathcal{D}_{\text{del}}$, the subset with sub-agent invocation, to isolate delegation-specific benefits. We now turn to the full benchmark $\mathcal{D}$ for a global comparison between MAD-SAE and SA, examining macroscopic performance–efficiency/cost trade-offs and the distinct behaviors each execution principle induces.

\noindent \textbf{Trade-off Analysis.}
Across nearly all models, MAD-SAE carries a cost premium over SA (Fig.~\ref{fig:trade-off}, left), ranging from +\$0.1 for GLM-5.3-Flash to +\$4.2 for GPT-5.6-Sol per task. For open-source models under the Spec variant, this premium typically buys a performance gain: Qwen-3.8-Max, for instance, trades a \$1.0 cost increase for a 10.2-point rise in Success Rates (SR). K3 (Spec) is the sole configuration to escape this trade-off entirely, cutting cost by \$0.4 per task while still lifting SR by 10.2 points. By contrast, closed-source models incur a double penalty: inflated cost alongside degraded performance, echoing the negative utilities reported in Table~\ref{tab:variants_models_comparison}. The same pattern recurs for efficiency (Fig.~\ref{fig:trade-off}, right): most open-source models (K3, Qwen-3.8-Max, GLM-5.3, GLM-5.3-Flash) attain both higher SR and fewer steps under MAD-SAE, whereas closed-source models cluster in the lower-right quadrant, suffering both lower success and more steps. Taken together, these results point to a clear practical takeaway: {For most open-source models, MAD-SAE with Spec is the preferable configuration, delivering gains in both performance and efficiency with more cost. For closed-source models, in contrast, it tends to raise cost while degrading both performance and efficiency, making SA the more practical choice}.

\input{common/figures_tex/case_study_2.tex}
\noindent \textbf{Case Study.}
The trade-off analysis shows that MAD-SAE does not always realize its advantages, depending on the setting. Pairing each MAD-SAE (Spec) result with its SA counterpart across eight models yields 625 \textit{beneficial} and 703 \textit{adverse} cases, revealing MAD-SAE as a double-edged sword whose net payoff hinges on orchestration quality. The full taxonomy and underlying reasons are detailed in Appx.~\ref{appx:cases_taxonomy} and~\ref{appx:underlying_reasons}. Detailed model-wise case analysis refer to Appx.~\ref{appx:further_case_study}.

Strikingly, \textbf{Main-Agent-Only (MAO)} (see Fig.~\ref{fig:better_mao_mm3}) behavior, where the main-agent solves the task unassisted, dominates the beneficial partition. It accounts for 305 cases (48.8\% of all benefits). MAO also appears in 254 adverse cases, but its beneficial occurrences still outnumber its adverse ones, suggesting that models prompted under MAD-SAE occasionally overfit to its semantic patterns yet serendipitously converge on more efficient single-agent solutions.

Beyond MAO, \textbf{Context Offloading} (see Fig.~\ref{fig:context_offload_glm5_3}) and \textbf{Delegated Parallelism} (see Fig.~\ref{fig:delegated_parallelism_glm5_3}) are the primary benefit drivers. Context Offloading accounts for 148 Cheaper-or-Efficient-Success cases by letting isolated sub-agent contexts absorb heavy work so the main-agent can focus on core solving, while Delegated Parallelism drives 88 Cheaper-or-Efficient-Success and 41 Failure-to-Success cases by converting an infeasible serial workload within a bounded context into a feasible parallel one. Conversely, ill-posed delegation backfires: \textbf{Unnecessary Delegation} (see Fig.~\ref{fig:unnessary_delegation_dsv4}) inflates overhead in 124 cases despite preserving success, \textbf{Delegated Failure Propagation} (see Fig.~\ref{fig:delegated_failure_propa_gpt_5_6}) accounts for over half of the Success-to-Failure cases, and \textbf{Execution Bloat} (see Fig.~\ref{fig:execution_bloat_gpt_5_6}) dominates the Costly-and-Inefficient-Success cases. 

These findings provide two takeaways: (1) MAD-SAE can unexpectedly casue  a better single-agent solution compared to the SA execution paradigm;
(2) Context offloading and parallelism of sub-agents convert extra cost into efficiency and performance gains, whereas unnecessary, unverified or redundant delegation turns the same cost into pure overhead.

\section{Conclusion}
We present MS-Eval to quantify MAD-SAE via a benchmark, controlled harness, delegation and orchestration metrics. Across four variants and eight frontier models, we find (1) Spec is the best practice; (2) open-source models are largely competitive, with few exceptions; and (3) MAD-SAE's higher cost yields contingent payoff, benefiting most open-source models in both performance and efficiency while leaving closed-source models with higher cost and degraded performance, with these gains traced to better single-agent solutions, context offloading, and sub-task parallelism rather than delegation for its own sake. Thus, MS-Eval assesses the sub-agent delegation and orchestration comprehensively, revealing the benefits and drawbacks of MAD-SAE in MCP-intensive working scenarios.


%% file: common/figures_tex/msq_rank.tex
\vspace{-0.2cm}
\begin{figure}[!htbp]
    \centering
    \includegraphics[width=0.9\textwidth]{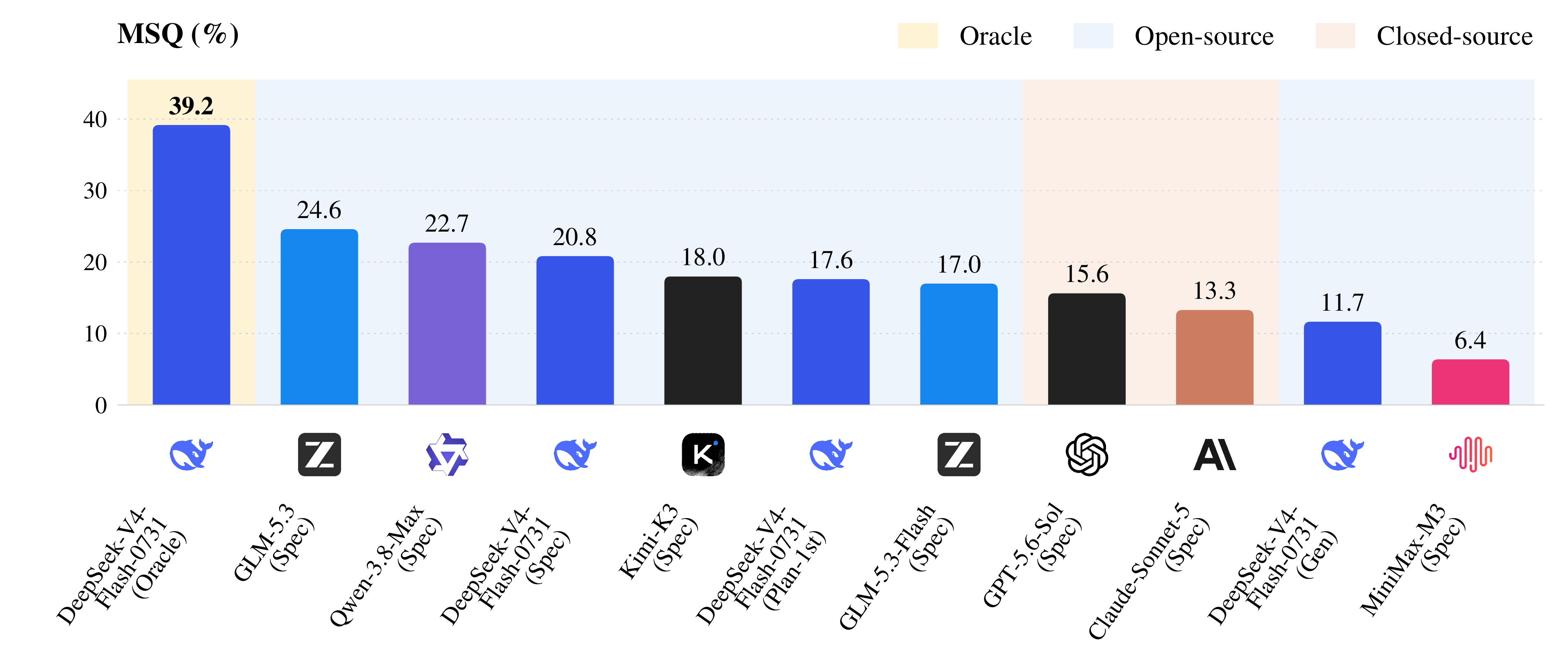}
    \vspace{-0.5cm}
    \caption{{The leaderboard of the MSQ Index}. The index measures the comprehensive quality of \textit{Main-Agent Delegation and Sub-Agent Execution} paradigm (detailed in \S~\ref{sec:eval}), which is dominated by open-source models. Oracle, highlighted in yellow, probes the upper bound of orchestration utilities.}
    \label{fig:msq_rank}
\end{figure}

%% file: common/figures_tex/main_picture.tex
\begin{figure}[!htbp]
    \centering
    \includegraphics[width=1.0\textwidth]{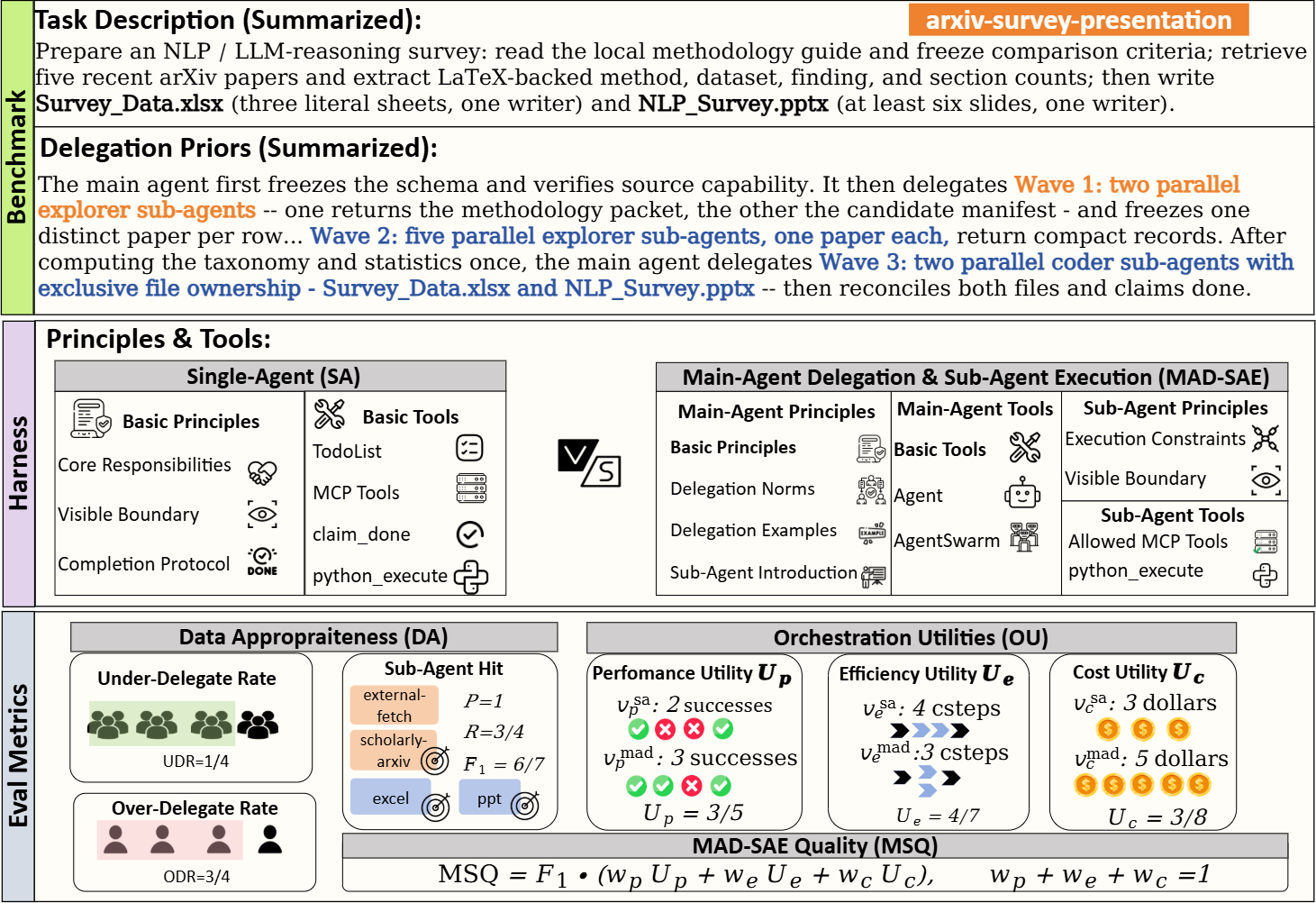}
    \vspace{-0.5cm}
    \caption{{The Overview of MS-Eval Kit.} \textbf{Benchmark} exemplifies \texttt{arxiv-survey-presentation} with its Task Description and Delegation Priors with 3 waves and 9 sub-agents annotation (\S~\ref{sec:benchmark}). \textbf{Harness} defines principles and tool permissions governing agent behavior across execution paradigms (\S~\ref{sec:harness}). \textbf{Eval Metrics} visualizes MAD-SAE metric computation (\S~\ref{sec:eval}).}
    \label{fig:main_picture}
\end{figure}


%% file: common/tables/all_res_table.tex
\vspace{-0.5cm}
\begin{table}[H]
  \centering
  \small
  \caption{
      {The Comparison of MAD-SAE variants and frontier models.}
      (a) In the \textbf{variants comparison}, shaded area distinguishes Oracle from practical variants; bold marks the best practical result.
      (b)In the \textbf{model comparison}, bold and underline denote the best and second-best results, respectively.}  
  \label{tab:variants_models_comparison}

  \begingroup
  \footnotesize
  \setlength{\tabcolsep}{2.2pt}
  \renewcommand{\arraystretch}{1.12}

  \newcommand{\scorecell}[2]{#1}
  \newcommand{\paramcell}[2]{%
      \begin{tabular}[c]{@{}c@{}}
          #1\\[-0.25ex]
          {\tiny (#2)}
      \end{tabular}%
  }

  \resizebox{\textwidth}{!}{%
  \begin{tabular}{@{}c|c|cccccc|cccc!{\vrule width 0.45pt}c@{}}
  \toprule

  \multirow{2}{*}{\textbf{Setting}}
    & \multirow{2}{*}{\textbf{Param.}}
    & \multicolumn{6}{c|}{\textbf{Delegation Appropriateness (DA)}}
    & \multicolumn{4}{c|}{\textbf{Orchestration Utilities (OU)}}
    & \multirow{2}{*}{\textbf{MSQ}$\uparrow$} \\
  \cmidrule(lr){3-8}
  \cmidrule(lr){9-12}
    & &
    UDR$\downarrow$ &
    ODR$\downarrow$ &
    $|\mathcal{D}_{\mathrm{del}}|$ &
    $\mathcal{P_{\text{hit}}}\uparrow$ &
    $\mathcal{R_{\text{hit}}}\uparrow$ &
    $\mathcal{F}_{\text{1 hit}}\uparrow$ &
    $U_p\uparrow$ &
    $U_e\uparrow$ &
    $U_c\uparrow$ &
    Avg.$\uparrow$ &
    \\
  \midrule

  \multicolumn{13}{c}{%
      \textit{(a) MAD-SAE Variants Comparison on Deepseek-V4-Flash-0731}
  } \\
  \midrule

  Gen
    & \multirow{4}{*}{\paramcell{284B}{A13B}}
    & 52.8\%
    & \textbf{9.0\%}
    & 49
    & \textbf{48.0\%}
    & 18.8\%
    & 27.1\%
    & \scorecell{48.7\%}{38-40}
    & \scorecell{\textbf{48.9\%}}{52.6-50.3}
    & \scorecell{31.2\%}{0.5-0.2}
    & 42.9\%
    & 11.6\%
    \\

  Spec
    &
    & 16.8\%
    & 57.1\%
    & 118
    & 41.0\%
    & \textbf{56.2\%}
    & \textbf{47.4\%}
    & \scorecell{\textbf{50.4\%}}{104-102}
    & \scorecell{48.0\%}{40.9-37.8}
    & \scorecell{\textbf{33.1\%}}{0.3-0.1}
    & \textbf{43.8\%}
    & \textbf{20.8\%}
    \\

  Plan-1st
    &
    & \textbf{15.7\%}
    & 61.0\%
    & 122
    & 38.4\%
    & 51.7\%
    & 44.1\%
    & \scorecell{48.0\%}{99-107}
    & \scorecell{42.1\%}{51.8-37.7}
    & \scorecell{29.4\%}{0.3-0.1}
    & 39.9\%
    & 17.6\%
    \\

  \cellcolor{gray!20}Oracle
    &
    & \cellcolor{gray!20}2.2\%
    & \cellcolor{gray!20}1.3\%
    & \cellcolor{gray!20}88
    & \cellcolor{gray!20}88.8\%
    & \cellcolor{gray!20}96.6\%
    & \cellcolor{gray!20}92.5\%
    & \cellcolor{gray!20}\scorecell{49.6\%}{71-72}
    & \cellcolor{gray!20}\scorecell{53.5\%}{39.3-45.3}
    & \cellcolor{gray!20}\scorecell{23.6\%}{0.6-0.2}
    & \cellcolor{gray!20}42.2\%
    & \cellcolor{gray!20}39.1\%
    \\

  \midrule[0.8pt]
  \multicolumn{13}{c}{\textit{(b) Model Comparison with Spec}} \\
  \midrule

  \multicolumn{13}{l}{\textit{Closed-source Models}} \\

  Claude-Sonnet-5
    & -
    & 39.3\%
    & 29.8\%
    & 77
    & 31.1\%
    & 31.7\%
    & 31.4\%
    & \scorecell{45.9\%}{45-53}
    & \scorecell{46.8\%}{54.4-48.0}
    & \scorecell{34.1\%}{3.9-2.0}
    & 42.3\%
    & 13.2\%
    \\

  GPT-5.6-Sol
    & -
    & \textbf{7.8\%}
    & 93.5\%
    & 154
    & 29.6\%
    & \textbf{66.7\%}
    & 41.0\%
    & \scorecell{42.6\%}{84-113}
    & \scorecell{44.0\%}{37.0-29.2}
    & \scorecell{27.4\%}{3.6-1.3}
    & 38.0\%
    & 15.6\%
    \\

  \midrule
  \multicolumn{13}{l}{\textit{Open-source Models}} \\

  Qwen-3.8-Max
    & \paramcell{2.4T}{A95B}
    & 15.7\%
    & 59.7\%
    & 121
    & \underline{46.7\%}
    & 48.6\%
    & \underline{47.6\%}
    & \scorecell{\underline{54.4\%}}{86-72}
    & \scorecell{49.9\%}{39.5-39.5}
    & \scorecell{38.4\%}{1.8-1.1}
    & 47.6\%
    & \underline{22.7\%}
    \\

  GLM-5.3
    & \paramcell{744B}{A40B}
    & \underline{8.9\%}
    & 64.9\%
    & 131
    & 46.4\%
    & \underline{56.9\%}
    & \textbf{51.1\%}
    & \scorecell{50.8\%}{89-86}
    & \scorecell{\underline{52.7\%}}{49.0-54.6}
    & \scorecell{\underline{40.5\%}}{2.3-1.5}
    & \underline{48.0\%}
    & \textbf{24.6\%}
    \\

  Kimi-K3
    & \paramcell{2.8T}{A104B}
    & 50.5\%
    & \underline{16.8\%}
    & 57
    & \textbf{52.7\%}
    & 25.2\%
    & 34.1\%
    & \scorecell{\textbf{56.4\%}}{48-37}
    & \scorecell{50.3\%}{43.5-44.2}
    & \scorecell{\textbf{50.9\%}}{4.4-4.6}
    & \textbf{52.6\%}
    & 17.9\%
    \\

  MiniMax-M3
    & \paramcell{428B}{A23B}
    & 79.7\%
    & \textbf{9.0\%}
    & 25
    & 29.4\%
    & 10.5\%
    & 15.5\%
    & \scorecell{42.8\%}{9-12}
    & \scorecell{42.7\%}{66.3-49.5}
    & \scorecell{37.6\%}{0.4-0.2}
    & 41.0\%
    & 6.3\%
    \\

  GLM-5.3-Flash
    & \paramcell{320B}{A18B}
    & 26.9\%
    & 50.6\%
    & 104
    & 43.2\%
    & 32.4\%
    & 37.0\%
    & \scorecell{51.3\%}{74-70}
    & \scorecell{\textbf{54.0\%}}{35.7-42.0}
    & \scorecell{32.0\%}{0.3-0.1}
    & 45.8\%
    & 16.9\%
    \\

  DeepSeek-V4-Flash-0731
    & \paramcell{284B}{A13B}
    & 16.8\%
    & 57.1\%
    & 118
    & 41.0\%
    & 56.2\%
    & 47.4\%
    & \scorecell{50.4\%}{104-102}
    & \scorecell{48.0\%}{40.9-37.8}
    & \scorecell{33.1\%}{0.3-0.1}
    & 43.8\%
    & 20.8\%
    \\

  \bottomrule
  \end{tabular}%
  }
  \endgroup
\end{table}

%% file: common/figures_tex/performance_cost_efficiency.tex
\begin{figure}[h]
    \centering
    \includegraphics[width=1.0\textwidth]{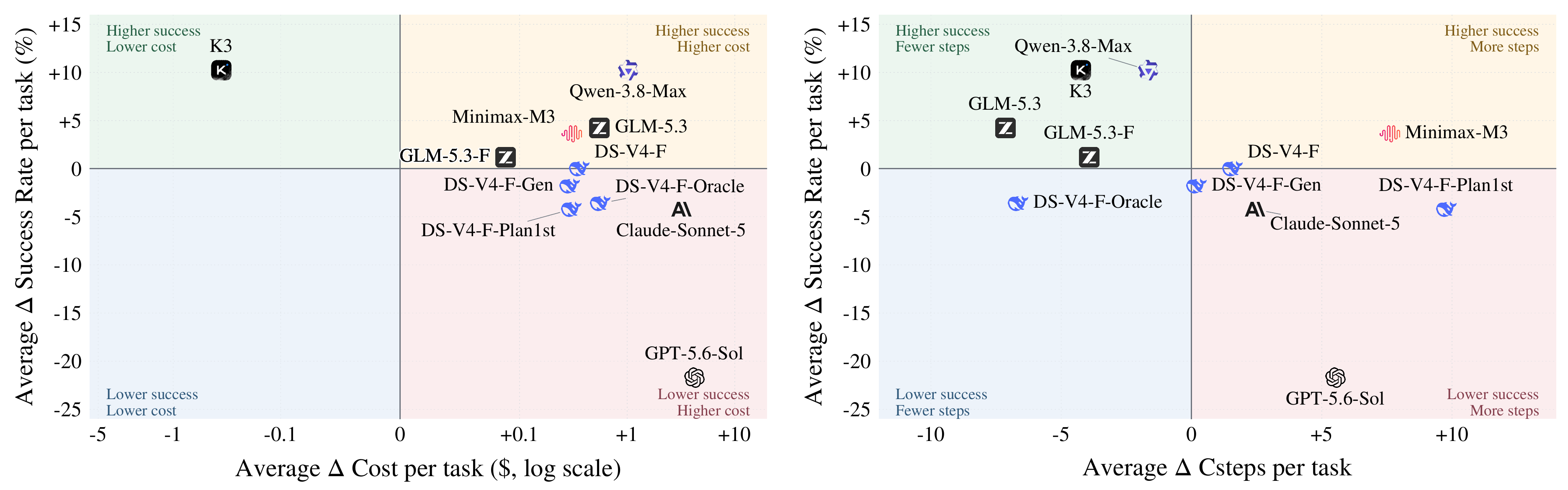}
    \vspace{-0.5cm}
    \caption{Performance-cost (\textbf{left}) and performance-efficiency (\textbf{right}) trade-offs of MAD-SAE relative to SA across the full benchmark $\mathcal{D}$.
Each point represents a model-variant configuration, with changes computed as MAD-SAE minus SA. {All tags are not explicitly stated are in the Spec mode.}
Upper-left regions indicate higher success rates with lower cost or fewer steps.
}
    \label{fig:trade-off}
\end{figure}

%% file: common/figures_tex/case_study_2.tex
\begin{figure}[H]
    \centering
    \includegraphics[width=0.9\textwidth]{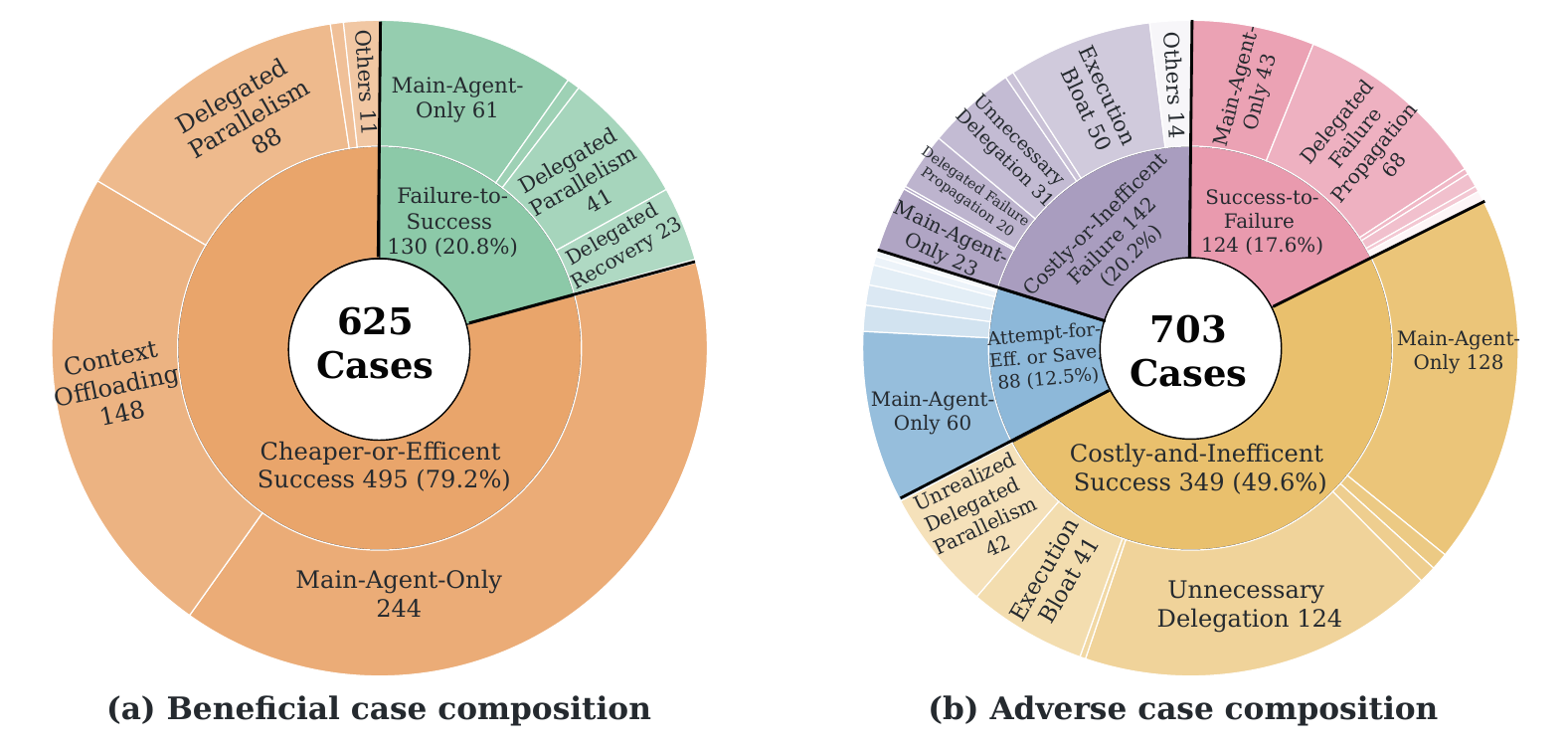}
    \caption{The distribution of beneficial and adverse categories, alongside with their underlying reasons.}
    \label{fig:case_study}
\end{figure}
\vspace{-0.5cm}

%% file: venues/iclr/bibliography.tex
\bibliographystyle{venues/iclr/configuration/iclr2027_conference}
\bibliography{common/references}

%% file: venues/iclr/appendix.tex
\clearpage
\appendix
\input{common/appendix.tex}

%% file: common/appendix.tex



\section{Sub-agent Delegation Statistics}
\label{appx:descriptive_statistics}

After annotating delegation priors for all tasks in MS-Eval, we obtain 89 tasks in $\mathcal{D}_{\text{need}}$ that require sub-agents. We group them by the sub-agent types they demand. Read-Only tasks require only exploratory read-only sub-agents for information gathering, with no editing involved. Writable tasks require only editing-capable sub-agents and need no dedicated exploration. Read \& Write tasks require both, exploratory read-only sub-agents for information gathering and editing sub-agents for producing deliverables.

Table~\ref{tab:domain-descriptive-statistics} summarizes their distribution across domains and these three groups. Among the 89 tasks, Read \& Write tasks are the most prevalent at 44, nearly half, followed by 23 Writable tasks and 22 Read-Only tasks. For the 22 Read-Only tasks, each involves on average 1.0 delegation wave and 8.6 sub-agents. The 23 Writable tasks average 1.0 wave and 2.3 sub-agents. The 44 Read \& Write tasks average 2.2 waves and 7.1 sub-agents. \textbf{Notably, Read-Only tasks demand far more sub-agents than Writable tasks, indicating that exploration is inherently more parallelizable than editing, since it can be sharded across many independent information sources.} Read \& Write tasks require roughly as many sub-agents as Read-Only tasks but twice as many waves, implying their combination of exploration and editing.

Specifically, Enterprise Data Warehouse is dominated by Writable tasks (10 of 15), whereas Stock Market and Railway System consist mostly of Read \& Write tasks (4 of 5 each). Learning Management System and Academic Research exhibit balanced mixtures of all three types.

\input{common/tables/domain_descriptive_statistics.tex}

\section{Further Implementation Details}
\label{appx:further_implementation_details}

\subsection{Measurements for Performance, Efficiency, and Cost}
\label{sec:measurements}

We evaluate MAD-SAE utilities along three dimensions: performance, efficiency, and cost. For each dimension, we define a corresponding measurement as follows.

\noindent \textbf{Success Rate (SR).}
We use Success Rate (SR) to measure task performance. Given a task set $\mathcal{D}$, an evaluator determines whether each task is successfully completed. We denote the subset of successfully solved tasks as $\mathcal{D}^{+}$, and compute SR as:
\begin{equation}
    \label{eq:sr}
    SR = \frac{|\mathcal{D}^{+}|}{|\mathcal{D}|}.
\end{equation}
For evaluating the performance utility $U_p$, we instantiate $\mathcal{D}$ as $\mathcal{D}_{\mathrm{del}}$, the subset of tasks where MAD-SAE actually invokes at least one non-planning sub-agent. This isolates the performance impact introduced by sub-agent orchestration from tasks that are solved entirely by the main-agent.

\noindent \textbf{Critical Steps (CSteps).} Following Kimi-K2.5~\citep{kimiteam2026kimik25visualagentic}, we use Critical Steps (CSteps). A step is counted whenever tools are called or a textual response is returned. An episode consists of $T$ sequential stages: each main-agent step forms one stage, and each parallel sub-agent wave is treated as a stage. For stage $t$, let $S_{\mathrm{main}}^{(t)}$ denote the number of main-agent steps and $S_{\mathrm{sub},i}^{(t)}$ the steps taken by the $i$-th sub-agent in the corresponding wave. CSteps are computed as:
\begin{equation}
    \label{eq:csteps}
    \mathrm{CSteps}
    =
    \sum_{t=1}^{T}
    \left(
        S_{\mathrm{main}}^{(t)}
        +
        \max_i S_{\mathrm{sub},i}^{(t)}
    \right).
\end{equation}
For SA, CSteps reduces to the total number of sequential steps. For MAD-SAE, it captures the bottleneck effect of parallel execution, where each wave completes only when its slowest sub-agent finishes. Since successful task completion is the primary objective, we evaluate efficiency utility $U_e$ as the average CSteps over successfully completed tasks $\mathcal{D}^{+}$ for each model-variant setting.

\noindent \textbf{API Billing.} 
We use the official API pricing for each model to compute the financial cost of all evaluation experiments. As shown in Table~\ref{tab:hyperparameters}, input, cached input, and output tokens are billed separately. Using the token counts from Kimi-Code~\citep{moonshot_kimi_code_agent_tools} for input, cached input, and output tokens ($N_{\text{input}}, N_{\text{cached}}, N_{\text{output}}$), we compute the total financial cost as:
\begin{equation}
  \label{eq:api_billing}
  \text{cost} = \sum_{i \in \{\text{input}, \text{cached}, \text{output}\}} N_{i} \cdot \text{Price}_{i}.
\end{equation}
Following Kimi's official pricing scheme, we do not bill \texttt{inputCacheCreation}. However, since it essentially writes prefixes into the KV cache and thus occupies the context window, it is counted toward the cumulative input tokens, affecting tiered-pricing models such as GPT-5.6-Sol and MiniMax-M3.

\subsection{Function Labeling for Sub-agent Hit}
\label{appx:function_labeling_for_sub-agent_hit}
Computing Sub-agent Hit Scores in \S~\ref{sec:eval} requires constructing an atomic matching unit, termed \emph{sub-task unit}, with specific function labels on both golden and delegated sub-agents. This design avoids penalizing semantically equivalent delegations that differ only in sub-agent count or specialization. In other words, we match \emph{delegation responsibilities} rather than raw sub-agent calls.

We provide 14 function labels for read-only sub-tasks in $\mathcal{V}_{\mathrm{read}}$:
\begin{equation}
\label{eq:read_only_function_labels}
\begin{aligned}
\mathcal{V}_{\mathrm{read}} = \{ & \texttt{rail\_12306}, \texttt{woocommerce}, \texttt{yahoo-finance}, \texttt{snowflake}, \\
& \texttt{scholarly-arxiv}, \texttt{external-fetch}, \texttt{notion}, \texttt{youtube}, \\
& \texttt{playwright-web}, \texttt{filesystem-workspace}, \texttt{canvas}, \texttt{howtocook}, \\
& \texttt{emails}, \texttt{github} \}.
\end{aligned}
\end{equation}
Each label corresponds to a specific domain of exploratory and investigative sub-tasks.

We also provide 11 function labels for editing and writing sub-tasks in $\mathcal{V}_{\mathrm{write}}$:
\begin{equation}
\label{eq:writable_function_labels}
\begin{aligned}
\mathcal{V}_{\mathrm{write}} = \{ & \texttt{excel}, \texttt{word}, \texttt{ppt}, \texttt{notion}, \texttt{email}, \texttt{calendar}, \\
& \texttt{gsheet}, \texttt{gform}, \texttt{pdf}, \texttt{processor}, \texttt{other-write} \}.
\end{aligned}
\end{equation}
Each label corresponds to a writing sub-task targeting a specific object, with \texttt{other-write} serving as a fallback when no other writing-related label applies.
The \texttt{planer} sub-agent is not included in the matching process, considering it does not resolve any specific sub-tasks by design.

\subsection{Model Details}
\label{appx:eval_config}

Table~\ref{tab:hyperparameters} summarizes the inference configuration of each model, with reasoning effort set to each official API's default. It also lists the official API prices, which we use in Eq.~\ref{eq:api_billing} to estimate the cost of our experiments.

For DeepSeek-V4-Flash-0731, we adopt the peak-hour pricing effective from 00:00, August 17, 2026 (Beijing time)~\citep{deepseek_api_pricing}. For MiniMax-M3, we use the permanent 50\% discount offered on the official website under the standard service~\citep{minimax_m3_api_pricing}. All other models use the non-discounted standard prices from their official documentation~\citep{qwen38max_api_pricing, zai_api_pricing, kimi_k3_api_pricing, claude_api_pricing, gpt56sol_api_pricing}. For models billed in CNY, we convert to USD at 1 USD = 6.71 CNY.

\input{common/tables/hyperparameters.tex}

\subsection{MAD-SAE Variants Illustration}
\label{appx:mad-sae-variants-illustration}
Our experiments include a Single-Agent (SA) baseline and four MAD-SAE variants: General (Gen), Specialized (Spec), Plan-1st, and Oracle. All configurations correspond to execution strategies supported by mainstream agent harness frameworks~\citep{moonshot_kimi_code_agent_tools,codex,claude-code,cursor,trae}. Table~\ref{tab:mad-sae-variants} provides an intuitive comparison of these execution paradigms and variants. 

\input{common/tables/modes_outline.tex}

Naturally, the SA paradigm, serving as the most direct runtime solution, is universally supported. The MAD-SAE variants are defined as follows:
\begin{itemize}[leftmargin=12pt]
 \item \textbf{General (Gen):} Employs three generally defined sub-agents: a read-only \texttt{Planner} for strategy formulation, a read-only \texttt{Explorer} for investigation, and a writable \texttt{Coder} for execution and delivery. This aligns with Kimi-Code's built-in taxonomy~\citep{moonshot_kimi_code_agent_tools} and corresponds to the \texttt{general-purpose} sub-agents in Claude Code~\citep{claude-code} and TRAE~\citep{trae} designed for complex research and code modification.

 \item \textbf{Specialized (Spec):} Retains the Gen \texttt{Planner} but granularizes the other agents into domain-specific roles, an approach advocated by Cursor~\citep{cursor_subagents}. The \texttt{Explorer} is divided into four read-only investigators: {Academic-Literature-Researcher} (ALR), {Enterprise-Data-Analyst} (EDA), {Financial-Market-Analyst} (FMA), and {Web-Domain-Researcher} (WDR). Meanwhile, the \texttt{Coder} is split into an {Office-Report-Builder} (for Office documents) and an {External-Workflow-Operator} (for targets like Notion and Google Calendar).
 
\item \textbf{Plan-1st:} Adopts the Gen sub-agent trio but enforces a mandatory initial step where the \texttt{Planner} determines task decomposition and delegation. This planning-first mechanism is supported across all harness frameworks~\citep{moonshot_kimi_code_agent_tools,codex,claude-code,cursor,trae}.

\item \textbf{Oracle:} Utilizes the Gen trio but injects golden priors, specifying exactly which sub-agent handles which sub-task and when, directly into the user query. This aligns with the explicit sub-agent invocation feature highlighted in the official documentation of various frameworks~\citep{moonshot_kimi_code_agent_tools,codex,claude-code,cursor,trae}.

\end{itemize}

\clearpage
\newpage
\section{Details for Case Study}
\label{appx:cases_demo}
\subsection{Taxonomy}
\label{appx:cases_taxonomy}

We perform pairwise comparisons between the Spec variant of MAD-SAE and Single-Agent (SA) across all tasks and eight models. Based on whether MAD-SAE brings performance, efficiency, or cost benefits over SA that would motivate replacing SA, we partition all cases into beneficial and adverse cases, as shown in Table~\ref{tab:case_taxonomy}.

The 625 beneficial cases fall into two categories:

\begin{enumerate}[leftmargin=12pt]
\item \textbf{Failure-to-Success (F2S)} includes 130 cases where MAD-SAE succeeds while SA fails. We assign all such cases to this category regardless of any efficiency or cost advantage, since task success is the first principle.

\item \textbf{Cheaper-or-Efficient Success (CES)} includes 495 cases where both MAD-SAE and SA succeed, and MAD-SAE is either more efficient (strictly fewer critical steps) or cheaper (strictly lower economic cost) than SA.

\end{enumerate}

However, MAD-SAE (Spec) does not always realize advantages over SA. We therefore have 703 adverse cases, which fall into four categories:

\begin{enumerate}[leftmargin=12pt]
\item \textbf{Success-to-Failure (S2F)} includes 124 cases where MAD-SAE fails on tasks that SA solves correctly, indicating a performance regression.

\textbf{Costly-and-Inefficient Success (CIS)} includes 349 cases where both succeed, but MAD-SAE incurs strictly higher cost than SA and achieves no reduction in critical steps, adding overhead without either efficiency or cost savings.


\item \textbf{Attempt-for-Efficiency-or-Saving (AES)} includes 88 cases where both fail, but MAD-SAE attempts more efficient or cheaper strategies that proved effective in beneficial cases.

\item \textbf{Costly-or-Inefficient Failure (CIF)} includes 142 cases where both fail, but MAD-SAE uses strictly more steps or incurs strictly higher cost than SA.
\end{enumerate}

\input{common/tables/case_taxonomy.tex}

\subsection{Underlying Reasons}
\label{appx:underlying_reasons}

To understand why MAD-SAE yields benefits or drawbacks, we conduct a case study. For beneficial cases, we identify five underlying reasons:
\begin{enumerate}[leftmargin=12pt]
  \item \textbf{Main-Agent-Only (MAO):} Under MAD-SAE, no sub-agent is delegated. The benefit comes from the main-agent finding a correct, cheaper, or more step-efficient solution than SA.

  \item  \textbf{Context Offloading (CO):} Isolated sub-agent contexts absorb heavy tasks that would otherwise occupy the SA context, allowing the main-agent to focus on the core of the task and thereby gain benefits.
  
  \item  \textbf{Delegated Parallelism (DP):} MAD-SAE enables parallel sub-agent delegation, decomposing what would be serial work into concurrently executed sub-tasks and thus offering efficiency gains.
  
  \item  \textbf{Delegated Recovery (DR):} Sub-agents introduce verifying and recovery operations rarely seen in SA execution, validating deliverables and fixing bugs during the runtime session, which improves robustness.
  
  \item  \textbf{Others (OTH):} Cases not covered by the four categories above.
  
\end{enumerate}

For adverse cases, the reasons are more varied and fall into twelve categories:

\begin{enumerate}[leftmargin=12pt]
\item \textbf{Main-Agent-Only (MAO):} Under MAD-SAE, no sub-agent is delegated. The drawback comes from the main-agent finding an incorrect, costlier, or more step-consuming solution than SA.

\item \textbf{Delegated Context Amplification (DCA):} Each sub-agent repeatedly carries a large context, multiplying token costs.

\item \textbf{Delegated Failure Propagation (DFP):} Failures, corrupted outputs, or unverified outputs in the delegation chain propagate and degrade the overall result.

\item \textbf{Unnecessary Delegation (UD):} Delegation is not necessary for the task, and the orchestration, handoff, and verification costs exceed the actual gains.

\item \textbf{Misdirected Delegation (MD):} Delegated content mismatches task requirements, such as specification drift, capability mismatch, wrong tool use, or incorrect hierarchy construction.

\item \textbf{Execution Bloat (EB):} Redundant verification, rework, repeated operations, or detours inflate steps or costs.

\item \textbf{Unrealized Delegated Parallelism (UDP):} The run incurs the intended delegation overhead, but the delegated parallel work does not deliver the expected efficiency benefit.

\item \textbf{Context Offloading Attempt (COA):} Specific to AES cases. Under MAD-SAE, the model attempts to share the main-agent's context burden by delegating sub-agents, but this does not turn SA failure into MAD-SAE success.

\item \textbf{Delegated Parallelism Attempt (DPA):} Specific to AES cases. Under MAD-SAE, the model attempts to delegate sub-agents in parallel and improve the task solving efficiency, but this does not turn SA failure into MAD-SAE success.

\item \textbf{Delegated Recovery Attempt (DRA):} Specific to AES cases. Under MAD-SAE, sub-agents attempt verifying and recovery operations, but this does not turn SA failure into MAD-SAE success.

\item \textbf{Missing Final Integration (MFI):} Specific to AES cases. Delegation completes the main work, but the main-agent fails to finish final integration, summarization, or acceptance.

\item  \textbf{Others (OTH):} Cases not covered by the categories above.

\end{enumerate}
The distribution of underlying reasons varies across categories. We visualize and analyze this in the case study in \S~\ref{sec:mad-sae-analysis}.

\clearpage
\newpage
\section{Further Case Study}
\label{appx:further_case_study}
\vspace{-0.3cm}
\S~\ref{sec:mad-sae-analysis} quantified the overall benefits and drawbacks of MAD-SAE (Spec) relative to SA across the eight models in Table~\ref{tab:hyperparameters} and summarized the underlying mechanisms in Fig.~\ref{fig:case_study}. Here we ask a more precise question: for each model, do the observed gains or regressions arise from the main agent's own execution or from sub-agent delegation? This distinction matters because Fig.~\ref{fig:case_study} shows that many beneficial and adverse cases derive from Main-Agent-Only (MAO) behavior, with 305 such cases among the beneficial cases and 254 among the adverse cases. We aim to determine whether this pattern is specific to certain models or shared across all models, which would clarify the source of MAD-SAE's overall advantage or disadvantage.

To support this analysis, Table~\ref{tab:per_model_case_attribution} reports the number of beneficial and adverse cases per model, together with a breakdown into MAO and Non-MAO cases. Non-MAO cases are those whose underlying reasons are not labeled as MAO and whose outcomes mainly arise from sub-agent delegation. We also compare Failure-to-Success (F2S) and Success-to-Failure (S2F) cases per model, since these flip the success outcome when switching from SA to MAD-SAE and thus drive changes in the success rate. We discuss models with more beneficial than adverse cases in Appx.~\ref{appx:benefits-outweigh-drawbacks}, and the reverse in Appx.~\ref{appx:drawbacks-outweigh-benefits}.
\vspace{-0.2cm}
\input{common/tables/benefits_and_drawbacks_per_model.tex}

\subsection{MAD-SAE (Spec) Benefits Outweigh Drawbacks}
\label{appx:benefits-outweigh-drawbacks}

\noindent\textbf{GLM-5.3: gains driven by effective delegation, still at a cost.} Delegation clearly accounts for 63 of the 89 beneficial cases (70.8\%), excluding one Non-MAO case labeled Other, with \textbf{Context Offloading} (Fig.~\ref{fig:context_offload_glm5_3}) and \textbf{Delegated Parallelism}  (Fig.~\ref{fig:delegated_parallelism_glm5_3}) contributing 31 and 27 cases, respectively. Similarly, 11 of the 16 Failure-to-Success cases stem from delegationX. GLM-5.3's improvement therefore mainly reflects sub-agents absorbing context and executing in parallel, while Main-Agent-Only improvement plays a smaller role. This aligns with its high efficiency utility ($U_e=52.7\%$) in Table~\ref{tab:per_model_case_attribution}. Its $U_c=40.5\%$ further shows that these gains still come at added cost, consistent with the higher-cost, higher-success pattern it occupies, e.g., Costly-and-Inefficient-Success, in Fig.~\ref{fig:trade-off}.
\input{common/figures_tex/model_wise_case_study/glm_5_3.tex}

\noindent\textbf{GLM-5.3-Flash: a narrow edge carried entirely by delegation.} Its Main-Agent-Only cases split 32 beneficial to 41 adverse, indicating that MAO does not provide notably better solutions in this setting. Its overall edge over SA therefore comes entirely from delegation. All 53 beneficial Non-MAO cases carry an identified delegation mechanism (Fig.~\ref{fig:glm_5_3_flash_case_study}), with 48 falling under Cheaper-or-Efficient-Success: 33 from Context Offloading and 15 from Delegated Parallelism. Given a net gain of only 2 successful tasks by comparing Failure-to-Success (7 cases) with Success-to-Failure (5 cases), GLM-5.3-Flash's more salient benefit is execution efficiency, matching its top-ranked $U_e=54.0\%$ in Table~\ref{tab:per_model_case_attribution}. Its $U_c=32.0\%$ indicates that this efficiency still comes at extra cost.
\input{common/figures_tex/model_wise_case_study/glm_5_3_flash.tex}

\noindent\textbf{K3: main-agent improvement leads, but delegation contributes clearly too.} Main-Agent-Only (MAO) accounts for 89 of K3's 130 beneficial cases (68.5\%), including 76 Cheaper-or-Efficient-Success and 13 Failure-to-Success cases. MAO yields a net advantage of 67 cases (89 beneficial vs. 22 adverse), and beneficial cases outnumber adverse cases by 94 overall. This indicates much of K3's MAD-SAE gain comes from the main agent's own execution path.  Nevertheless, the remaining 41 beneficial cases are all attributed to an identified delegation mechanism, e.g., Context Offloading, Delegated Parallelism, etc. On top of that, K3's Failure-to-Success cases significantly outnumber its Success-to-Failure cases, primarily from 13 MAO and 9 \textbf{Delegated Parallelism} (Fig.~\ref{fig:delegated_parallelism_k3}) cases. These results demonstrate clearly that K3's MAD-SAE benefits are jointly supported by both the better main-agent execution and the high-quality delegation, which explain K3's broad improvement in performance, cost, and efficiency in Fig.~\ref{fig:trade-off}, as well as three orachestration utilities all exceeding 50\% in Table~\ref{tab:per_model_case_attribution}. However, broad gains do not imply that appropriate delegation is fully recalled, according to K3's  50.5\% UDR.
\input{common/figures_tex/model_wise_case_study/kimi_k3.tex}


\subsection{MAD-SAE (Spec) Drawbacks Outweigh Benefits}
\label{appx:drawbacks-outweigh-benefits}

\noindent\textbf{Qwen-3.8-Max: more adverse cases, yet the largest success-rate gain.} Although adverse cases outnumber beneficial ones by 6, Qwen-3.8-Max achieves the largest success-rate gain (+10.2 points), because its 20 Failure-to-Success (F2S) cases far exceed its 3 Success-to-Failure (S2F) cases. Delegation plays a central role in these flips: 14 of the 20 F2S cases are delegation-attributed (Fig.~\ref{fig:qwen_3_8_max_case_study}), led by Delegated Recovery (7) and Delegated Parallelism (5). This gain, however, comes with overhead: 25 of its 48 Costly-and-Inefficient-Success cases are attributed to Unnecessary Delegation. Delegation thus simultaneously drives additional successes and introduces redundancy on a subset of tasks, a pattern consistent with its utilities in Table~\ref{tab:per_model_case_attribution}: $U_p=54.4\%$, $U_c=38.4\%$, and $U_e=49.9\%$, indicating higher success and higher cost with roughly flat efficiency. The step-count reduction over the full benchmark is examined separately in Fig.~\ref{fig:trade-off}.
\input{common/figures_tex/model_wise_case_study/qwen_3_8_max.tex}

\noindent \textbf{DeepSeek-V4-Flash-0731: a flat success rate masking two offsetting effects.} Failure-to-Success (F2S) and Success-to-Failure (S2F) are tied at 11 cases each, matching its zero success-rate change in Fig.~\ref{fig:trade-off}. This balance arises from two offsetting effects: Main-Agent-Only contributes a net loss (F2S 4 v.s. S2F 6), while delegation-attributed cases yield a net gain of 2 (F2S 7 v.s. S2F 5), so the two roughly cancel. This is consistent with a slightly positive $U_p=50.4\%$ in Table~\ref{tab:per_model_case_attribution}. Its adverse cases are heavily concentrated in Costly-and-Inefficient-Success (71 of 91, 78.0\%; Fig.~\ref{fig:dsv4_flash_0731_case_study}), of which 40 stem from \textbf{Unnecessary Delegation} (Fig.~\ref{fig:unnessary_delegation_dsv4}), the single largest source of drawbacks across all its adverse cases (43 occurrences in total). Overall, delegation provides only a limited correctness benefit here, while extensive redundant delegation still adds overhead, consistent with $U_e=48.0\%$ and $U_c=33.1\%$ in the same table.
\input{common/figures_tex/model_wise_case_study/deepseek_v4_flash_0731.tex}

\noindent\textbf{Claude-Sonnet-5: failure propagation outweighs main-agent gains.} MAO's beneficial and adverse counts are nearly balanced (46 vs.\ 43), with its Failure-to-Success (F2S) and Success-to-Failure (S2F) both equal to 7, so MAO contributes no net change. Claude-Sonnet-5's overall success-rate still falls by 7 tasks. The driver is Delegated Failure Propagation, responsible for 14 of its 22 S2F cases (Fig.~\ref{fig:claude_sonnet_5_case_study}), showing that insufficiently verified delegation outcomes are the main source of the regression. Its overhead problem, however, is not purely a delegation story: 31 of its 48 Costly-and-Inefficient-Success cases are labeled MAO. Claude-Sonnet-5 is thus hurt by both delegation failures and inefficient main-agent execution, consistent with all three utilities in Table~\ref{tab:per_model_case_attribution} falling below 50\% and with its more expensive, more inefficient, lower success position in Fig.~\ref{fig:trade-off}.
\input{common/figures_tex/model_wise_case_study/claude_sonnet_5.tex}

\noindent\textbf{GPT-5.6-Sol: over-delegation undermines both correctness and efficiency.} With 137 adverse cases against only 29 beneficial ones, and all 7 of its Main-Agent-Only (MAO) cases falling on the adverse side, GPT-5.6-Sol's problems are concentrated in delegation. \textbf{Delegated Failure Propagation} (Fig.~\ref{fig:delegated_failure_propa_gpt_5_6}) accounts for 33 of its 45 Success-to-Failure cases (Fig.~\ref{fig:gpt_5_6_sol_case_study}); among its 55 Costly-and-Inefficient-Success cases, \textbf{Unnecessary Delegation} and \textbf{Unrealized Delegated Parallelism} (Fig.~\ref{fig:udp_gpt_5_6}) contribute 25 and 20 respectively. Failure propagation explains the correctness regression, and redundant, unrealized delegation explains the execution overhead. This is consistent with its 93.5\% ODR, all three utilities below 50\% in Table~\ref{tab:per_model_case_attribution}, and its pronounced regression in Fig.~\ref{fig:trade-off}: frequent sub-agent invocation here never converts into effective collaboration.
\input{common/figures_tex/model_wise_case_study/gpt_5_6_sol.tex}

\noindent\textbf{MiniMax-M3: outcomes dominated by main-agent behavior.} \textbf{Main-Agent-Only (MAO)} (Fig.~\ref{fig:better_mao_mm3}) dominates both sides (Fig.~\ref{fig:minimax_m3_case_study}), accounting for 55 of 63 beneficial and 79 of 103 adverse cases. Of its 26 Failure-to-Success (F2S) cases, 24 are labeled MAO. Together with 14 MAO-labeled Success-to-Failure (S2F) cases, these MAO cases net a gain of 10 successful tasks, offsetting a net loss of 4 from the remaining Non-MAO successes and yielding the model's overall success-rate gain of 3.6 points. At the same time, 33 of its 42 Costly-and-Inefficient-Success cases are also labeled MAO, indicating that the main-agent likewise generates many successes with added overhead and no efficiency gain. This composition echoes its UDR of 79.7\% in Table~\ref{tab:per_model_case_attribution}, although a high MAO share cannot itself substitute for UDR, which is computed from task-level priors. Its performance gain on $\mathcal{D}$ and its $U_p=42.8\%<50\%$ in the same table are not in tension: main-agent improvement masks a regression within the delegation subset, and the larger adverse count overall cannot be attributed to delegation overhead alone.
\input{common/figures_tex/model_wise_case_study/minimax_m3.tex}

\noindent \textbf{Takeaway.} This model-wise analysis shows that MAD-SAE's overall value rests on the joint contribution of main-agent change and delegation mechanisms, and their relative weight is highly model-dependent. Deciding whether to adopt MAD-SAE should draw on full-benchmark success rate, cost, and efficiency together; judging whether delegation itself is effective requires specific mechanisms and delegation-subset metrics in tandem.

\clearpage
\newpage
\subsection{Typical Cases Illustration}
\label{appx:typical_cases_illustration}

We select several typical cases to illustrate the benefits of MAD-SAE over the SA baseline in Appx.~\ref{appx:beneficial_cases_illustration} and its drawbacks in Appx.~\ref{appx:adverse_cases_illustration}.

\subsubsection{Beneficial Cases}
\label{appx:beneficial_cases_illustration}

\noindent \textbf{Context Offloading} is the primary component of GLM-5.3's Cheaper-or-Efficient-Success category (see Fig.~\ref{fig:glm_5_3_case_study}). We use the \texttt{canvas-course-notion-wiki} case from this category for illustration in Figure~\ref{fig:context_offload_glm5_3}. SA encounters numerous Notion API issues when solving the task, whereas Spec resolves these Notion API bugs more efficiently through a dedicated sub-agent. This offloads the error messages that would otherwise occupy the main-agent's context, keeping it cleaner and allowing it to focus on core task-solving logic, thereby improving efficiency.

\noindent \textbf{Delegated Parallelism} is an important component of GLM-5.3's Cheaper-or-Efficient-Success category (see Fig.~\ref{fig:glm_5_3_case_study}). We use the \texttt{canvas-enrollment-ppt-gsheet} case from this category for illustration in Fig.~\ref{fig:delegated_parallelism_glm5_3}. The optimal solution annotated in our Delegation Priors is to use parallelized explorers to read course information separately, followed by three dedicated writable sub-agents responsible for producing and delivering xlsx, pptx, and Google Sheet files. Spec perfectly matches this optimal solution, totaling 23 critical steps and saving 13 steps compared to SA. SA suffers from serial execution, where xlsx, pptx, and Google Sheet must be completed one after another, which weakens execution efficiency.

For Kimi-K3, Delegated Parallelism helps correct failures (see Fig.~\ref{fig:kimi_k3_case_study}). Fig.~\ref{fig:delegated_parallelism_k3} shows the \texttt{terminal-yf-canvas-excel-gcal-email} task from Kimi-K3's Failure-to-Success category. SA's serial solution hits the 100-step limit and fails to deliver, whereas Spec distributes the execution pressure of a single agent through reasonably parallelized paginated course exploration, completing a successful delivery in only 67 steps.

\noindent \textbf{Better Main-Agent-Only Solutions} dominates Minimax-M3's Failure-to-Success category and is the main reason MAD-SAE (Spec) improves this model's performance (see Fig.~\ref{fig:minimax_m3_case_study}). As shown in Fig.~\ref{fig:better_mao_mm3} (\texttt{scholarly-arxiv-survey-word-notion-gcal}), although the Delegation Prior expects parallel paper explorers and writable sub-agents to solve the problem more efficiently, in Spec the main-agent completes the task alone. By applying some verification during problem solving, it reduces the SA approach that hit the 100-step limit to a success in 48 steps.

\input{common/figures_tex/case_study_classic_cases/context_offload.tex}
\input{common/figures_tex/case_study_classic_cases/delegated_parallelism.tex}
\input{common/figures_tex/case_study_classic_cases/delegated_parallelism_k3.tex}
\input{common/figures_tex/case_study_classic_cases/better_main_agent_only.tex}

\subsubsection{Adverse Cases}
\label{appx:adverse_cases_illustration}
\noindent \textbf{Unnecessary Delegation} plays an important role in DeepSeek-V4-Flash-0731's Costly-and-Inefficient-Success category (see Fig.~\ref{fig:dsv4_flash_0731_case_study}). We select the \texttt{academic-literature-review} case (Fig.~\ref{fig:unnessary_delegation_dsv4}) from this category for a comparative illustration between MAD-SAE (Spec) and the SA baseline. The optimal solution in our annotated delegation priors requires only one wave of exploratory sub-agents to conduct parallel exploration of five non-overlapping RAG topics. The subsequent process of producing Literature\_Review.docx is not cumbersome and can be handled by the main-agent itself. Spec perfectly uses five academic-literature-researcher sub-agents for parallel paper exploration, but then assigns an additional office-report-builder specifically to generate the docx report. This operation is semantically valid and reasonable, yet the office-report-builder consumes 70 steps, more than SA's total critical steps (CSteps) of 51. Combined with the main-agent's execution steps, its total CSteps reach 97, nearly twice that of SA. This significantly affects execution efficiency and increases execution cost.

\noindent \textbf{Delegated Failure Propagation} dominates GPT-5.6-Sol's Success-to-Failure cases (see Fig.~\ref{fig:gpt_5_6_sol_case_study}). We use \texttt{canvas-ta-workload-excel-email} for visualizatio in Fig.~\ref{fig:delegated_failure_propa_gpt_5_6}. Here, Spec's \texttt{enterprise-data-analyst} runs only a few steps and returns a compressed, lossy summary, which the downstream Excel, Notion, Email builders directly trust such the summary. This error propagation causes GPT-5.6-Sol to fail a task that SA solves correctly with 21 CSteps, despite spending 41 CSteps.

\noindent \textbf{Unrealized Delegated Parallelism} is the second most common factor behind GPT-5.6-Sol's Costly-and-Inefficient-Success (see Fig.~\ref{fig:gpt_5_6_sol_case_study}). As shown in Fig.~\ref{fig:udp_gpt_5_6} (\texttt{canvas-enrollment-ppt-gsheet}), ideally, course information acquisition and deliverable generation should be completed synchronously by parallelized sub-agents to accelerate task solving. However, GPT-5.6-Sol calls four sub-agents and executes them serially. This unnecessary and unreasonable delegation makes Spec require 63 steps for a task that SA can complete in 25 critical steps. This contrasts sharply with GLM-5.3 (Fig.~\ref{fig:delegated_parallelism_glm5_3}), which efficiently solves the same task by reasonably parallelizing sub-agent scheduling.

\noindent \textbf{Execution Bloat} accounts for a considerable proportion of GPT-5.6-Sol's Costly-and-Inefficient-Success category  (see Fig.~\ref{fig:gpt_5_6_sol_case_study}). We use the \texttt{terminal-arxiv-scholarly-notion-word-excel} case in this setting for illustration in Fig.~\ref{fig:execution_bloat_gpt_5_6}. In the Delegation Priors, we expect parallelized exploratory sub-agents to first collect information, followed by parallelized writable sub-agents responsible for editing and delivering Notion, Word, and Excel files. However, this Spec generates five serial single-agent waves, including a resume of an academic-literature-researcher. The sub-agent delegation here achieves no parallelism and instead introduces rework costs. This makes task solving more complex, requiring 18 more steps than the SA solution.

\input{common/figures_tex/case_study_classic_cases/unnessary_delegation.tex}
\input{common/figures_tex/case_study_classic_cases/delegated_failure_propa.tex}
\input{common/figures_tex/case_study_classic_cases/udp.tex}
\input{common/figures_tex/case_study_classic_cases/execution_bloat.tex}

\clearpage
\newpage

\section{MS-Eval Kit Details}
\label{app:ms-eval-kit-details}
Fig.~\ref{fig:main_picture} shows the overall architecture of MS-Eval with the example of \texttt{arxiv-survey-presentation}. We continue with this task and provide more complete information to facilitate understanding of the MS-Eval Benchmark (Appx.~\ref{appx:benchmark}) and Harness (Appx.~\ref{appx:harness}). In Appx.~\ref{appx:benchmark}, we provide the full task description and delegation prior for \texttt{arxiv-survey-presentation}, supplementing the summarized content in Fig.~\ref{fig:main_picture} (Benchmark). In Appx.~\ref{appx:harness}, we provide the principles and tools for the basic, main-agent, and sub-agent levels, illustrating the core constraints needed to govern agent behavior across different paradigms in working scenarios.

\subsection{Benchmark}
\label{appx:benchmark}

\input{common/appendix_prompt_setup}

\subsubsection{Task Description}
\label{app:arxiv-task-description}
We provide the full task description for \texttt{arxiv-survey-presentation} in Table~\ref{tab:arxiv-full-task}.
This is a representative working case from the Academic Research scenario. It requires the following MCP servers: \texttt{playwright\_with\_chunk} for web browsing, \texttt{terminal} for command-line operations, file inspection, and verification, \texttt{arxiv-latex} for searching, listing, reading, and downloading arXiv papers, and \texttt{excel} and \texttt{pptx} for creating and writing xlsx and pptx files. Using these MCP servers, the agent must complete three milestone sub-tasks: (1) inspect survey-methodology pages to understand the local arXiv candidate space relevant to the task, (2) select exactly five source-supported papers, and (3) create the two required files Survey\_Data.xlsx and NLP\_Survey.pptx. This illustrates the reliance of working scenarios on domain-specific tools and the complexity of tasks involving multiple sub-tasks.

\subsubsection{Delegation Priors}
\label{app:arxiv-delegation-priors}
We exhibit the detailed and specific delegation prior for each task, as shown in Table~\ref{tab:arxiv-full-prior}. We assign three waves of sub-agents to complete the three milestone sub-tasks mentioned in Appx.\ref{app:arxiv-task-description}. The first wave consists of two exploratory sub-agents that perform read-only tasks in parallel. One inspects the survey-methodology page, and the other scouts the task-relevant local arXiv candidate space. This allows the agent to fully understand the current environment state. The second wave consists of five read-only exploratory sub-agents running in parallel. Each sub-agent investigates one paper and returns compact records. After obtaining these records, the third wave assigns two writable sub-agents to simultaneously edit the deliverables. They are responsible for Survey\_Data.xlsx and NLP\_Survey.pptx, respectively. Such detailed delegation priors not only indicate whether the current task requires delegating sub-agents, but also provide golden priors for the Oracle Study. This helps us understand MAD-SAE's utility under near-perfect delegation.

\subsection{Harness}
\label{appx:harness}
For the single-agent (SA) paradigm, we use the Basic Principles \& Tools in Appx.~\ref{app:basic-runtime-prompt} to govern agent behavior. For the MAD-SAE paradigm, the main-agent inherits these Basic Principles \& Tools while adding variant-specific sub-agent delegation information for different MAD-SAE variants (Appx.~\ref{app:main-runtime-prompts}). This includes adding Delegation Norms, Delegation Examples, and Sub-Agent Introduction to the principles, as mentioned in Fig.~\ref{fig:main_picture} (Harness). For the tools, we add \texttt{Agent} and \texttt{AgentSwarm} to enable the main-agent to delegate sub-agents. For sub-agent principles and tools (Appx.~\ref{app:subagent-principles-tools}), different sub-agent types may have different allowed MCP tools, but none are granted access to \texttt{Agent}, \texttt{AgentSwarm}, or the task-termination tool \texttt{claim\_done}.

\subsubsection{Basic Principles \& Tools}
\label{app:basic-runtime-prompt}
Table~\ref{tab:single-agent-principles} presents the Basic Principles \& Tools that govern SA agents and also underlie main-agent control in MAD-SAE. The runtime profile specifies name, description, and override, where override determines whether to replace a same-named built-in profile. tools grants all task-authorized MCP tools via \texttt{mcp\_\_*}, while disallowedTools disables Kimi-Code built-ins, ensuring all functionality is accessed only through \texttt{mcp\_\_*}. subagents is set to None because SA requires no sub-agent calls. The profile further defines the accessible workspace, \textbf{Core Responsibilities, Visible Boundary, and Completion Protocol}, which guide valid task solving, restrict resource access, and specify termination. Notably, many models fail to call \texttt{mcp\_\_local\_\_claim\_done} despite this requirement.

\subsubsection{Main-Agent Principles \& Tools}
\label{app:main-runtime-prompts}
\noindent \textbf{General (Gen)} is a MAD-SAE variant that defines three sub-agents: a read-only \texttt{Plan} for strategy planning, a read-only \texttt{Explore} for investigation, and a writable \texttt{Coder} for editing. As shown in Table~\ref{tab:gen-principles}, the \texttt{subagents} field in the Runtime Profile changes from \texttt{None} in Table~\ref{tab:single-agent-principles} to names of three sub-agents. Relative to Table~\ref{tab:single-agent-principles}, Gen adds two \textbf{Core Responsibilities}: assessing whether to delegate sub-agents and reusing sub-agent outputs under their return contracts. It further introduces Subagent Delegation and Sub-Agent Orchestration Rules, collectively called \textbf{Delegation Norms}, which specify when and how to schedule sub-agents. The Available Sub-Agent Types section, referred to as \textbf{Sub-Agent Introduction}, briefly describes each sub-agent type and its strengths. Finally, the Examples section, termed \textbf{Delegation Examples}, illustrates in what contexts to delegate and which sub-agents to schedule.

\noindent \textbf{Specialized (Spec)} is a MAD-SAE variant that inherits Gen's \texttt{Planner}, while Gen's \texttt{Explorer} and \texttt{Coder} are split into five and two more domain-specific sub-agents, respectively. As shown in Table~\ref{tab:spec-principles}, Main-Agent Principles \& Tools (Spec) and Main-Agent Principles \& Tools (Gen) in Table~\ref{tab:gen-principles} are largely identical, except that the subagents parameter, the corresponding entries in Sub-Agent Orchestration Rules, Available Sub-Agent Types, and Examples are replaced with Spec-specific content.

\noindent \textbf{Plan-1st} extends Gen's sub-agent definitions with a mandatory initial planning step: the first action must be a \texttt{Planner} call that determines sub-task partitioning and subsequent sub-agent scheduling. Consequently, relative to Gen (Table~\ref{tab:gen-principles}), Plan-1st adds a Plan-First Protocol (Mandatory) to enforce this initial planning, as shown in Table~\ref{tab:planfirst-principles}.

\noindent \textbf{Oracle} serves as the variant for the golden study under MAD-SAE. It fully inherits Main-Agent Principles \& Tools (Gen) in Table~\ref{tab:gen-principles}, except that it appends the delegation prior directly to the end of the task description.

\subsubsection{Sub-Agent Principles \& Tools}
\label{app:subagent-principles-tools}
Tables~\ref{tab:planner-principles},\ref{tab:explorer-principles},\ref{tab:coder-principles} present the principles and tools for the \texttt{Planner}, \texttt{Explorer}, and \texttt{Coder} sub-agents, while Tables~\ref{tab:alr-principles},\ref{tab:eda-principles},\ref{tab:fma-principles},\ref{tab:wdr-principles},\ref{tab:ewo-principles},\ref{tab:orb-principles} present those for the Academic-Literature-Researcher, Enterprise-Data-Analyst, Financial-Market-Analyst, Web-Domain-Researcher, External-Workflow-Operator, and Office-Report-Builder sub-agents. For all sub-agents, the Runtime Profile shares the same parameter schema as the Basic Principles \& Tools in Table~\ref{tab:single-agent-principles}, with an additional \texttt{whenToUse} field specifying when the sub-agent is dispatched. The \texttt{tools} and \texttt{disallowedTools} fields define which tools are available or unavailable, determined by the sub-agent's read-only or writable nature and its domain-specialized role. In particular, \texttt{Planner}, \texttt{Explorer}, Academic-Literature-Researcher, Enterprise-Data-Analyst, Financial-Market-Analyst, and Web-Domain-Researcher are read-only, whereas \texttt{Coder}, External-Workflow-Operator, and Office-Report-Builder are writable. The Constraints section restricts each sub-agent to domain-specific read-only or writable operations, and the Visible Boundary limits the scope of resources accessible to the sub-agent.

\input{common/tables/principle_tools/arxiv_full_task_table}

\input{common/tables/principle_tools/arxiv_full_prior_table}

\input{common/tables/principle_tools/single_agent_principles_table}
\input{common/tables/principle_tools/gen_principles_table}

\input{common/tables/principle_tools/spec_principles_table}

\input{common/tables/principle_tools/planfirst_principles_table}

\input{common/tables/principle_tools/planner_principles_table}

\input{common/tables/principle_tools/explorer_principles_table}

\input{common/tables/principle_tools/coder_principles_table}

\input{common/tables/principle_tools/alr_principles_table}

\input{common/tables/principle_tools/eda_principles_table}

\input{common/tables/principle_tools/fma_principles_table}

\input{common/tables/principle_tools/wdr_principles_table}

\input{common/tables/principle_tools/ewo_principles_table}

\input{common/tables/principle_tools/orb_principles_table}

\endgroup

%% file: common/tables/domain_descriptive_statistics.tex
\begin{table}[!htbp]
  \centering
  \caption{Distribution of the 89 tasks in $\mathcal{D}_{\text{need}}$ by domain and sub-agent type. Tasks are grouped into Read-Only, Writable, and Read \& Write. For each group, we report the number of tasks and the min/max/mean of delegation waves and sub-agents.}
  \label{tab:domain-descriptive-statistics}
  \begingroup
  \footnotesize
  \setlength{\tabcolsep}{3pt}
  \renewcommand{\arraystretch}{1.15}
  \resizebox{\textwidth}{!}{%
  \begin{tabular}{@{}ccccccccccc@{}}
  \toprule
  \multirow[c]{2}{*}{\textbf{Domain}} &
  \multirow[c]{2}{*}{\shortstack{\textbf{Task}\\\textbf{Number}}} &
  \multicolumn{3}{c}{\textbf{Read-Only}} &
  \multicolumn{3}{c}{\textbf{Writable}} &
  \multicolumn{3}{c}{\textbf{Read \& Write}} \\
  \cmidrule(lr){3-5}
  \cmidrule(lr){6-8}
  \cmidrule(lr){9-11}
  & & \textbf{Number} & \textbf{Wave} & \shortstack{\textbf{Sub-}\\\textbf{Agent}} &
  \textbf{Number} & \textbf{Wave} & \shortstack{\textbf{Sub-}\\\textbf{Agent}} &
  \textbf{Number} & \textbf{Wave} & \shortstack{\textbf{Sub-}\\\textbf{Agent}} \\
  \midrule
  Learning Management System & 25 & 11 & 1/1/1.0 & 3/22/14.3 & 5 & 1/1/1.0 & 1/3/2.2 & 9 & 2/3/2.1 & 4/25/11.2 \\
  Enterprise Data Warehouse & 15 & 0 & - & - & 10 & 1/2/1.1 & 1/4/2.4 & 5 & 2/2/2.0 & 4/5/4.6 \\
  E-Commerce & 15 & 4 & 1/1/1.0 & 2/2/2.0 & 2 & 1/1/1.0 & 2/2/2.0 & 9 & 2/3/2.3 & 2/7/4.1 \\
  Stock Market & 5 & 1 & 1/1/1.0 & 2/2/2.0 & 0 & - & - & 4 & 2/3/2.3 & 6/9/7.3 \\
  Video Platform & 4 & 0 & - & - & 2 & 1/1/1.0 & 1/3/2.0 & 2 & 2/2/2.0 & 4/4/4.0 \\
  Railway System & 5 & 1 & 1/1/1.0 & 2/2/2.0 & 0 & - & - & 4 & 2/2/2.0 & 3/34/11.8 \\
  Academic Research & 16 & 4 & 1/1/1.0 & 3/5/4.3 & 3 & 1/1/1.0 & 2/3/2.3 & 9 & 2/3/2.1 & 3/9/6.3 \\
  Food and Catering & 4 & 1 & 1/1/1.0 & 4/4/4.0 & 1 & 1/1/1.0 & 3/3/3.0 & 2 & 2/3/2.5 & 5/6/5.5 \\
  \midrule
  \textbf{ALL Data} & \textbf{89} & \textbf{22} & 1/1/1.0 & 2/22/8.6 & \textbf{23} & 1/2/1.0 & 1/4/2.3 & \textbf{44} & 2/3/2.2 & 2/34/7.1 \\
  \bottomrule
  \end{tabular}%
  }
  \endgroup
\end{table}

%% file: common/tables/hyperparameters.tex
\begin{table}[h]
    \centering
    \caption{Inference configuration and official API pricing for each model. Prices are in USD per 1M tokens. For tiered models, we report \texttt{low}/\texttt{high} prices, where tiers are triggered when input tokens exceed 272K for GPT-5.6-Sol and 512K for MiniMax-M3.}
    \label{tab:hyperparameters}
    \resizebox{\textwidth}{!}{%
    \begin{tabular}{@{}cccccc@{}}
    \toprule
    \textbf{Model} & \textbf{Reasoning Effort} & \textbf{Temperature (Top-p)} & \textbf{Input (\$)} & \textbf{Cached Input (\$)} & \textbf{Output (\$)} \\ \midrule
    GPT-5.6-Sol & medium & - & 4 / 8 & 0.4 / 0.8 & 20 / 30 \\
    Claude-Sonnet-5 & adaptive & - & 2 & 0.2 & 10 \\
    Qwen-3.8-Max & enable & 1.0 (1.0) & 1.79 & 0.22 & 5.37 \\
    Kimi-K3 & max & 1.0 (0.95) & 2.98 & 0.30 & 14.91 \\
    GLM-5.3 & max & 1.0 (0.95) & 1.4 & 0.26 & 4.4 \\
    MiniMax-M3 & adaptive & 1.0 (0.95) & 0.31 / 0.63 & 0.06 / 0.13 & 1.25 / 2.5 \\
    GLM-5.3-Flash & max & 1.0 (0.95) & 0.15 & 0.03 & 0.5 \\
    DeepSeek-V4-Flash-0731 & high & 1.0 (0.95) & 0.45 & 0.02 & 1.34 \\ \bottomrule
    \end{tabular}%
    }
\end{table}

%% file: common/tables/modes_outline.tex
\begin{table}[h]
  \centering
  \caption{The outline of common MAD-SAE variants.
  \textcolor{readonlyorange}{Orange} denotes read-only sub-agents;
  \textcolor{writableblue}{blue} denotes writable sub-agents.}
  \label{tab:mad-sae-variants}
  \resizebox{\textwidth}{!}{%
  \footnotesize
  \begin{tabular}{@{}lp{0.40\textwidth}
    >{\raggedright\arraybackslash}p{0.24\textwidth}
    >{\raggedright\arraybackslash}p{0.24\textwidth}@{}}
    \toprule
    \multicolumn{1}{@{}c}{\textbf{Mode}} &
    \multicolumn{1}{c}{\textbf{Description}} &
    \multicolumn{1}{c}{\textbf{Sub-Agent Types}} &
    \multicolumn{1}{c@{}}{\textbf{Relevant Harness}} \\
    \midrule
    Single-Agent &
    The main-agent resolves the task independently without delegation. &
    -- &
    Kimi-Code, Codex, Claude Code, Cursor, TRAE \\
    General (Gen) &
    The main-agent autonomously decides when and how to delegate to
    generally defined sub-agents. &
    \ro{Planner}, \ro{Explorer}, \rw{Coder} &
    Kimi-Code, Claude Code, TRAE \\
    Specialized (Spec) &
    The main-agent autonomously decides when and how to delegate to
    domain-specialized sub-agents. &
    \ro{Planner}, \ro{ALR}, \ro{EDA},
    \rw{EWO},
    \ro{FMA},
    \rw{ORB},
    \ro{WDR} &
    Cursor \\
    Plan-1st &
    The main-agent is forced to call the planner for task decomposition and
    delegation planning at first, then follows the plan. &
    \ro{Planner}, \ro{Explorer}, \rw{Coder} &
    Kimi-Code, Codex, Claude Code, Cursor, TRAE \\
    Oracle &
    The main-agent views annotated delegation priors and is required to
    follow them. &
    \ro{Planner}, \ro{Explorer}, \rw{Coder} &
    Kimi-Code, Codex, Claude Code, Cursor, TRAE \\
    \bottomrule
  \end{tabular}%
  }
\end{table}

%% file: common/tables/case_taxonomy.tex
\begin{table}[h]
\centering
\caption{The taxonomy of MAD-SAE (Spec) cases. Across both partitions, $\checkmark$ denotes strictly fewer critical steps (Efficiency) or strictly lower cost (Cost), and $\times$ denotes the absence of that gain (equality or an increase); $=$ denotes equality. CIS additionally requires strictly higher cost, and CIF requires a strict increase in critical steps or cost.}
\label{tab:case_taxonomy}
\resizebox{\textwidth}{!}{%
\begin{tabular}{@{}ccccc@{}}
\toprule
Partition                               & Category                                                  & Success Situation                            & Efficiency Benefits & Cost Benefits \\ \midrule
\multirow{7}{*}{\shortstack{Beneficial\\Cases (625)}} & \multirow{4}{*}{Failure-to-Success (F2S, 130)}            & \multirow{4}{*}{\shortstack{MAD-SAE success,\\SA failure}} & $\checkmark$            & $\checkmark$  \\
                                        &                                                           &                                              & $\checkmark$            & $\times$      \\
                                        &                                                           &                                              & $\times$                & $\checkmark$  \\
                                        &                                                           &                                              & $\times$                & $\times$      \\ \cmidrule(lr){2-5}
                                        & \multirow{3}{*}{Cheaper-or-Efficient Success (CES, 495)} & \multirow{3}{*}{\shortstack{MAD-SAE success,\\SA success}} & $\checkmark$ & $\times$      \\
                                        &                                                           &                                              & $\checkmark$            & $\checkmark$  \\
                                        &                                                           &                                              & $\times$                & $\checkmark$  \\ \midrule
\multirow{10}{*}{\shortstack{Adverse\\Cases (703)}}    & \multirow{4}{*}{Success-to-Failure (S2F, 124)}            & \multirow{4}{*}{\shortstack{MAD-SAE failure,\\SA success}} & $\checkmark$            & $\checkmark$  \\
                                        &                                                           &                                              & $\checkmark$            & $\times$      \\
                                        &                                                           &                                              & $\times$                & $\checkmark$  \\
                                        &                                                           &                                              & $\times$                & $\times$      \\ \cmidrule(lr){2-5}
                                        &                                                           & \multirow{2}{*}{\shortstack{MAD-SAE success,\\SA success}} & \multirow{2}{*}{$\times$} & \multirow{2}{*}{$\times$} \\
                                        & \multirow{-2}{*}{Costly-and-Inefficient Success (CIS, 349)} &                                              &                           &                           \\ \cmidrule(lr){2-5}
                                        & Attempt-for-Efficiency-or-Saving (AES, 88)                & \multirow{4}{*}{\shortstack{MAD-SAE failure,\\SA failure}} & $\checkmark$ or $=$ & $\checkmark$  \\ \cmidrule(lr){2-2}\cmidrule(lr){4-5}
                                        & \multirow{3}{*}{Costly-or-Inefficient Failure (CIF, 142)} &                         & $\checkmark$            & $\times$      \\
                                        &                                                           &                                              & $\times$                & $\checkmark$  \\
                                        &                                                           &                                              & $\times$                & $\times$      \\ \bottomrule
\end{tabular}%
}
\end{table}

%% file: common/tables/benefits_and_drawbacks_per_model.tex
\begin{table}[H]
    \centering
    \small
    \caption{Per-model case composition and delegation-quality metrics under MAD-SAE (Spec); UDR, ODR, and $U_p$/$U_e$/$U_c$ are reproduced from Table~\ref{tab:variants_models_comparison}. $B/A$: beneficial/adverse case counts, with the larger side of each pair bolded (ties unbolded). $\Delta\mathrm{SR}$: success-rate change on $\mathcal{D}$ from converting SA to MAD-SAE (Spec). Bold/underline mark best/second-best for UDR and ODR.}
    \label{tab:per_model_case_attribution}
    \resizebox{\textwidth}{!}{%
    \begin{tabular}{@{}l|cccc|c|cc|ccc@{}}
    \toprule
    \multirow{2}{*}{Model} & \multicolumn{4}{c|}{Case Composition (full $\mathcal{D}$)} & \multirow{2}{*}{$\Delta\mathrm{SR}$} & \multicolumn{2}{c|}{Delegation Appropriateness} & \multicolumn{3}{c}{Orchestration Utilities} \\
    \cmidrule(lr){2-5} \cmidrule(lr){7-8} \cmidrule(lr){9-11}
     & $B/A$ & MAO ($B/A$) & Non-MAO ($B/A$) & F2S/S2F & & UDR$\downarrow$ & ODR$\downarrow$ & $U_p\uparrow$ & $U_e\uparrow$ & $U_c\uparrow$ \\
    \midrule
    GLM-5.3                 & \textbf{89}/77  & \textbf{25}/21 & \textbf{64}/56  & \textbf{16}/9  & $+4.2$  & \underline{8.9\%}    & 64.9\%             & \textbf{50.8\%} & \textbf{52.7\%} & 40.5\%          \\
    GLM-5.3-Flash           & \textbf{85}/81  & 32/\textbf{41} & \textbf{53}/40  & \textbf{7}/5   & $+1.2$  & 26.9\%               & 50.6\%             & \textbf{51.3\%} & \textbf{54.0\%} & 32.0\%          \\
    Kimi-K3                 & \textbf{130}/36 & \textbf{89}/22 & \textbf{41}/14  & \textbf{26}/9  & $+10.2$ & 50.5\%               & \underline{16.8\%} & \textbf{56.4\%} & \textbf{50.3\%} & \textbf{50.9\%} \\
    \midrule
    Qwen-3.8-Max             & 80/\textbf{86}  & \textbf{32}/22 & 48/\textbf{64}  & \textbf{20}/3  & $+10.2$ & 15.7\%               & 59.7\%             & \textbf{54.4\%} & 49.9\%          & 38.4\%          \\
    DeepSeek-V4-Flash-0731   & 75/\textbf{91}  & \textbf{26}/19 & 49/\textbf{72}  & 11/11          & $0.0$   & 16.8\%               & 57.1\%             & \textbf{50.4\%} & 48.0\%          & 33.1\%          \\
    Claude-Sonnet-5          & 74/\textbf{92}  & \textbf{46}/43 & 28/\textbf{49}  & 15/\textbf{22} & $-4.2$  & 39.3\%               & 29.8\%             & 45.9\%          & 46.8\%          & 34.1\%          \\
    GPT-5.6-Sol              & 29/\textbf{137} & 0/\textbf{7}   & 29/\textbf{130} & 9/\textbf{45}  & $-21.7$ & \textbf{7.8\%}       & 93.5\%             & 42.6\%          & 44.0\%          & 27.4\%          \\
    MiniMax-M3               & 63/\textbf{103} & 55/\textbf{79} & 8/\textbf{24}   & \textbf{26}/20 & $+3.6$  & 79.7\%               & \textbf{9.0\%}     & 42.8\%          & 42.7\%          & 37.6\%          \\
    \bottomrule
    \end{tabular}%
    }
\end{table}


%% file: common/figures_tex/model_wise_case_study/glm_5_3.tex
\begin{figure}[H]
    \centering
    \includegraphics[width=0.7\textwidth]{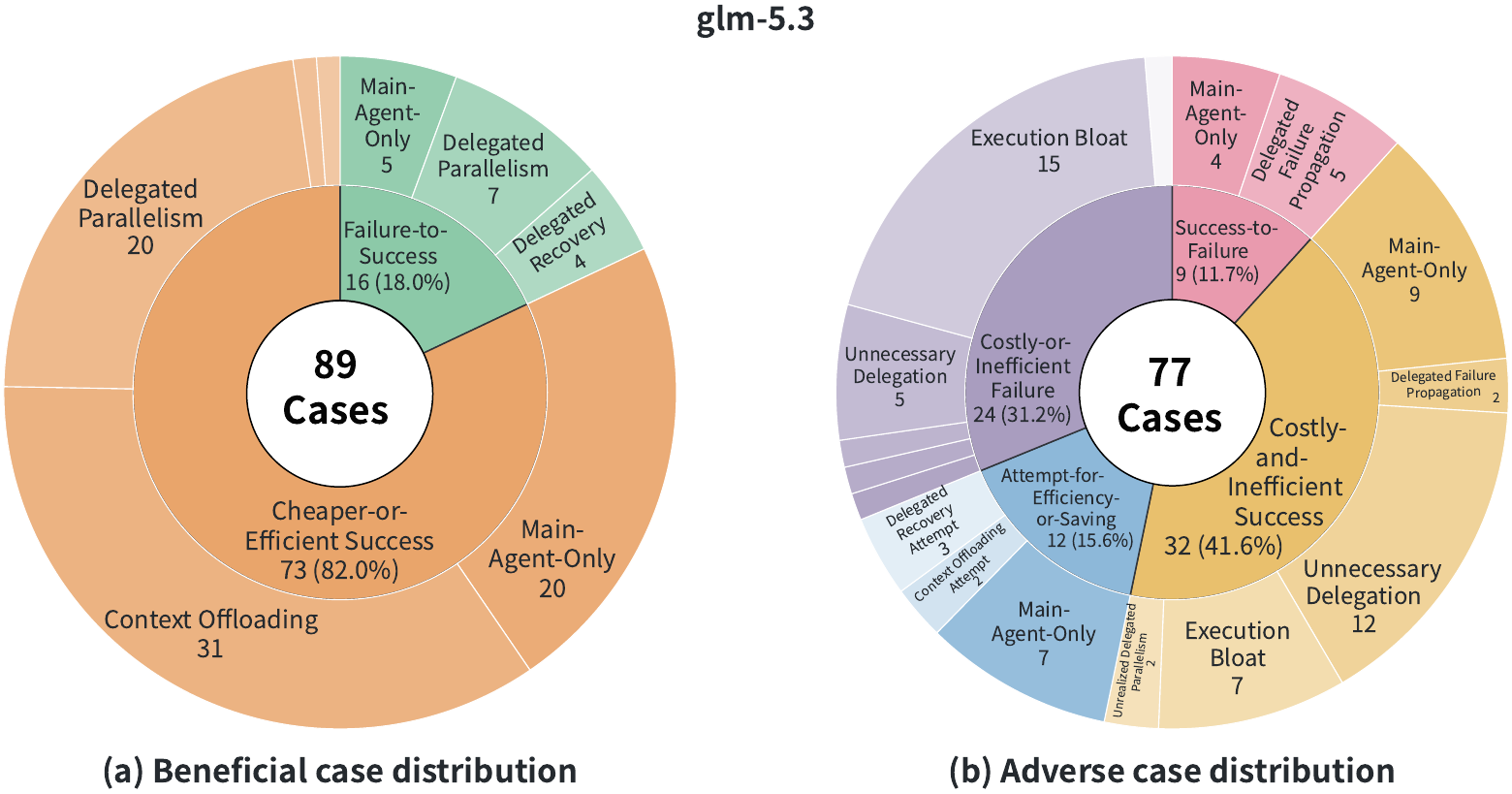}
    \caption{The distribution of beneficial and adverse categories, alongside with their underlying reasons of GLM-5.3.}
    \label{fig:glm_5_3_case_study}
\end{figure}

%% file: common/figures_tex/model_wise_case_study/glm_5_3_flash.tex
\begin{figure}[H]
    \centering
    \includegraphics[width=0.7\textwidth]{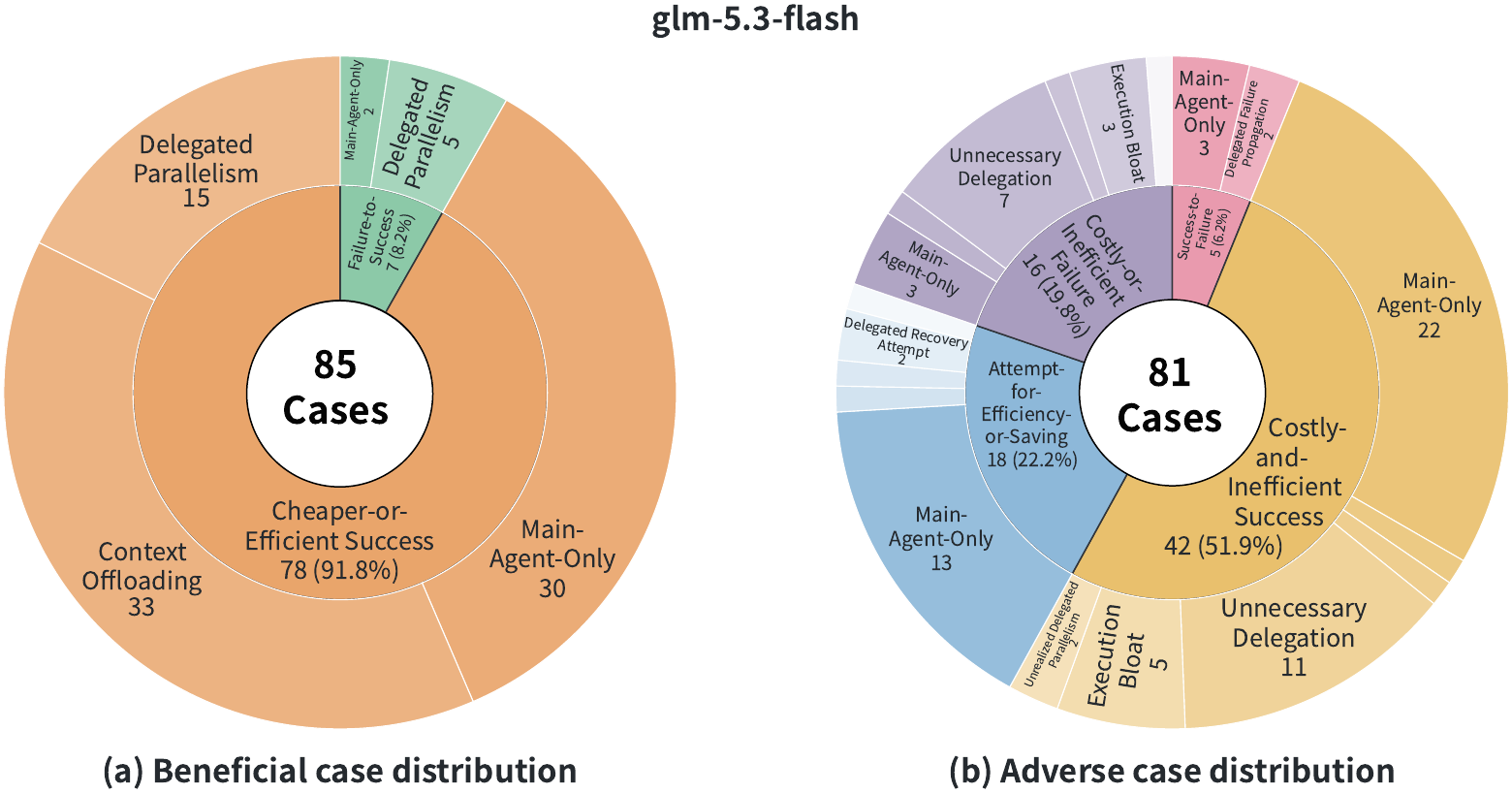}
    \caption{The distribution of beneficial and adverse categories, alongside with their underlying reasons of GLM-5.3-Flash.}
    \label{fig:glm_5_3_flash_case_study}
\end{figure}
\vspace{-0.5cm}

%% file: common/figures_tex/model_wise_case_study/kimi_k3.tex
\begin{figure}[H]
    \centering
    \includegraphics[width=0.7\textwidth]{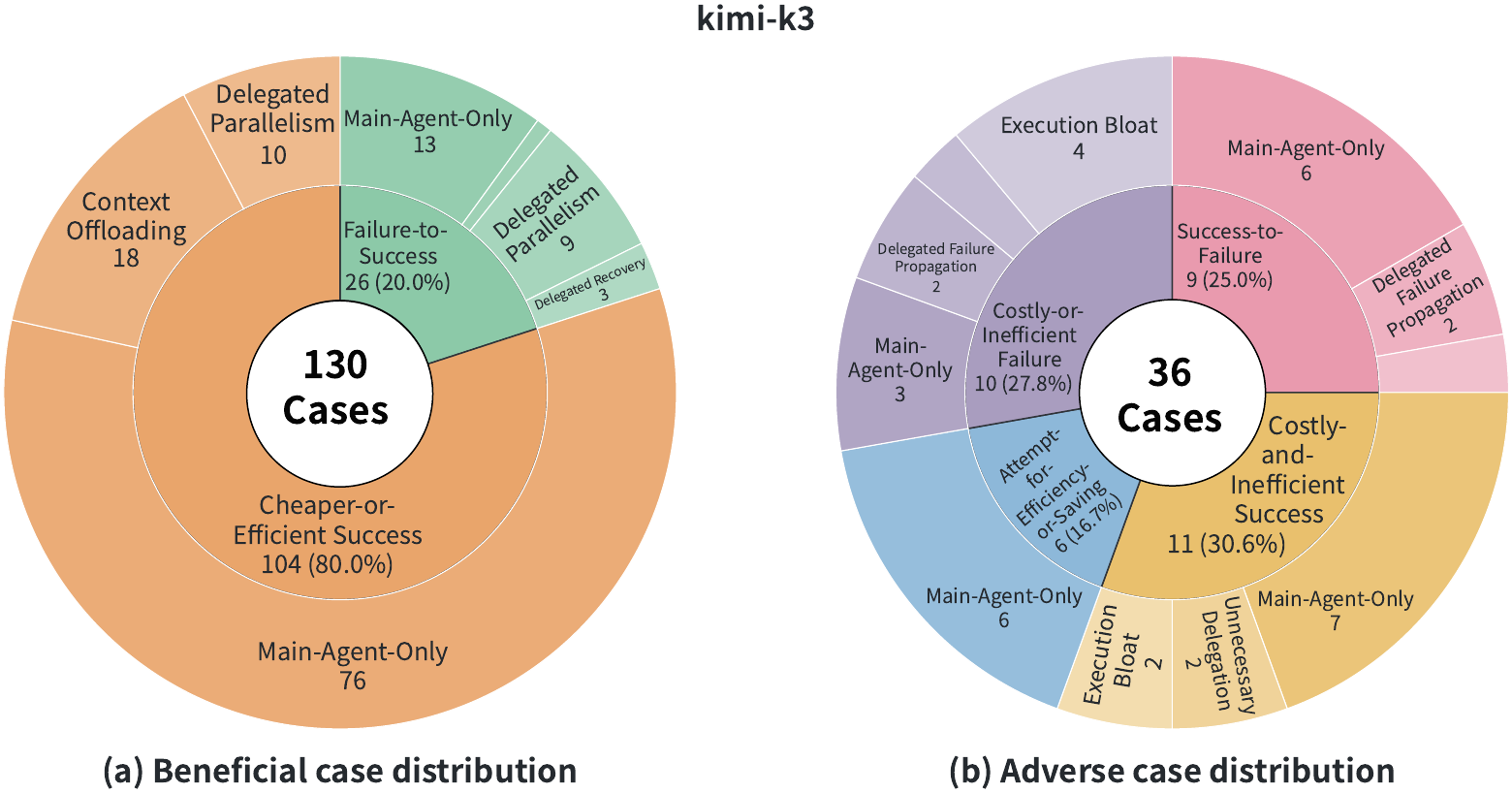}
    \caption{The distribution of beneficial and adverse categories, alongside with their underlying reasons of Kimi-K3.}
    \label{fig:kimi_k3_case_study}
\end{figure}
\vspace{-0.5cm}

%% file: common/figures_tex/model_wise_case_study/qwen_3_8_max.tex
\begin{figure}[H]
    \centering
    \includegraphics[width=0.7\textwidth]{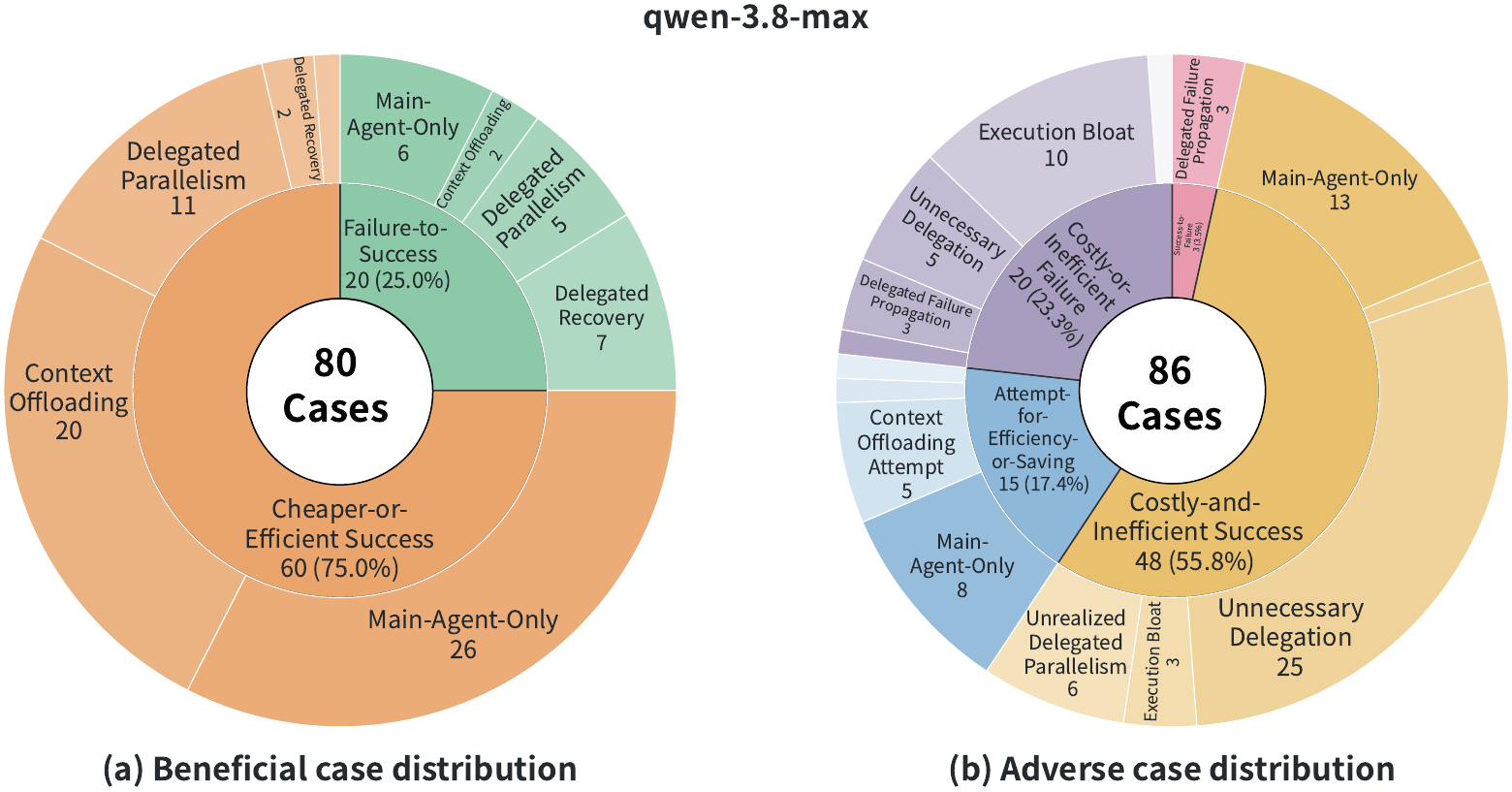}
    \caption{The distribution of beneficial and adverse categories, alongside with their underlying reasons of Qwen-3.8-Max.}
    \label{fig:qwen_3_8_max_case_study}
\end{figure}
\vspace{-0.5cm}

%% file: common/figures_tex/model_wise_case_study/deepseek_v4_flash_0731.tex
\begin{figure}[H]
    \centering
    \includegraphics[width=0.7\textwidth]{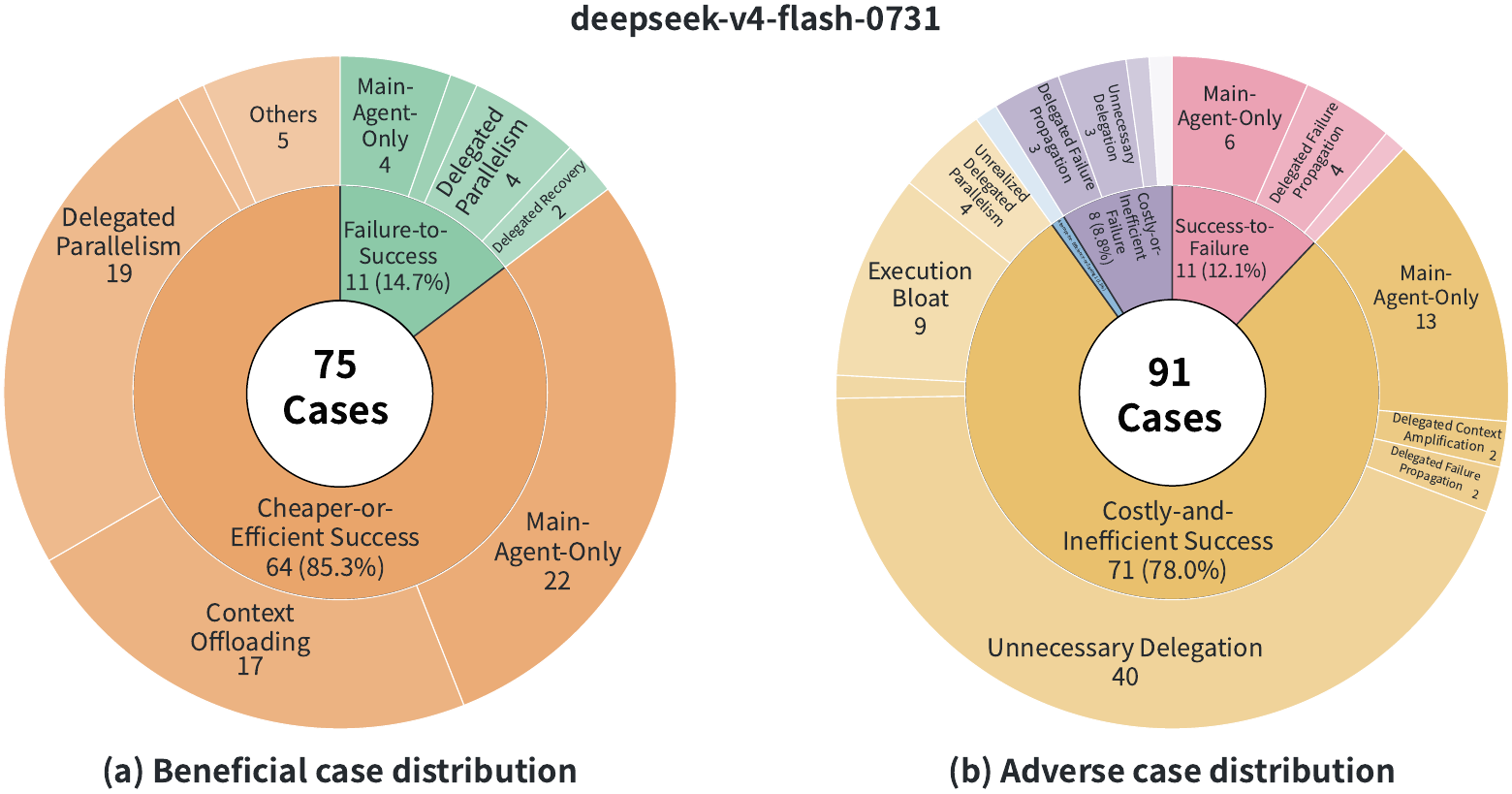}
    \caption{The distribution of beneficial and adverse categories, alongside with their underlying reasons of DeepSeek-V4-Flash-0731.}
    \label{fig:dsv4_flash_0731_case_study}
\end{figure}
\vspace{-0.5cm}

%% file: common/figures_tex/model_wise_case_study/claude_sonnet_5.tex
\begin{figure}[H]
    \centering
    \includegraphics[width=0.7\textwidth]{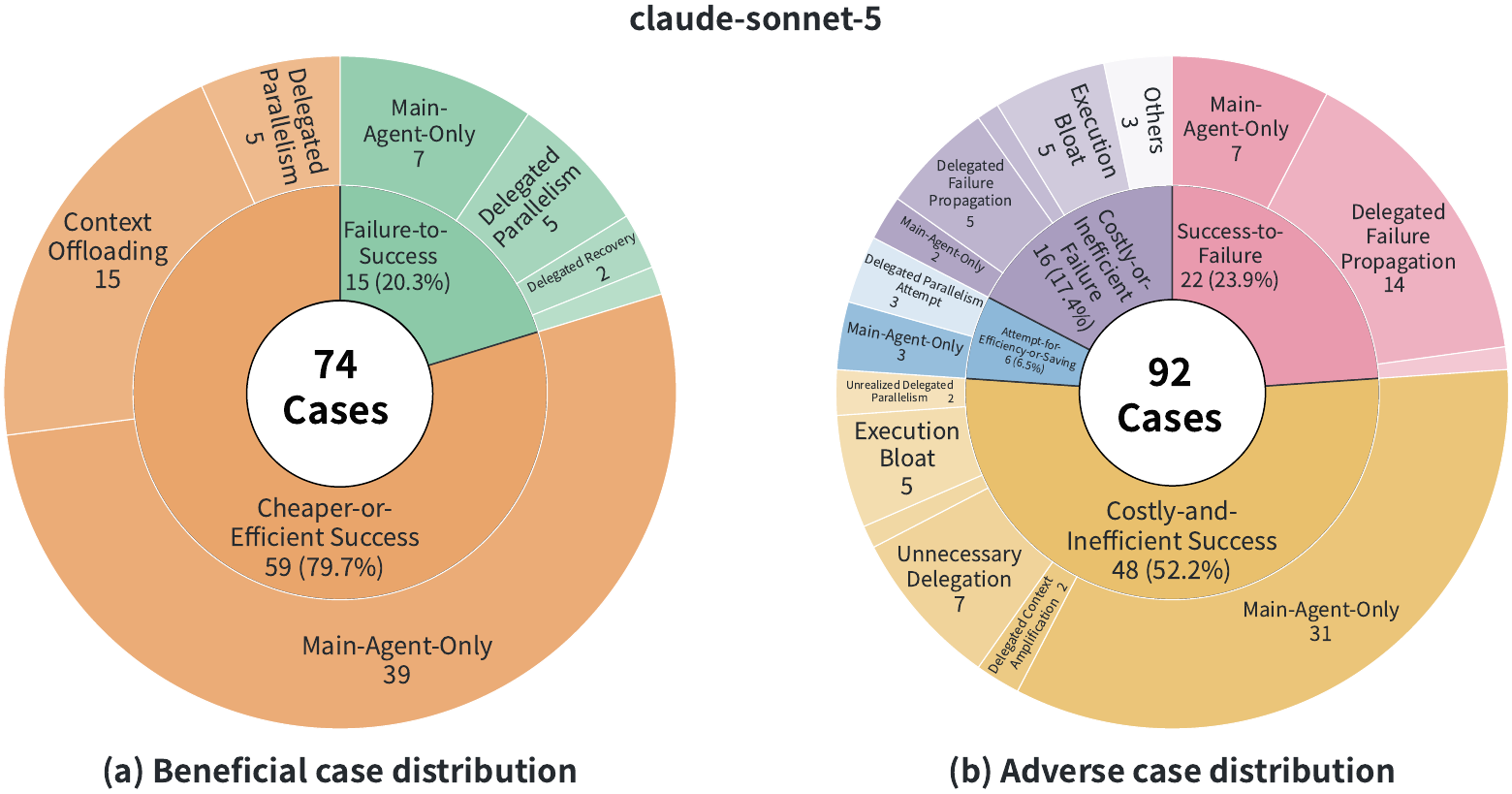}
    \caption{The distribution of beneficial and adverse categories, alongside with their underlying reasons of Claude-Sonnet-5.}
    \label{fig:claude_sonnet_5_case_study}
\end{figure}
\vspace{-0.5cm}

%% file: common/figures_tex/model_wise_case_study/gpt_5_6_sol.tex
\begin{figure}[H]
    \centering
    \includegraphics[width=0.7\textwidth]{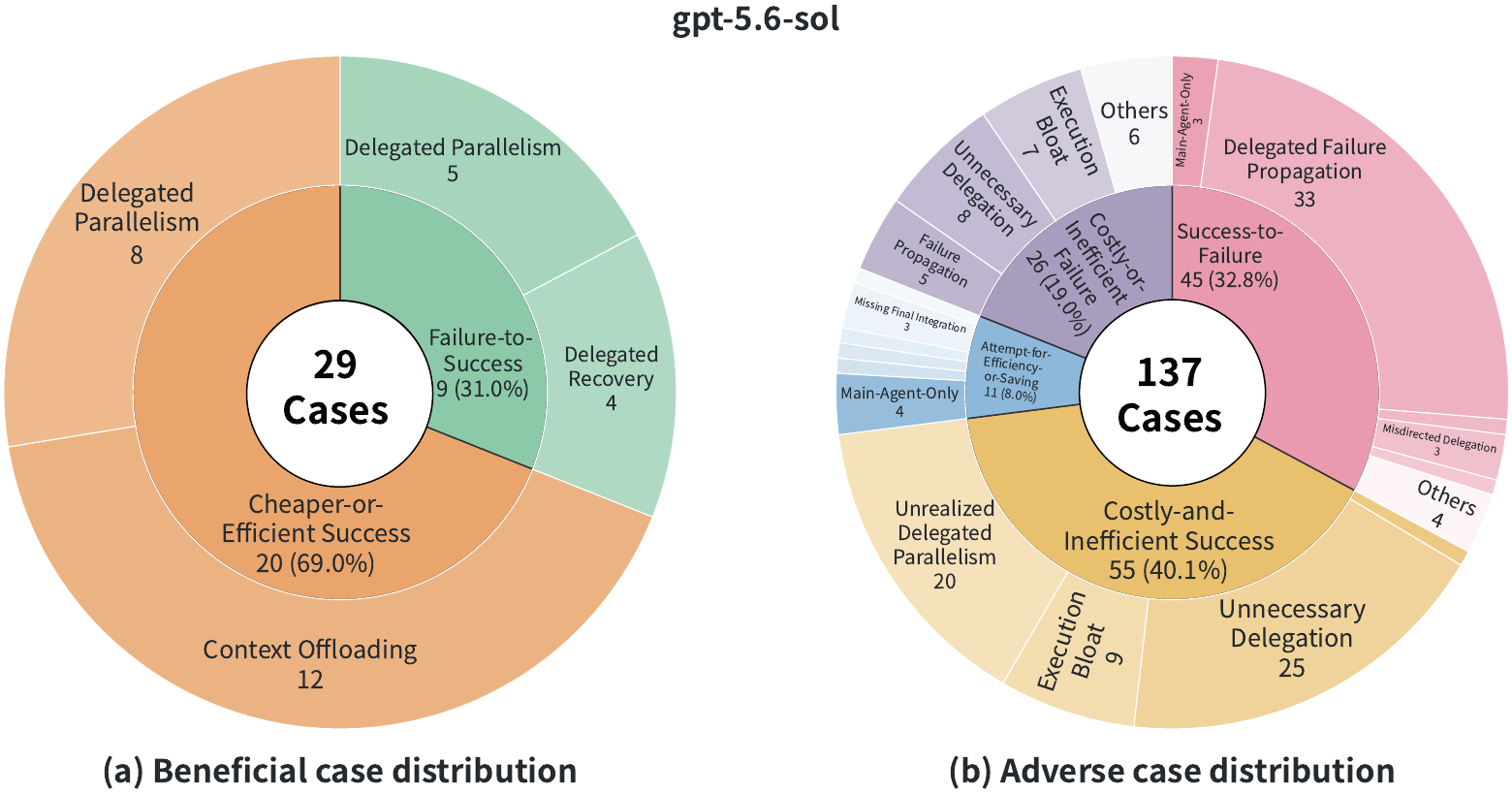}
    \caption{The distribution of beneficial and adverse categories, alongside with their underlying reasons of GPT-5.6-Sol.}
    \label{fig:gpt_5_6_sol_case_study}
\end{figure}
\vspace{-0.5cm}

%% file: common/figures_tex/model_wise_case_study/minimax_m3.tex
\begin{figure}[H]
    \centering
    \includegraphics[width=0.7\textwidth]{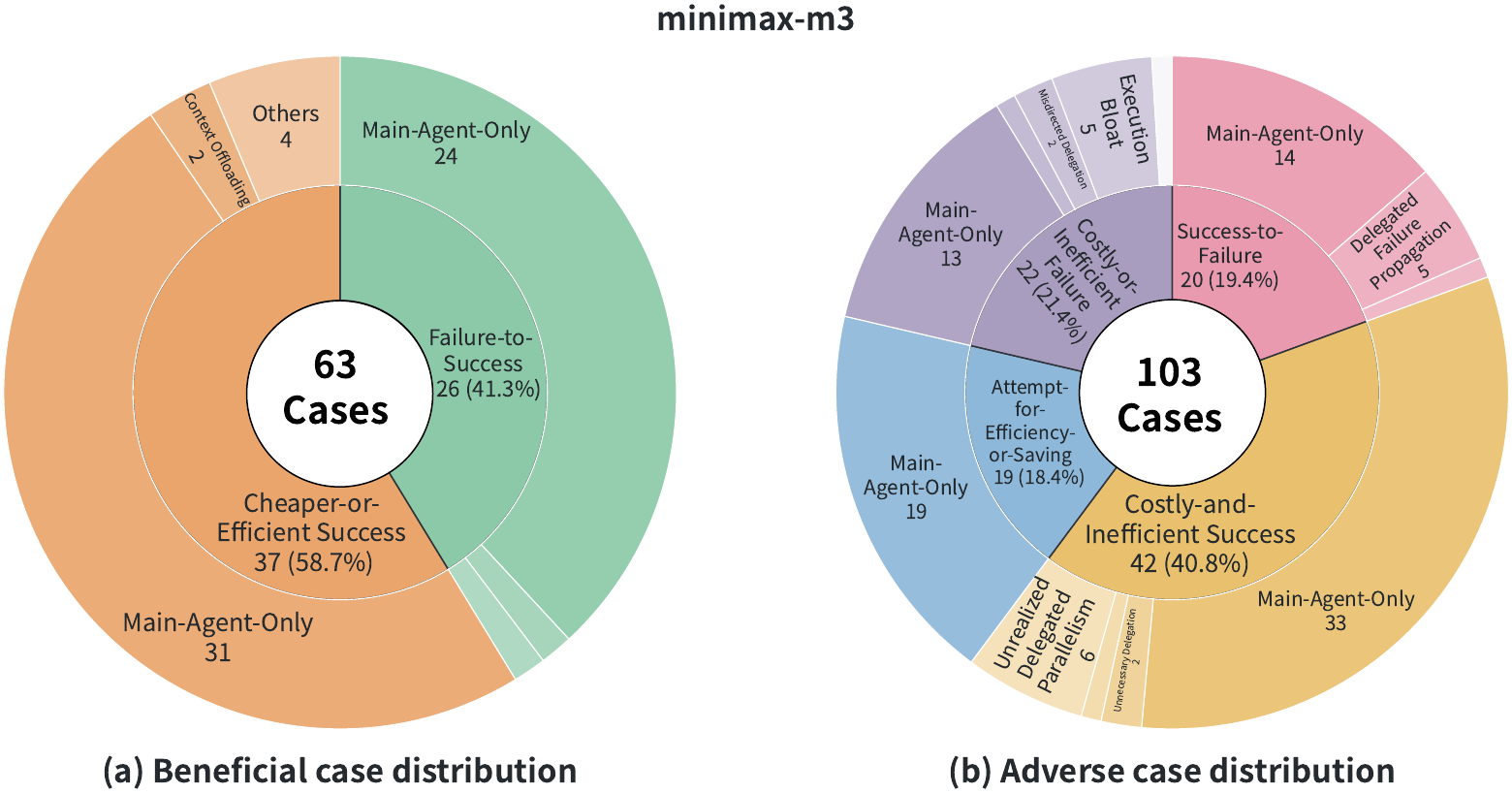}
    \caption{The distribution of beneficial and adverse categories, alongside with their underlying reasons of MiniMax-M3.}
    \label{fig:minimax_m3_case_study}
\end{figure}
\vspace{-0.5cm}

%% file: common/figures_tex/case_study_classic_cases/context_offload.tex
\begin{figure}[H]
    \centering
    \includegraphics[width=1.0\textwidth]{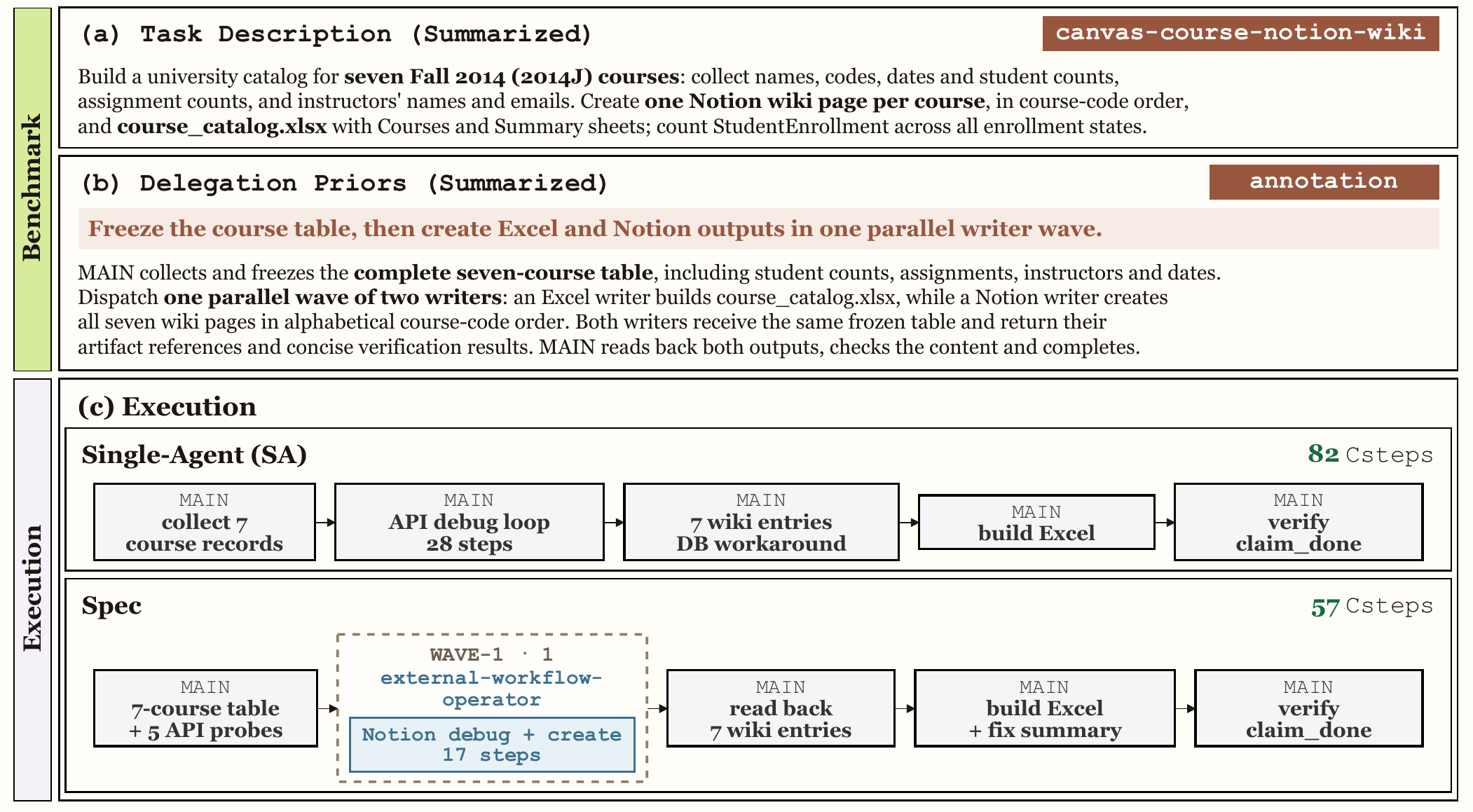}
    \caption{The Illustration of the Context Offloading benefit on GLM-5.3.}
    \label{fig:context_offload_glm5_3}
\end{figure}
\vspace{-0.5cm}

%% file: common/figures_tex/case_study_classic_cases/delegated_parallelism.tex
\begin{figure}[H]
    \centering
    \includegraphics[width=1.0\textwidth]{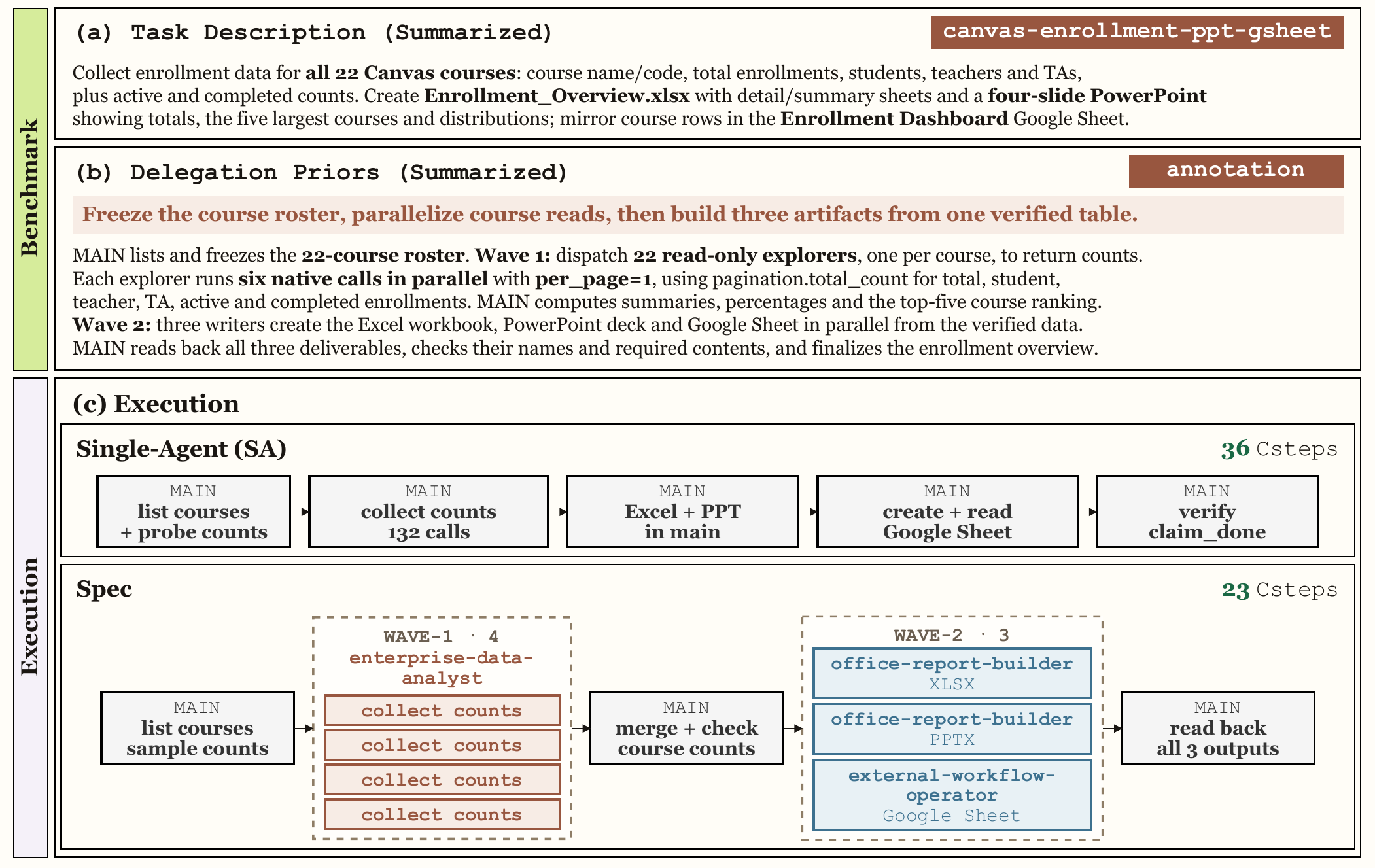}
    \caption{The Illustration of the Delegated Parallelism benefit on GLM-5.3.}
    \label{fig:delegated_parallelism_glm5_3}
\end{figure}
\vspace{-0.5cm}

%% file: common/figures_tex/case_study_classic_cases/delegated_parallelism_k3.tex
\begin{figure}[H]
    \centering
    \includegraphics[width=1.0\textwidth]{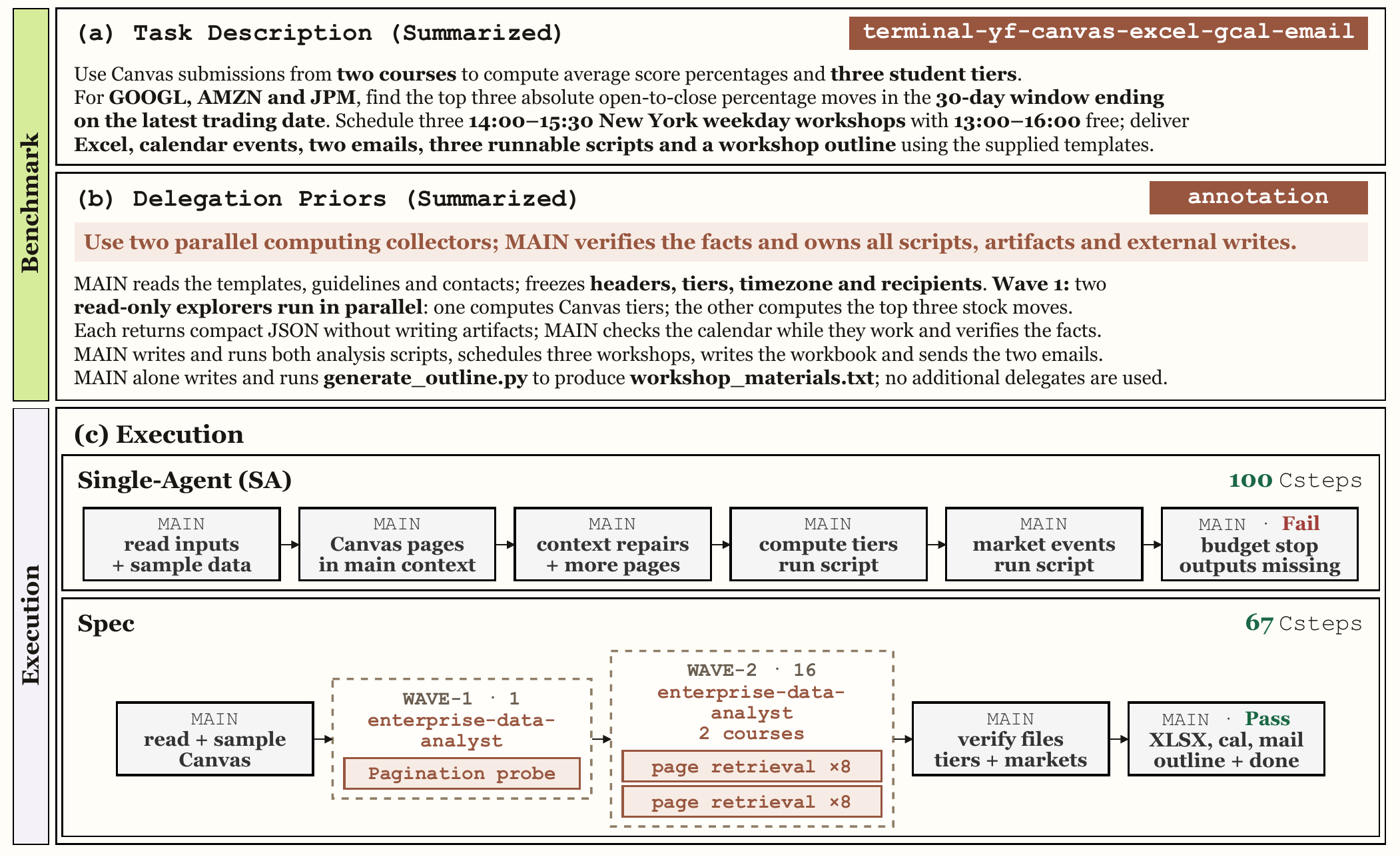}
    \caption{The Illustration of the Delegated Parallelism benefit on Kimi-K3.}
    \label{fig:delegated_parallelism_k3}
\end{figure}
\vspace{-0.5cm}

%% file: common/figures_tex/case_study_classic_cases/better_main_agent_only.tex
\begin{figure}[H]
    \centering
    \includegraphics[width=1.0\textwidth]{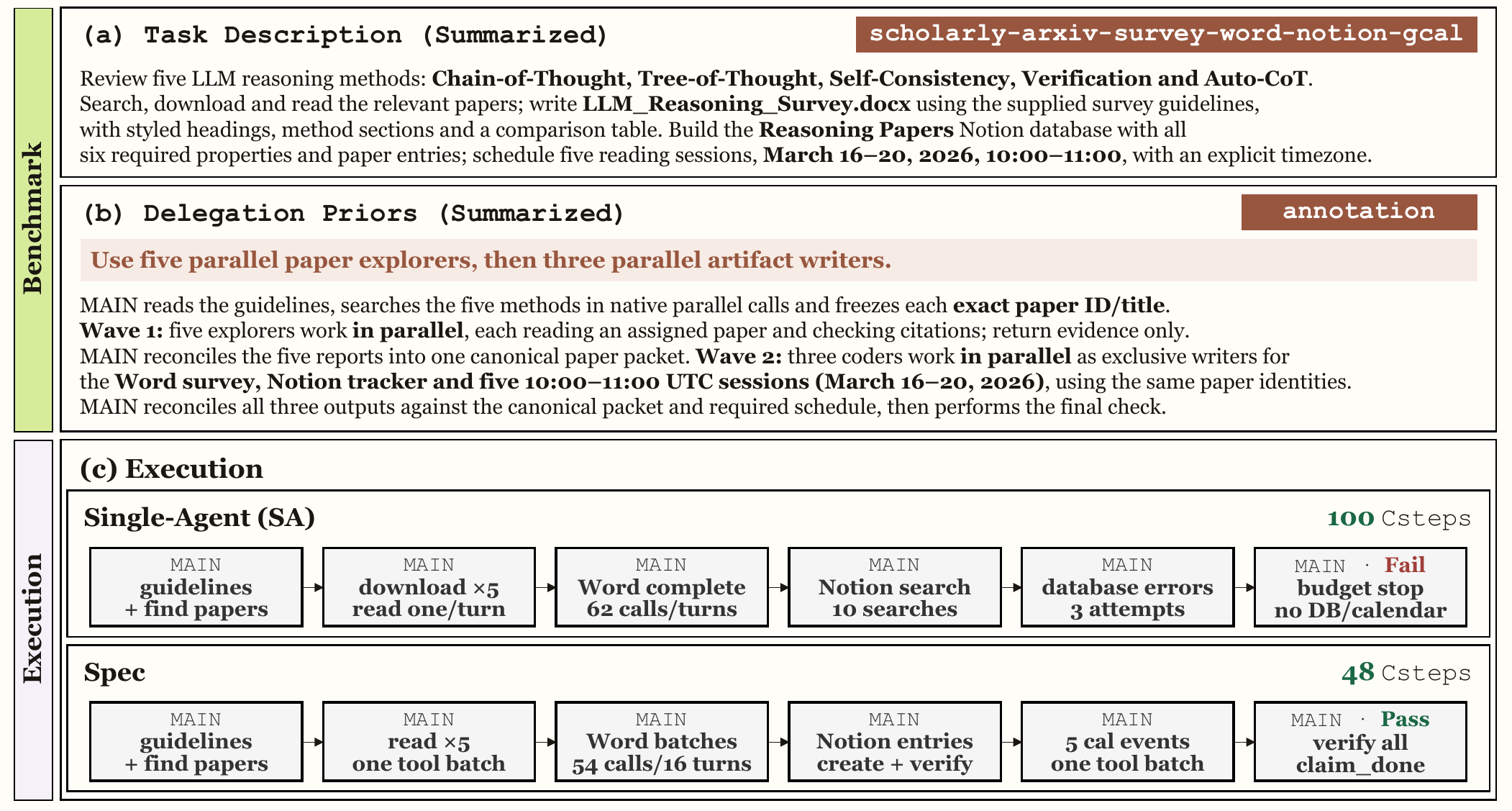}
    \caption{The Illustration of the Better Main-Agent-Only solution on Minimax-M3.}
    \label{fig:better_mao_mm3}
\end{figure}
\vspace{-0.5cm}

%% file: common/figures_tex/case_study_classic_cases/unnessary_delegation.tex
\begin{figure}[H]
    \centering
    \includegraphics[width=1.0\textwidth]{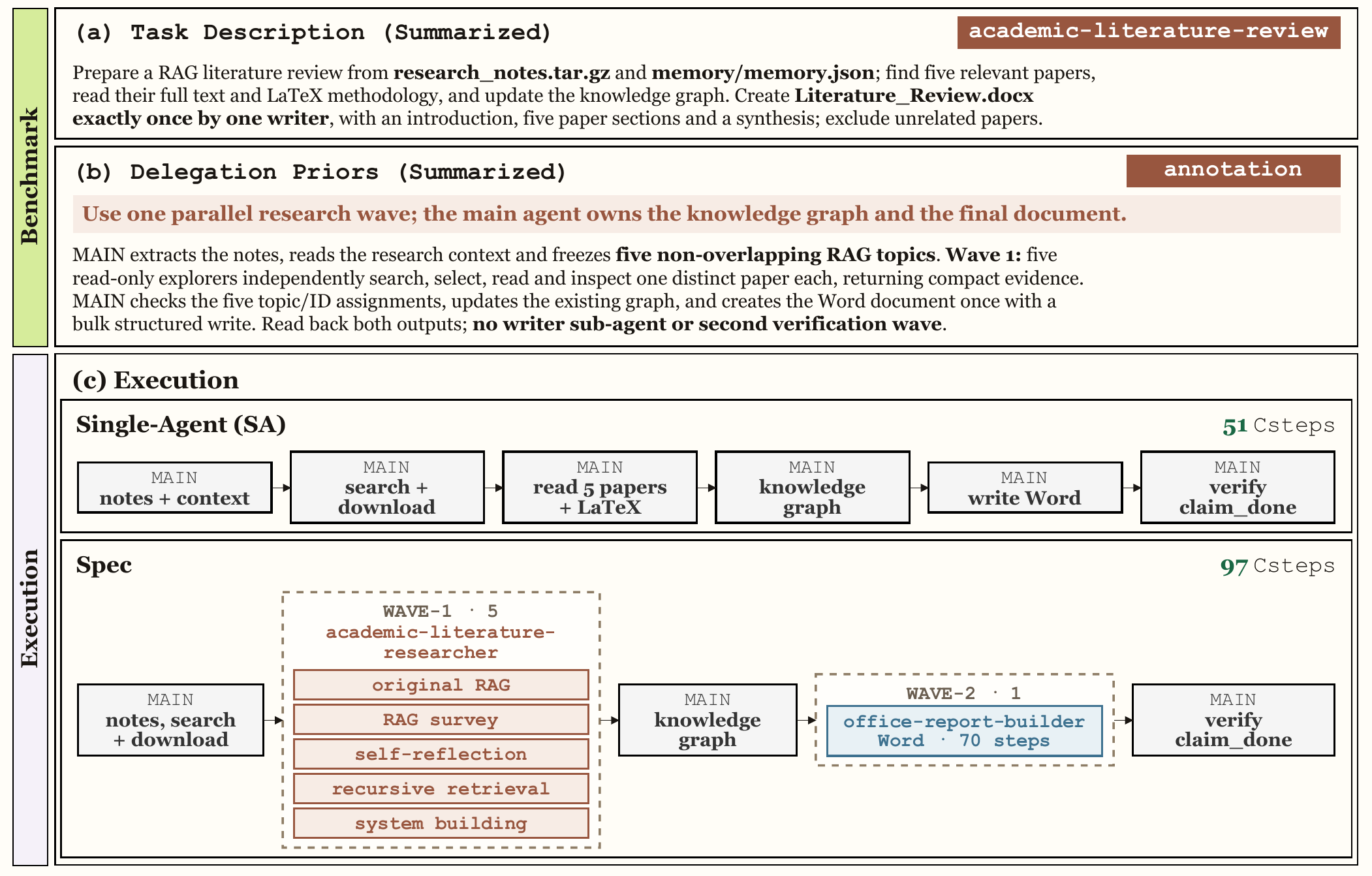}
    \caption{The Illustration of the Unnecessary Delegation drawback on DeepSeek-V4-Flash-0731.}
    \label{fig:unnessary_delegation_dsv4}
\end{figure}
\vspace{-0.5cm}

%% file: common/figures_tex/case_study_classic_cases/delegated_failure_propa.tex
\begin{figure}[H]
    \centering
    \includegraphics[width=1.0\textwidth]{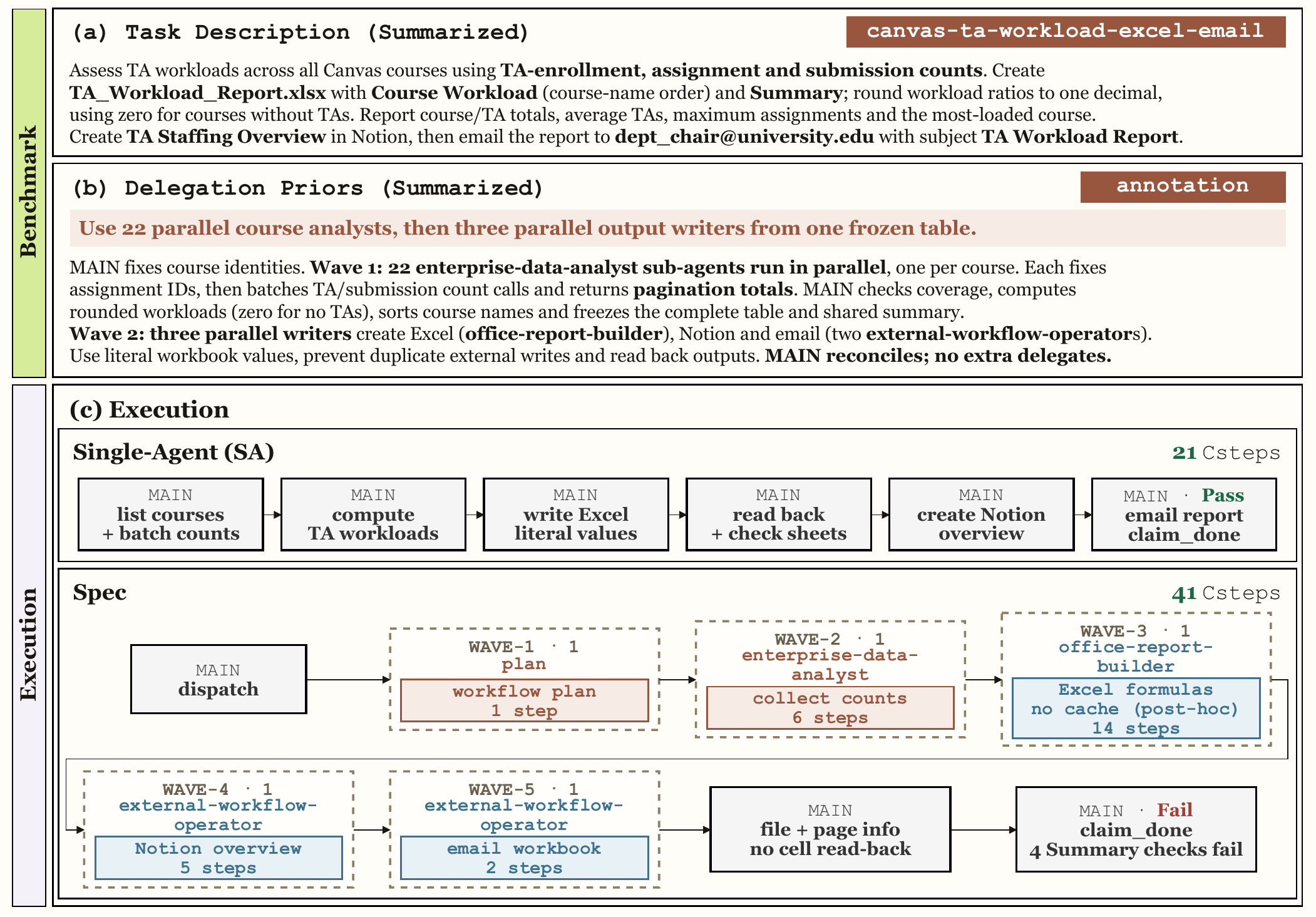}
    \caption{The Illustration of the Delegated Falure Propagation drawback on GPT-5.6-Sol.}
    \label{fig:delegated_failure_propa_gpt_5_6}
\end{figure}
\vspace{-0.5cm}

%% file: common/figures_tex/case_study_classic_cases/udp.tex
\begin{figure}[H]
    \centering
    \includegraphics[width=1.0\textwidth]{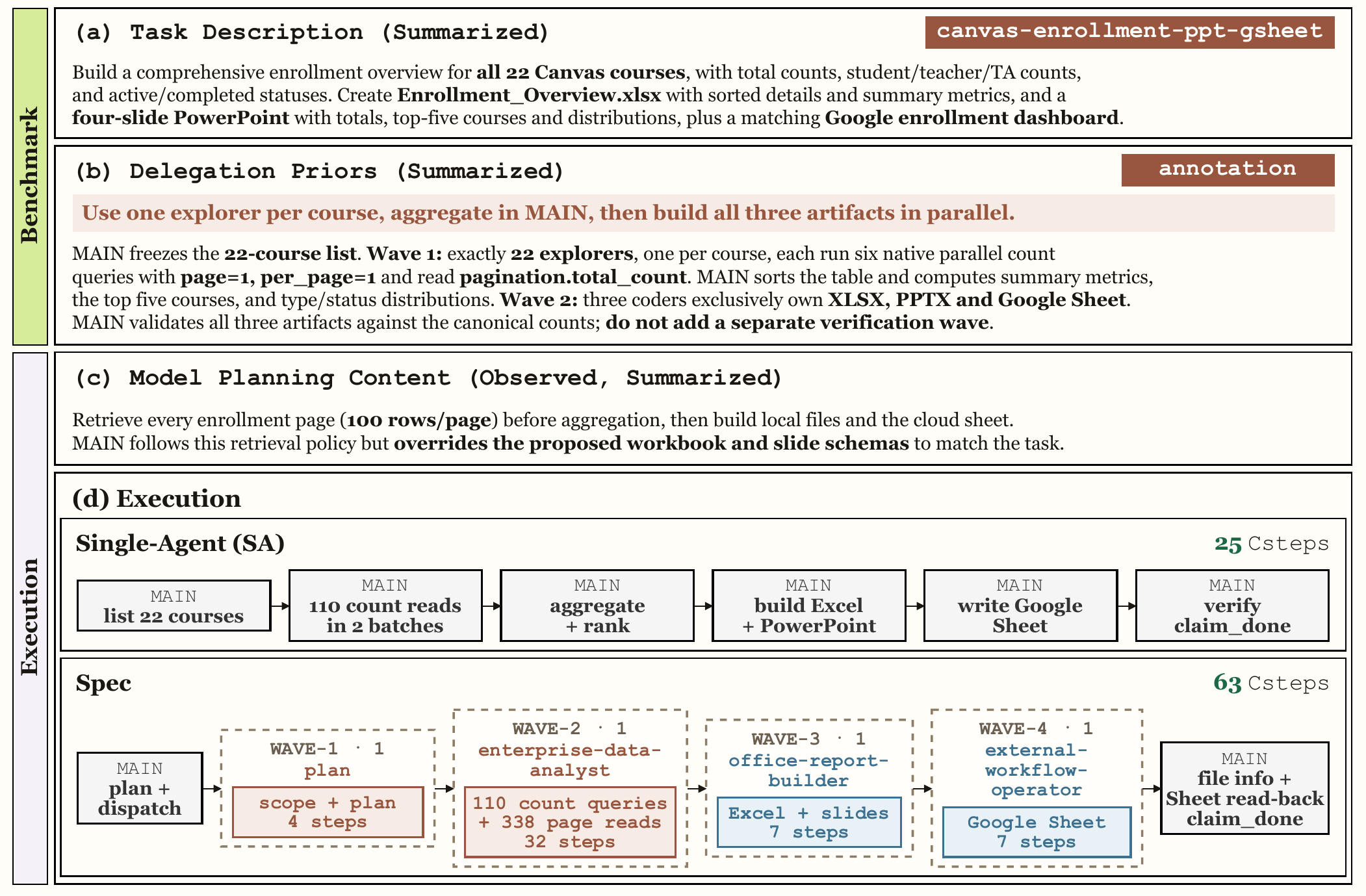}
    \caption{The Illustration of the Unrealized Delegated Parallelism drawback on GPT-5.6-Sol.}
    \label{fig:udp_gpt_5_6}
\end{figure}
\vspace{-0.5cm}

%% file: common/figures_tex/case_study_classic_cases/execution_bloat.tex
\begin{figure}[H]
    \centering
    \includegraphics[width=1.0\textwidth]{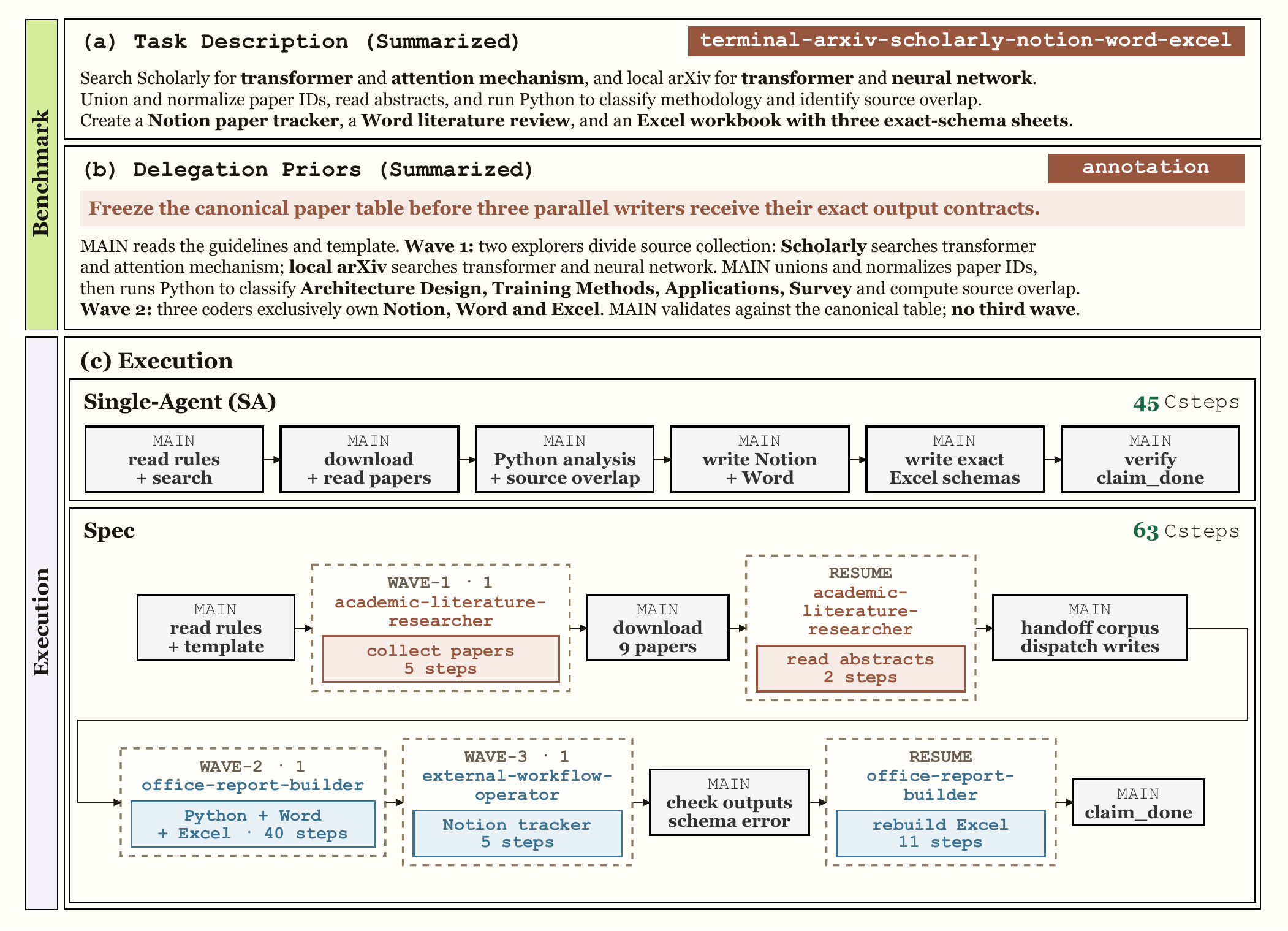}
    \caption{The Illustration of the Execution Bloat drawback on GPT-5.6-Sol.}
    \label{fig:execution_bloat_gpt_5_6}
\end{figure}
\vspace{-0.5cm}

%% file: common/appendix_prompt_setup.tex
\begingroup
\raggedbottom
\definecolor{PromptOrange}{HTML}{D97724}
\newcounter{promptbox}[section]
\renewcommand{\thepromptbox}{\thesection.\arabic{promptbox}}

%% file: common/tables/principle_tools/arxiv_full_task_table.tex
\begin{table}[H]
\centering
\scriptsize
\begin{tabular}{@{}p{\linewidth}@{}}
\toprule
\textbf{Task Description} \\
\midrule
\setlength{\parskip}{0.8em}
I am a PhD student preparing a survey presentation on recent advances in natural language processing, specifically focusing on large language models and their reasoning capabilities. I need to gather information from multiple sources, organize it systematically, and create both a data spreadsheet and a presentation.

First, visit the survey methodology guide at \url{http://localhost:30231} to learn about how to structure a literature survey. The main page explains the overall approach including comparison criteria and taxonomy development. There is also a guide page at \url{http://localhost:30231/guide.html} with tips on presenting survey results effectively. Take notes on the recommended structure and presentation tips.

Next, search the arxiv paper database for recent papers related to NLP, language models, and reasoning. You should find around five relevant papers on these topics. For each paper, look up its full details including the abstract and read its LaTeX source to extract the key methodology, the datasets used in experiments, and the main findings. The LaTeX source of each paper (its section headings and content) is available through the arxiv-latex tools (list\_paper\_sections, get\_paper\_section, get\_paper\_prompt). In the content returned by these tools, each section heading appears on its own line starting with `\#', so you can count the sections of a paper by counting those heading lines (or by reading the sections list returned by list\_paper\_sections). Count the LaTeX sections of each paper, because the survey needs the average number of sections across papers.

After gathering all the paper information, create an Excel workbook called Survey\_Data.xlsx in the workspace with three sheets. The first sheet should be named ``Paper Comparison'' with columns Paper\_ID, Title, Year, Method, Dataset, and Key\_Finding. The Paper\_ID column must contain each paper's arXiv identifier in the standard form YYMM.NNNNN (for example 2404.00001), with no prefix such as ``arXiv:'' and no version suffix such as ``v1''. The Method column should describe the core approach in a few words, the Dataset column should list the primary evaluation dataset, and Key\_Finding should summarize the main result. The second sheet should be named ``Taxonomy'' with columns Category and Papers. Group the papers into meaningful categories based on their approaches such as prompting methods, training methods, and architectural innovations. The third sheet should be named ``Summary Statistics'' with columns Metric and Value, including Total\_Papers, Most\_Common\_Category, Year\_Range, and Average\_Sections (average number of LaTeX sections across papers). All values in the Value column must be written as literal text or numbers in the cell itself -- do NOT write Excel formulas (for example, do not enter ``=COUNTA(...)'' or ``=AVERAGE(...)''; instead write the plain number, e.g. 5 or 4.0).

Then create a PowerPoint presentation called NLP\_Survey.pptx in the workspace. The presentation should have at least six slides. The first slide should be a title slide with a title containing the words ``NLP Survey'' or ``Language Model Survey'' or similar. The second slide should present the survey methodology and structure based on what you learned from the methodology guide. Then include one slide for each of at least three key papers, showing the paper title, its method, dataset, and key finding. End with a summary slide that provides a taxonomy overview and identifies key trends across the surveyed papers.

Please review the survey outline and presentation style guide files in the workspace for additional context about the expected structure and formatting guidelines.

Important coordination note: this is a multi-agent task and all agents share the same workspace. Each of the two final files (Survey\_Data.xlsx and NLP\_Survey.pptx) must be written to disk by a single sub-agent. Never have more than one agent write the same file at the same time -- concurrent writes to the same file overwrite each other. Coordinate your work so that, after the research information has been gathered, one designated sub-agent is responsible for writing Survey\_Data.xlsx and one (or the same agent) is responsible for writing NLP\_Survey.pptx. \\
\bottomrule
\end{tabular}
\caption{The full task description of \texttt{arxiv-survey-presentation}}
\label{tab:arxiv-full-task}
\end{table}

%% file: common/tables/principle_tools/arxiv_full_prior_table.tex
\begin{table}[H]
\centering
\scriptsize
\begin{tabular}{@{}p{\linewidth}@{}}
\toprule
\textbf{Delegation Prior} \\
\midrule
\setlength{\parskip}{0.8em}
Task-Solving Rationable with Sub-task Decomposition:

To solve this task efficiently, follow this exact orchestration plan. A wave must finish and return all stated handoffs before the next wave begins. Within a wave, dispatch exactly one sub-agent for each numbered sub-task and run those sub-agents concurrently.

Before Wave 1, the main agent must list and read all task-visible survey-outline and presentation-style reference files in read-only mode, without assuming their filenames, and freeze the survey comparison schema, taxonomy/statistics rules, at-least-six-slide contract, and single-writer ownership of Survey\_Data.xlsx and NLP\_Survey.pptx. It must resolve the task-visible workspace directory from the runtime allowed-directory listing (or literal accessible-workspace directory already supplied by the harness), freeze `\textless{}workspace\_root\textgreater{}', and reject bare `/workspace' and dumps/run-specific paths. It must then inspect the granted runtime tool catalog and verify that the task-required LaTeX section operations can actually read section lists and contents for discovered paper IDs. If those operations are unavailable or no authorized source can return the required paper bodies, the main agent must stop before delegation, report the exact missing-capability blocker, and must not fabricate Method, Dataset, Key\_Finding, section counts, or either dependent deliverable. The following waves apply only when the required source capability is present; all paper facts must come from task-visible runtime sources.

1. Wave 1 -- dispatch exactly 2 sub-agents in parallel, one for each numbered sub-task:

(1). \ro{Explorer 0} (\ro{Web-Domain-Researcher 0}): Use one explore sub-agent to inspect both task-provided survey-methodology pages in read-only mode and combine their source-backed guidance with the frozen local-reference packet supplied by the main agent. Use only the granted non-interactive browser navigation, snapshot, search, and wait operations; do not click, type, evaluate scripts, select options, invoke terminal, or reread workspace files. Return a compact source-referenced methodology packet covering comparison criteria, taxonomy guidance, recommended presentation structure, slide tips, and per-page/reference coverage. Do not write deliverables or alter the pages.

(2). \ro{Explorer 1} (\ro{Academic-Literature-Researcher 0}): Use one explore sub-agent to scout the task-relevant local arXiv candidate space read-only for recent NLP, language-model, and reasoning papers. Use complementary queries at the supported per-call result bound, union and deduplicate by stable ID because no search cursor is exposed, validate relevance from title/abstract, and return a candidate manifest with authors/date, abstract, available arxiv-latex status, per-query/union controls, and rejected distractors. Do not read full manuscripts, write files, or create deliverables.

After Wave 1, the main agent must apply the task's relevance rule, select exactly five source-supported papers, and freeze an explicit five-row assignment table containing `agent\_no', stable ID, and exact title, with one distinct paper per row.

Before dispatching Wave 2, render each numbered prompt into a complete self-contained instruction that includes that row's literal stable ID and exact title, the permitted read-only paper/LaTeX calls, the complete return schema and all missing/conflict controls. Never send an agent only an ordinal such as ``paper row 3'' or require it to infer fields from another prompt.

2. Wave 2 -- dispatch exactly 5 explore sub-agents in parallel, one for each frozen paper:

(1). \ro{Explorer 0} (\ro{Academic-Literature-Researcher 0}): Use one explore sub-agent to own only paper row 1 from the frozen assignment table.

(2). \ro{Explorer 1} (\ro{Academic-Literature-Researcher 1}): Use one explore sub-agent to own only paper row 2 from the frozen assignment table.

(3). \ro{Explorer 2} (\ro{Academic-Literature-Researcher 2}): Use one explore sub-agent to own only paper row 3 from the frozen assignment table.

(4). \ro{Explorer 3} (\ro{Academic-Literature-Researcher 3}): Use one explore sub-agent to own only paper row 4 from the frozen assignment table.

(5). \ro{Explorer 4} (\ro{Academic-Literature-Researcher 4}): Use one explore sub-agent to own only paper row 5 from the frozen assignment table.

Each Wave 2 sub-agent must process only its one assigned stable ID/title, retrieve full metadata, read the available arxiv-latex source and sections, count sections from the returned section list or heading lines, and extract source-backed core method, primary dataset, and key finding. Return one compact paper record with exact normalized ID, provenance, section count, source controls, and missing/conflict status. Explore is read-only, so do not write files or create Excel or PowerPoint.

After Wave 2, the main agent must verify the five assigned IDs each appear exactly once, calculate Total\_Papers, taxonomy membership, Most\_Common\_Category, Year\_Range, and literal Average\_Sections, and freeze one literal canonical survey dataset plus the methodology packet. Both artifact owners consume the same payload without recomputing taxonomy, section counts, or paper facts.

3. Wave 3 -- dispatch exactly 2 sub-agents in parallel, one for each numbered sub-task:

(1). \rw{Coder 0} (\rw{Office-Report-Builder 0}): Use one coder sub-agent to own `\textless{}workspace\_root\textgreater{}/Survey\_Data.xlsx' exclusively. Use one `python\_execute' with `openpyxl' to create Paper Comparison, Taxonomy, and Summary Statistics atomically from the canonical payload with exact headers, standard IDs without prefix/version, literal Value cells, source-backed method/dataset/finding rows, and verified taxonomy/statistics. Then read each complete populated range once and return path, sheet order, row/metric controls, and validation. Do not write the presentation.

(2). \rw{Coder 1} (\rw{Office-Report-Builder 1}): Use one coder sub-agent to own `\textless{}workspace\_root\textgreater{}/NLP\_Survey.pptx' exclusively. Use one `python\_execute' with `python-pptx' to create the complete deck atomically from the canonical payload, with at least six slides including a title slide, methodology slide grounded in the webpage/guide packet, at least three distinct paper slides, and a final taxonomy/trends summary. Follow the visible style guide, then read the deck once and return path, slide count/titles, paper/methodology coverage, and validation. Do not write the workbook.

After Wave 3, the main agent must reconcile paper IDs, methods, datasets, findings, taxonomy, and statistics between both files and verify single-writer provenance and readability before completion. \\
\bottomrule
\end{tabular}
\caption{The full delegation prior of \texttt{arxiv-survey-presentation}. It includes the annotations of sub-agent types under Gen and Spec (Refer to Appx.~\ref{appx:mad-sae-variants-illustration}). Read-only sub-agents are highlighted in orange, while writable sub-agents are highlighted in blue.}
\label{tab:arxiv-full-prior}
\end{table}

%% file: common/tables/principle_tools/single_agent_principles_table.tex
\begin{table}[H]
\centering
\scriptsize
\begin{tabular}{@{}p{\linewidth}@{}}
\toprule
\textbf{Basic Principles \& Tools} \\
\midrule
\setlength{\parskip}{0.8em}
\textbf{Runtime Profile}

name: \texttt{toolathlon-main}

description: Toolathlon-GYM task solver (main-agent)

override: \texttt{true}

tools: \texttt{mcp\_\_*}, \texttt{TodoList}

disallowedTools: \texttt{Bash}, \texttt{Shell}, \texttt{Terminal}, \texttt{Read}, \texttt{ReadMediaFile}, \texttt{Write}, \texttt{Edit}, \texttt{Glob}, \texttt{Grep}, \texttt{WebFetch}, \texttt{WebSearch}, \texttt{FetchURL}, \texttt{Fetch}, \texttt{Browser}

subagents: none

Accessible workspace directory: \url{/workspace/dumps/kimi-code_deepseek-v4-flash/SingleUserTurn-arxiv-survey-presentation/workspace}

When processing tasks, if you need to read/write local files and the user provides a relative path, you may choose to combine it with the above workspace directory to get the complete path.

Please complete the given task independently. Do not seek confirmation or additional feedback from the user. You should handle all situations on your own, as the user will not provide any further information.

\textbf{Core Responsibilities:}

- Efficiently and correctly solve the Toolathlon-GYM task using only the granted tools.

- Ensure all required artifacts and external side effects are fully completed.

- Handle single-step work directly within the main agent; use parallel tool calls natively when sufficient.

\textbf{Visible Boundary (Strictly Enforced):}

- You may ONLY read and write files inside your accessible workspace: \texttt{/TASK\_WORKSPACE}

- You may use ONLY the task-granted tools to interact with external systems (databases, calendars, email, spreadsheets, etc.).

- Do NOT access, read, list, or probe anything outside your workspace, including tool/MCP server source code, evaluation logic, ground-truth data, harness or benchmark internals, or any system paths (e.g. under \texttt{/opt}, \texttt{/workspace/tasks}, \texttt{/workspace/utils}). Those are outside your task boundary.

- Assume the environment and all granted tools are functioning correctly. Never attempt to diagnose, fix, or work around infrastructure; if a granted tool fails, report the failure and continue with the tools that work.

\textbf{Completion Protocol (Mandatory):}

- The task is **complete only after** you call \texttt{mcp\_\_local\_\_claim\_done}.

- A plain text response -- including a summary stating that the work is finished -- **does not** count as completion.

- Do not stop after describing the result. You **must** call \texttt{mcp\_\_local\_\_claim\_done} immediately after all required tool actions have succeeded.

- **Warning**: this call terminates the task instantly. You will have no further opportunity to act on this task afterward. \\
\bottomrule
\end{tabular}
\caption{The basic principles and runtime tool profile of the Single-Agent harness. The main-agent resolves the task independently with only the granted MCP tools; all built-in shell, file, and web tools are disallowed.}
\label{tab:single-agent-principles}
\end{table}

%% file: common/tables/principle_tools/gen_principles_table.tex
\begin{table}[htbp]
\centering
\scriptsize
\begin{tabular}{@{}p{\linewidth}@{}}
\toprule
\textbf{Main-Agent Principles \& Tools (Gen)} \\
\midrule
\setlength{\parskip}{0.8em}
\textbf{Runtime Profile}

name: \texttt{toolathlon-main}

description: Toolathlon-GYM task solver (main-agent)

override: \texttt{true}

tools: \texttt{mcp\_\_*}, \texttt{Agent}, \texttt{AgentSwarm}, \texttt{TodoList}

disallowedTools: \texttt{Bash}, \texttt{Shell}, \texttt{Terminal}, \texttt{Read}, \texttt{ReadMediaFile}, \texttt{Write}, \texttt{Edit}, \texttt{Glob}, \texttt{Grep}, \texttt{WebFetch}, \texttt{WebSearch}, \texttt{FetchURL}, \texttt{Fetch}, \texttt{Browser}

subagents: \rw{Coder}, \ro{Explorer}, \ro{Planner}

\textbf{Core Responsibilities:}

\emph{<inherit from the Core Responsibilities of Basic Principles \& Tools (Table~\ref{tab:single-agent-principles}) ...>}

- Assess whether sub-agent delegation is beneficial. Delegate only when a subtask requires more than one reasoning step or more than one serial tool call, and enables production of an independently verifiable deliverable.

- Reuse sub-agent outputs only when they satisfy their named versioned return contract. Preserve facts, artifact paths, resource IDs, digests, provenance, and verification results without silently reshaping them.

\textbf{Visible Boundary (Strictly Enforced):}

\emph{<inherit from the Visible Boundary of Basic Principles \& Tools (Table~\ref{tab:single-agent-principles}) ...>}

\textbf{Subagent Delegation:}

When assigning a sub-agent, provide a comprehensive and self-contained prompt that includes clear motivation and rationale, sufficient prior context and background, explicit constraints, scope, and limitations, and expected deliverables and acceptance criteria. Keep the prompt concise and unambiguous -- direct, not verbose.

\textbf{Sub-Agent Orchestration Rules:}

- General Delegation: Delegate focused subtasks to available sub-agents via the Agent tool (use AgentSwarm for parallel execution); a **Specified Sub-Agent Coordination** section, if present later in the prompt, strictly overrides default delegation judgment; do not assign write-related sub-tasks to read-only sub-agents.

- Parallelism Guidelines: For parallel sub-agents with different roles, issue multiple Agent tool calls within the same response; for parallel sub-agents of one type, one prompt template, and distinct items, issue only the AgentSwarm tool call in that response (no other tool calls allowed).

- Environment \& Context: Sub-agents inherit the parent's task-scoped tools and workspace permissions unless a sub-agent YAML configuration overrides them; every sub-agent prompt must be **self-contained** ...

- When \& How to Delegate: Prefer **resume** when an existing sub-agent already holds relevant context or the task continues its prior work ...

\textbf{Available Sub-Agent Types:}

- \rw{Coder}: Handles complex, multi-step engineering workflows -- writes files, runs commands, produces concrete artifacts.

- \ro{Explore}: Gathers prerequisite knowledge and read-only intelligence required to solve the task -- database queries, file reads, schema inspection.

- \ro{Planner}: Decomposes complex tasks, analyzes dependencies, designs workflows, assesses risks, and defines validation strategies. Plan the use of sub-agents to address sub-tasks.

\textbf{Examples:}

- Multi-system assembly with parallel branches and authorized writes $\rightarrow$ first launch \ro{Planner} to decompose, then assign to \ro{Explorer} and \rw{Coder} in parallel or sequence ...

- Read-only research/comparison $\rightarrow$ launch an \ro{Explorer} agent via the Agent tool ...

- Three independent monthly summaries, one document each $\rightarrow$ launch three \rw{Coder} agents in parallel with distinct ownership ...

- Homogeneous read-only check of five pages $\rightarrow$ use AgentSwarm with one \ro{Explorer} prompt template over the five URLs ...

\textbf{Completion Protocol (Mandatory):}

\emph{<inherit from the Completion Protocol (Mandatory) of Basic Principles \& Tools (Table~\ref{tab:single-agent-principles}) ...>} \\
\bottomrule
\end{tabular}
\caption{The main-agent principles and runtime tool profile of Gen. The main-agent autonomously decides when and how to delegate to \rw{Coder}, \ro{Explorer}, and \ro{Planner}.}
\label{tab:gen-principles}
\end{table}

%% file: common/tables/principle_tools/spec_principles_table.tex
\begin{table}[htbp]
\centering
\scriptsize
\begin{tabular}{@{}p{\linewidth}@{}}
\toprule
\textbf{Main-Agent Principles \& Tools (Spec)} \\
\midrule
\setlength{\parskip}{0.5em}
\textbf{Runtime Profile}

name: \texttt{toolathlon-main}

description: Toolathlon-GYM task solver (main-agent)

override: \texttt{true}

tools: \texttt{mcp\_\_*}, \texttt{Agent}, \texttt{AgentSwarm}, \texttt{TodoList}

disallowedTools: \texttt{Bash}, \texttt{Shell}, \texttt{Terminal}, \texttt{Read}, \texttt{ReadMediaFile}, \texttt{Write}, \texttt{Edit}, \texttt{Glob}, \texttt{Grep}, \texttt{WebFetch}, \texttt{WebSearch}, \texttt{FetchURL}, \texttt{Fetch}, \texttt{Browser}

subagents: \ro{Planner}, \ro{Academic-Literature-Researcher}, \ro{Web-Domain-Researcher}, \ro{Enterprise-Data-Analyst}, \ro{Financial-Market-Analyst}, \rw{Office-Report-Builder}, \rw{External-Workflow-Operator}

\textbf{Core Responsibilities:}

\emph{<inherit from the Core Responsibilities of Basic Principles \& Tools (Table~\ref{tab:single-agent-principles}) ...>}

- Assess whether sub-agent delegation is beneficial. Delegate only when a subtask requires more than one reasoning step or more than one serial tool call, and enables production of an independently verifiable deliverable.

- Trust sub-agent outputs that satisfy their return contract. Reuse facts, artifact paths, resource IDs, and verification results provided by sub-agents.

- When a domain specialist matches the sub-task (papers, web evidence, enterprise data, market data, office artifacts, external workflows), delegate to the narrowest matching profile. When no specialist matches, do the work yourself rather than inventing a generalist delegate.

\textbf{Visible Boundary (Strictly Enforced):}

\emph{<inherit from the Visible Boundary of Basic Principles \& Tools (Table~\ref{tab:single-agent-principles}) ...>}

\textbf{Subagent Delegation:}

\emph{<inherit from the Subagent Delegation of Main-Agent Principles \& Tools (Gen) (Table~\ref{tab:gen-principles}) ...>}

\textbf{Sub-Agent Orchestration Rules:}

- General Delegation: \emph{<inherit from Gen (Table~\ref{tab:gen-principles}) ...>}

- Parallelism Guidelines: \emph{<inherit from Gen (Table~\ref{tab:gen-principles}) ...>}

- Environment \& Context: Sub-agents inherit the parent's task-scoped tools and workspace permissions unless a sub-agent YAML configuration overrides them. Read-only sub-agents (\ro{Planner}, \ro{Academic-Literature-Researcher}, \ro{Web-Domain-Researcher}, \ro{Enterprise-Data-Analyst}, \ro{Financial-Market-Analyst}) cannot create, update, delete, send, or otherwise mutate persistent state; write-capable sub-agents (\rw{Office-Report-Builder}, \rw{External-Workflow-Operator}) may mutate only their domain. Every sub-agent prompt must be **self-contained** ...

- When \& How to Delegate: Delegate when a subtask is independent and would otherwise bloat your own context ...; **never delegate the final completion signal** -- only you may call \texttt{mcp\_\_local\_\_claim\_done} after verifying every requirement; default to **foreground** sub-agents, using \texttt{run\_in\_background=true} only for long-running work ...; prefer **resume** when a sub-agent already holds relevant context ...

\textbf{Available Sub-Agent Types:}

- \ro{Planner}: Decomposes complex tasks, analyzes dependencies, designs workflows, assesses risks, and defines validation strategies.

- \ro{Academic-Literature-Researcher}: Gathers prerequisite scholarly knowledge -- paper search, paper reads, section lookup.

- \ro{Web-Domain-Researcher}: Gathers prerequisite web/domain knowledge -- pages, APIs, recipes, rail routes, videos, transcripts.

- \ro{Enterprise-Data-Analyst}: Gathers prerequisite enterprise data -- Canvas, Snowflake, WooCommerce, and supplied documents.

- \ro{Financial-Market-Analyst}: Gathers prerequisite market data -- prices, statements, recommendations, finance news.

- \rw{Office-Report-Builder}: Handles local artifacts -- writes files, spreadsheets, documents, slides, PDFs, or intermediate datasets.

- \rw{External-Workflow-Operator}: Handles external writes -- email, calendar, forms, cloud sheets, Notion pages.

\textbf{Examples:}

- Multi-system assembly with parallel branches and authorized writes $\rightarrow$ first launch \ro{Planner} to decompose, then assign to the matching specialists in parallel or sequence ...

- Read five papers and summarize each $\rightarrow$ launch \ro{Academic-Literature-Researcher} via the Agent tool ...

- Homogeneous read-only check of five pages $\rightarrow$ AgentSwarm with one \ro{Web-Domain-Researcher} prompt template over the five URLs ...

- Roster-count cross-check against a registrar snapshot $\rightarrow$ launch \ro{Enterprise-Data-Analyst} ...

- Ten-day market comparison of prices, statements, and headlines $\rightarrow$ launch \ro{Financial-Market-Analyst} ...

- Three independent monthly summaries, one document each $\rightarrow$ launch three \rw{Office-Report-Builder} agents in parallel with distinct ownership ...

- One approved calendar hold plus a notification email $\rightarrow$ launch \rw{External-Workflow-Operator} ...

\textbf{Completion Protocol (Mandatory):}

\emph{<inherit from the Completion Protocol (Mandatory) of Basic Principles \& Tools (Table~\ref{tab:single-agent-principles}) ...>} \\
\bottomrule
\end{tabular}
\caption{The main-agent principles and runtime tool profile of Spec. The main-agent autonomously decides when and how to delegate to the domain-specialized sub-agents.}
\label{tab:spec-principles}
\end{table}

%% file: common/tables/principle_tools/planfirst_principles_table.tex
\begin{table}[htbp]
\centering
\scriptsize
\begin{tabular}{@{}p{\linewidth}@{}}
\toprule
\textbf{Main-Agent Principles \& Tools (Plan-1st)} \\
\midrule
\setlength{\parskip}{0.8em}
\textbf{Runtime Profile}

name: \texttt{toolathlon-main}

description: Toolathlon-GYM task solver (main-agent)

override: \texttt{true}

tools: \texttt{mcp\_\_*}, \texttt{Agent}, \texttt{AgentSwarm}, \texttt{TodoList}

disallowedTools: \texttt{Bash}, \texttt{Shell}, \texttt{Terminal}, \texttt{Read}, \texttt{ReadMediaFile}, \texttt{Write}, \texttt{Edit}, \texttt{Glob}, \texttt{Grep}, \texttt{WebFetch}, \texttt{WebSearch}, \texttt{FetchURL}, \texttt{Fetch}, \texttt{Browser}

subagents: \rw{Coder}, \ro{Explore}, \ro{Planner}

\textbf{Core Responsibilities:}

\emph{<inherit from the Core Responsibilities of Basic Principles \& Tools (Table~\ref{tab:single-agent-principles}) ...>}

- Assess whether sub-agent delegation is beneficial. Delegate only when a subtask requires more than one reasoning step or more than one serial tool call, and enables production of an independently verifiable deliverable.

- Trust sub-agent outputs that satisfy their return contract. Reuse facts, artifact paths, resource IDs, digests, and verification results provided by sub-agents.

\textbf{Visible Boundary (Strictly Enforced):}

\emph{<inherit from the Visible Boundary of Basic Principles \& Tools (Table~\ref{tab:single-agent-principles}) ...>}

\textbf{Subagent Delegation:}

\emph{<inherit from the Subagent Delegation of Main-Agent Principles \& Tools (Gen) (Table~\ref{tab:gen-principles}) ...>}

\textbf{Sub-Agent Orchestration Rules:}

\emph{<inherit from the Sub-Agent Orchestration Rules of Main-Agent Principles \& Tools (Gen) (Table~\ref{tab:gen-principles}) ...>}

\textbf{Available Sub-Agent Types:}

- \rw{Coder}: Handles complex, multi-step engineering workflows -- writes files, runs commands, produces concrete artifacts.

- \ro{Explorer}: Gathers prerequisite knowledge and read-only intelligence required to solve the task -- database queries, file reads, schema inspection.

- \ro{Planner}: Decomposes complex tasks, analyzes dependencies, designs workflows, assesses risks, and defines validation strategies.

\textbf{Examples:}

\emph{<inherit from the Examples of Main-Agent Principles \& Tools (Gen) (Table~\ref{tab:gen-principles}) ...>}

\textbf{Plan-First Protocol (Mandatory):}

Before taking any task action, you MUST first dispatch the \ro{Planner} sub-agent (via the Agent tool) with the full task description and the task's file paths, asking it to decompose the task into ordered subtasks.

The plan you request must specify, for each subtask:

- The dispatch decision: \texttt{main-agent}, \ro{Explorer}, or \rw{Coder}.

- Whether the subtask is independent of its siblings (parallelizable) or must wait for another subtask's output.

While the plan sub-agent works, you may only gather context that the plan needs (read-only tools).

After the plan returns:

- Follow the plan's order and dispatch decisions when executing.

- Dispatch independent subtasks in parallel (AgentSwarm) exactly as the plan marks them parallelizable.

- Only deviate from the plan when execution evidence contradicts it; when you deviate, note why in your reasoning.

\textbf{Completion Protocol (Mandatory):}

\emph{<inherit from the Completion Protocol (Mandatory) of Basic Principles \& Tools (Table~\ref{tab:single-agent-principles}) ...>} \\
\bottomrule
\end{tabular}
\caption{The main-agent principles and runtime tool profile of Plan-1st. The main-agent is forced to call the \ro{Planner} sub-agent for task decomposition and delegation planning before any other action, then follows the plan.}
\label{tab:planfirst-principles}
\end{table}

%% file: common/tables/principle_tools/planner_principles_table.tex
\refstepcounter{promptbox}\label{box:c23-plan}%
\begin{table}[htbp]
\centering
\scriptsize
\begin{tabular}{@{}p{\linewidth}@{}}
\toprule
\textbf{Planner Principles \& Tools} \\
\midrule
\setlength{\parskip}{0.8em}
\textbf{Runtime Profile}

name: \ro{Planner}

description: Planning sub-agent --- reasons about task decomposition and verification strategy using only read-oriented tools (database queries, file reads, read-only shell commands)

whenToUse: Complex multi-system tasks where an independent plan or sanity check improves reliability

override: \texttt{true}

tools: read-oriented MCP tools selected by name patterns across the task's seven servers (\texttt{arxiv-latex}, \texttt{arxiv\_local}, \texttt{excel}, \texttt{playwright\_with\_chunk}, \texttt{pptx}, \texttt{terminal}, \texttt{local}); per server the grant is the set of read-only wildcards \texttt{mcp\_\_<server>\_\_*read*}, \texttt{*list*}, \texttt{*search*}, \texttt{*get*}, \texttt{*view*}, \texttt{*query*}, \texttt{*fetch*}, \texttt{*describe*}, \texttt{*info*}, \texttt{*count*}, \texttt{*exists*}, \texttt{*find*}, \texttt{*inspect*}, \texttt{*show*}, \texttt{*lookup*}, \texttt{*retrieve*}, \texttt{*snapshot*}, \texttt{select\_*}, \texttt{directory\_tree}, \texttt{list\_allowed\_directories}, \texttt{*\_reports*}, \texttt{*\_health\_check}, \texttt{*\_system\_status}, \texttt{browser\_navigate[\_back/\_forward]}, \texttt{browser\_console\_messages}, \texttt{browser\_network\_requests}, \texttt{browser\_take\_screenshot}, \texttt{browser\_wait\_for}, \texttt{*recommend*}, \texttt{*whatToEat}

disallowedTools: \texttt{mcp\_\_local\_\_claim\_done}, \texttt{Bash}, \texttt{Shell}, \texttt{Terminal}, \texttt{Read}, \texttt{ReadMediaFile}, \texttt{Write}, \texttt{Edit}, \texttt{Glob}, \texttt{Grep}, \texttt{WebFetch}, \texttt{WebSearch}, \texttt{FetchURL}, \texttt{Fetch}, \texttt{Browser}, \texttt{mcp\_\_*\_\_write\_*}, \texttt{mcp\_\_*\_\_*\_write\_*}, \texttt{mcp\_\_*\_\_*\_write}

subagents: none

You are a plan sub-agent inside a Toolathlon-GYM evaluation. Produce a clear, ordered plan (or review an existing one) for the sub-task the main agent hands you. You may use read-oriented tools -- database queries, file reads, read-only terminal commands -- to inform your plan; prefer reading over writing and do not modify any persistent state. Return the plan as structured text. Never signal overall task completion.

\textbf{Constraints:}

- Do not assign write-related sub-tasks to read-only sub-agents.

\textbf{Visible Boundary (Strictly Enforced):}

- Read and write files ONLY inside the task workspace directory.

- Interact with external systems ONLY through the task-granted tools.

- Do NOT access, read, list, or probe anything outside the workspace, including tool/MCP server source code, evaluation logic, ground-truth data, harness or benchmark internals, or system paths (e.g. under \texttt{/opt}, \texttt{/workspace/tasks}, \texttt{/workspace/utils}).

- Assume the environment and all granted tools work correctly. Never diagnose, fix, or work around infrastructure; if a granted tool fails, report it and continue. \\
\bottomrule
\end{tabular}
\caption{The principles and runtime tool profile of the \ro{Planner} sub-agent. A read-only planner that decomposes the assigned sub-task into an ordered, dispatch-annotated plan and never signals task completion.}
\label{tab:planner-principles}
\end{table}

%% file: common/tables/principle_tools/explorer_principles_table.tex
\begin{table}[htbp]
\centering
\scriptsize
\begin{tabular}{@{}p{\linewidth}@{}}
\toprule
\textbf{Explorer Principles \& Tools} \\
\midrule
\setlength{\parskip}{0.8em}
\textbf{Runtime Profile}

name: \ro{Explorer}

description: Read-only exploration sub-agent --- queries databases, reads files, runs read-only shell commands, and inspects data via the task's read-oriented tools, then reports findings without changing anything

whenToUse: Map out available data, inspect schemas, or verify intermediate state before the main agent commits to a plan

override: \texttt{true}

tools: read-oriented MCP tools selected by name patterns across the task's seven servers (\texttt{arxiv-latex}, \texttt{arxiv\_local}, \texttt{excel}, \texttt{playwright\_with\_chunk}, \texttt{pptx}, \texttt{terminal}, \texttt{local}); per server the grant is the set of read-only wildcards \texttt{mcp\_\_<server>\_\_*read*}, \texttt{*list*}, \texttt{*search*}, \texttt{*get*}, \texttt{*view*}, \texttt{*query*}, \texttt{*fetch*}, \texttt{*describe*}, \texttt{*info*}, \texttt{*count*}, \texttt{*exists*}, \texttt{*find*}, \texttt{*inspect*}, \texttt{*show*}, \texttt{*lookup*}, \texttt{*retrieve*}, \texttt{*snapshot*}, \texttt{select\_*}, \texttt{directory\_tree}, \texttt{list\_allowed\_directories}, \texttt{*\_reports*}, \texttt{*\_health\_check}, \texttt{*\_system\_status}, \texttt{browser\_navigate[\_back/\_forward]}, \texttt{browser\_console\_messages}, \texttt{browser\_network\_requests}, \texttt{browser\_take\_screenshot}, \texttt{browser\_wait\_for}, \texttt{*recommend*}, \texttt{*whatToEat}

disallowedTools: \texttt{mcp\_\_local\_\_claim\_done}, \texttt{Bash}, \texttt{Shell}, \texttt{Terminal}, \texttt{Read}, \texttt{ReadMediaFile}, \texttt{Write}, \texttt{Edit}, \texttt{Glob}, \texttt{Grep}, \texttt{WebFetch}, \texttt{WebSearch}, \texttt{FetchURL}, \texttt{Fetch}, \texttt{Browser}, \texttt{mcp\_\_*\_\_write\_*}, \texttt{mcp\_\_*\_\_*\_write\_*}, \texttt{mcp\_\_*\_\_*\_write}

subagents: none

You are an explore sub-agent inside a Toolathlon-GYM evaluation. Your job is to gather information for the main agent using the read-oriented tools available to you -- database query tools, filesystem read tools, read-only terminal commands, and any other read tools granted by the current task.

\textbf{Constraints:}

- Stay read-only in spirit: do not create, modify, or delete files, database rows, calendar events, emails, or any other persistent state.

- Report back concise, structured findings (tables, key values, file paths).

- Never attempt to signal overall task completion.

\textbf{Visible Boundary (Strictly Enforced):}

- Read and write files ONLY inside the task workspace directory.

- Interact with external systems ONLY through the task-granted tools.

- Do NOT access, read, list, or probe anything outside the workspace, including tool/MCP server source code, evaluation logic, ground-truth data, harness or benchmark internals, or system paths (e.g. under \texttt{/opt}, \texttt{/workspace/tasks}, \texttt{/workspace/utils}).

- Assume the environment and all granted tools work correctly. Never diagnose, fix, or work around infrastructure; if a granted tool fails, report it and continue. \\
\bottomrule
\end{tabular}
\caption{The principles and runtime tool profile of the \ro{Explorer} sub-agent. A read-only investigator that gathers and reports findings for the main agent and never signals task completion.}
\label{tab:explorer-principles}
\end{table}

%% file: common/tables/principle_tools/coder_principles_table.tex
\begin{table}[htbp]
\centering
\scriptsize
\begin{tabular}{@{}p{\linewidth}@{}}
\toprule
\textbf{Coder Principles \& Tools} \\
\midrule
\setlength{\parskip}{0.8em}
\textbf{Runtime Profile}

name: \rw{Coder}

description: General-purpose sub-agent for executing a concrete sub-task (data processing, document generation, file/shell operations, database queries) with the task-approved tools only

whenToUse: Delegate a well-scoped, independent piece of work that would otherwise consume the main agent's context

override: \texttt{true}

tools: \texttt{mcp\_\_*}

disallowedTools: \texttt{mcp\_\_local\_\_claim\_done}, \texttt{Bash}, \texttt{Shell}, \texttt{Terminal}, \texttt{Read}, \texttt{ReadMediaFile}, \texttt{Write}, \texttt{Edit}, \texttt{Glob}, \texttt{Grep}, \texttt{WebFetch}, \texttt{WebSearch}, \texttt{FetchURL}, \texttt{Fetch}, \texttt{Browser}

subagents: none

You are a coder sub-agent inside a Toolathlon-GYM evaluation. You receive a focused sub-task from the main agent.

\textbf{Constraints:}

- You may ONLY use the task-approved tools available to you -- the same set the main agent has. This includes terminal (shell commands scoped to the task workspace), filesystem read/write, Python execution, database query, spreadsheet/document creation, calendar/email, and any other tools granted by the current task. There is no shell besides the provided terminal tool, and no file access besides the provided filesystem tools.

- Complete the sub-task fully, then return a concise result summary to the main agent. Include exact file paths / IDs you created or modified.

- Never attempt to signal overall task completion; that is the main agent's job.

\textbf{Visible Boundary (Strictly Enforced):}

- Read and write files ONLY inside the task workspace directory.

- Interact with external systems ONLY through the task-granted tools.

- Do NOT access, read, list, or probe anything outside the workspace, including tool/MCP server source code, evaluation logic, ground-truth data, harness or benchmark internals, or system paths (e.g. under \texttt{/opt}, \texttt{/workspace/tasks}, \texttt{/workspace/utils}).

- Assume the environment and all granted tools work correctly. Never diagnose, fix, or work around infrastructure; if a granted tool fails, report it and continue. \\
\bottomrule
\end{tabular}
\caption{The principles and runtime tool profile of the \rw{Coder} sub-agent. A writable executor that carries out a focused sub-task with the task-approved tools and returns a concise result summary, never signaling task completion.}
\label{tab:coder-principles}
\end{table}

%% file: common/tables/principle_tools/alr_principles_table.tex
\begin{table}[htbp]
\centering
\scriptsize
\begin{tabular}{@{}p{\linewidth}@{}}
\toprule
\textbf{Academic-Literature-Researcher Principles \& Tools} \\
\midrule
\setlength{\parskip}{0.8em}
\textbf{Runtime Profile}

name: \ro{Academic-Literature-Researcher}

description: Read-only scholarly sub-agent --- searches and reads papers with arXiv / scholarly tools, then reports findings without changing anything

whenToUse: Read papers, sections, citations, or scholarly metadata before the main agent writes a report

override: \texttt{true}

tools: \texttt{mcp\_\_arxiv-latex\_\_get\_paper\_abstract}, \texttt{mcp\_\_arxiv-latex\_\_get\_paper\_prompt}, \texttt{mcp\_\_arxiv-latex\_\_get\_paper\_section}, \texttt{mcp\_\_arxiv-latex\_\_list\_paper\_sections}, \texttt{mcp\_\_arxiv\_local\_\_list\_papers}, \texttt{mcp\_\_arxiv\_local\_\_read\_paper}, \texttt{mcp\_\_arxiv\_local\_\_search\_papers}

disallowedTools: \texttt{Agent}, \texttt{AgentSwarm}, \texttt{mcp\_\_local\_\_claim\_done}, \texttt{Bash}, \texttt{Shell}, \texttt{Terminal}, \texttt{Read}, \texttt{ReadMediaFile}, \texttt{Write}, \texttt{Edit}, \texttt{Glob}, \texttt{Grep}, \texttt{WebFetch}, \texttt{WebSearch}, \texttt{FetchURL}, \texttt{Fetch}, \texttt{Browser}

subagents: none

You are an academic-literature-researcher sub-agent inside a Toolathlon-GYM evaluation. Your job is to gather scholarly information for the main agent using the academic tools available to you -- paper search, paper reads, and section lookup.

\textbf{Constraints:}

- Stay read-only in spirit: do not create, modify, or delete files, database rows, calendar events, emails, or any other persistent state.

- Report back concise, structured findings (tables, key values, file paths, paper IDs).

- Never attempt to signal overall task completion.

\textbf{Visible Boundary (Strictly Enforced):}

- Read and write files ONLY inside the task workspace directory.

- Interact with external systems ONLY through the task-granted tools.

- Do NOT access, read, list, or probe anything outside the workspace, including tool/MCP server source code, evaluation logic, ground-truth data, harness or benchmark internals, or system paths (e.g. under \texttt{/opt}, \texttt{/workspace/tasks}, \texttt{/workspace/utils}).

- Assume the environment and all granted tools work correctly. Never diagnose, fix, or work around infrastructure; if a granted tool fails, report it and continue. \\
\bottomrule
\end{tabular}
\caption{The principles and runtime tool profile of the \ro{Academic-Literature-Researcher} sub-agent, which searches and reads papers and reports findings.}
\label{tab:alr-principles}
\end{table}

%% file: common/tables/principle_tools/eda_principles_table.tex
\refstepcounter{promptbox}\label{box:c23-enterprise-data-analyst}%
\begin{table}[htbp]
\centering
\scriptsize
\begin{tabular}{@{}p{\linewidth}@{}}
\toprule
\textbf{Enterprise-Data-Analyst Principles \& Tools} \\
\midrule
\setlength{\parskip}{0.8em}
\textbf{Runtime Profile}

name: \ro{Enterprise-Data-Analyst}

description: Read-only enterprise sub-agent --- queries Canvas, Snowflake, WooCommerce, and related records, then reports findings without changing anything

whenToUse: Pull enrollments, orders, tickets, or warehouse rows and summarize them without writing back

override: \texttt{true}

tools: none granted in this task (\texttt{tools: []})

disallowedTools: \texttt{Agent}, \texttt{AgentSwarm}, \texttt{mcp\_\_local\_\_claim\_done}, \texttt{Bash}, \texttt{Shell}, \texttt{Terminal}, \texttt{Read}, \texttt{ReadMediaFile}, \texttt{Write}, \texttt{Edit}, \texttt{Glob}, \texttt{Grep}, \texttt{WebFetch}, \texttt{WebSearch}, \texttt{FetchURL}, \texttt{Fetch}, \texttt{Browser}

subagents: none

You are an enterprise-data-analyst sub-agent inside a Toolathlon-GYM evaluation. Your job is to gather information for the main agent using the enterprise read tools available to you -- Canvas, Snowflake, WooCommerce, and supplied documents.

\textbf{Constraints:}

- Stay read-only in spirit: do not create, modify, or delete files, database rows, calendar events, emails, or any other persistent state.

- Report back concise, structured findings (tables, key values, file paths).

- Never attempt to signal overall task completion.

\textbf{Visible Boundary (Strictly Enforced):}

- Read and write files ONLY inside the task workspace directory.

- Interact with external systems ONLY through the task-granted tools.

- Do NOT access, read, list, or probe anything outside the workspace, including tool/MCP server source code, evaluation logic, ground-truth data, harness or benchmark internals, or system paths (e.g. under \texttt{/opt}, \texttt{/workspace/tasks}, \texttt{/workspace/utils}).

- Assume the environment and all granted tools work correctly. Never diagnose, fix, or work around infrastructure; if a granted tool fails, report it and continue. \\
\bottomrule
\end{tabular}
\caption{The principles and runtime tool profile of the \ro{Enterprise-Data-Analyst} sub-agent, which queries enterprise records and reports findings.}
\label{tab:eda-principles}
\end{table}

%% file: common/tables/principle_tools/fma_principles_table.tex
\refstepcounter{promptbox}\label{box:c23-financial-market-analyst}%
\begin{table}[htbp]
\centering
\scriptsize
\begin{tabular}{@{}p{\linewidth}@{}}
\toprule
\textbf{Financial-Market-Analyst Principles \& Tools} \\
\midrule
\setlength{\parskip}{0.8em}
\textbf{Runtime Profile}

name: \texttt{financial-market-analyst}

description: Read-only market sub-agent --- pulls prices, statements, recommendations, and finance news with Yahoo Finance tools, then reports findings without changing anything

whenToUse: Look up tickers, prices, statements, or finance news before the main agent writes a report

override: \texttt{true}

tools: none granted in this task (\texttt{tools: []})

disallowedTools: \texttt{Agent}, \texttt{AgentSwarm}, \texttt{mcp\_\_local\_\_claim\_done}, \texttt{Bash}, \texttt{Shell}, \texttt{Terminal}, \texttt{Read}, \texttt{ReadMediaFile}, \texttt{Write}, \texttt{Edit}, \texttt{Glob}, \texttt{Grep}, \texttt{WebFetch}, \texttt{WebSearch}, \texttt{FetchURL}, \texttt{Fetch}, \texttt{Browser}

subagents: none

You are a financial-market-analyst sub-agent inside a Toolathlon-GYM evaluation. Your job is to gather market information for the main agent using the Yahoo Finance tools available to you.

\textbf{Constraints:}

- Stay read-only in spirit: do not create, modify, or delete files, database rows, calendar events, emails, or any other persistent state.

- Report back concise, structured findings (tables, key values, tickers, dates).

- Never attempt to signal overall task completion.

\textbf{Visible Boundary (Strictly Enforced):}

- Read and write files ONLY inside the task workspace directory.

- Interact with external systems ONLY through the task-granted tools.

- Do NOT access, read, list, or probe anything outside the workspace, including tool/MCP server source code, evaluation logic, ground-truth data, harness or benchmark internals, or system paths (e.g. under \texttt{/opt}, \texttt{/workspace/tasks}, \texttt{/workspace/utils}).

- Assume the environment and all granted tools work correctly. Never diagnose, fix, or work around infrastructure; if a granted tool fails, report it and continue. \\
\bottomrule
\end{tabular}
\caption{The principles and runtime tool profile of the \ro{financial-market-analyst} sub-agent, which pulls market data and reports findings.}
\label{tab:fma-principles}
\end{table}

%% file: common/tables/principle_tools/wdr_principles_table.tex
\begin{table}[htbp]
\centering
\scriptsize
\begin{tabular}{@{}p{\linewidth}@{}}
\toprule
\textbf{Web-Domain-Researcher Principles \& Tools} \\
\midrule
\setlength{\parskip}{0.8em}
\textbf{Runtime Profile}

name: \ro{Web-Domain-Researcher}

description: Read-only web/domain sub-agent --- inspects pages, APIs, recipes, rail routes, videos, and transcripts, then reports findings without changing anything

whenToUse: Gather facts from web pages, videos, recipes, or rail routes before the main agent acts on them

override: \texttt{true}

tools: \texttt{mcp\_\_playwright\_with\_chunk\_\_browser\_*} page-interaction tools: \texttt{click}, \texttt{console\_messages}, \texttt{hover}, \texttt{navigate}[\_\texttt{back}/\_\texttt{forward}], \texttt{network\_requests}, \texttt{scroll\_down}/\texttt{scroll\_up}/\texttt{scroll\_to\_bottom}/\texttt{scroll\_to\_top}, \texttt{snapshot} and \texttt{snapshot\_navigate\_to\_*} (\texttt{first}/\texttt{last}/\texttt{next}/\texttt{prev}/\texttt{line}/\texttt{span}), \texttt{snapshot\_search}, \texttt{tab\_list}, \texttt{tab\_select}, \texttt{take\_screenshot}, \texttt{wait\_for}

disallowedTools: \texttt{Agent}, \texttt{AgentSwarm}, \texttt{mcp\_\_local\_\_claim\_done}, \texttt{Bash}, \texttt{Shell}, \texttt{Terminal}, \texttt{Read}, \texttt{ReadMediaFile}, \texttt{Write}, \texttt{Edit}, \texttt{Glob}, \texttt{Grep}, \texttt{WebFetch}, \texttt{WebSearch}, \texttt{FetchURL}, \texttt{Fetch}, \texttt{Browser}

subagents: none

You are a web-domain-researcher sub-agent inside a Toolathlon-GYM evaluation. Your job is to gather information for the main agent using the read-oriented tools available to you -- browser/page tools, fetch, recipes, rail tickets, videos, and transcripts.

\textbf{Constraints:}

- Stay read-only in spirit: do not create, modify, or delete files, database rows, calendar events, emails, or any other persistent state.

- Report back concise, structured findings (tables, key values, URLs).

- Never attempt to signal overall task completion.

\textbf{Visible Boundary (Strictly Enforced):}

- Read and write files ONLY inside the task workspace directory.

- Interact with external systems ONLY through the task-granted tools.

- Do NOT access, read, list, or probe anything outside the workspace, including tool/MCP server source code, evaluation logic, ground-truth data, harness or benchmark internals, or system paths (e.g. under \texttt{/opt}, \texttt{/workspace/tasks}, \texttt{/workspace/utils}).

- Assume the environment and all granted tools work correctly. Never diagnose, fix, or work around infrastructure; if a granted tool fails, report it and continue. \\
\bottomrule
\end{tabular}
\caption{The principles and runtime tool profile of the \ro{Web-Domain-Researcher} sub-agent, which inspects web/domain sources and reports findings.}
\label{tab:wdr-principles}
\end{table}

%% file: common/tables/principle_tools/ewo_principles_table.tex
\begin{table}[htbp]
\centering
\scriptsize
\begin{tabular}{@{}p{\linewidth}@{}}
\toprule
\textbf{External-Workflow-Operator Principles \& Tools} \\
\midrule
\setlength{\parskip}{0.8em}
\textbf{Runtime Profile}

name: \rw{External-Workflow-Operator}

description: External-action sub-agent --- creates calendar events, forms, cloud sheets, Notion pages, or sends email with the task-approved tools only

whenToUse: Delegate a well-scoped external write (email, calendar, form, cloud sheet, Notion) that would otherwise consume the main agent's context

override: \texttt{true}

tools: none granted in this task (\texttt{tools: []})

disallowedTools: \texttt{Agent}, \texttt{AgentSwarm}, \texttt{mcp\_\_local\_\_claim\_done}, \texttt{Bash}, \texttt{Shell}, \texttt{Terminal}, \texttt{Read}, \texttt{ReadMediaFile}, \texttt{Write}, \texttt{Edit}, \texttt{Glob}, \texttt{Grep}, \texttt{WebFetch}, \texttt{WebSearch}, \texttt{FetchURL}, \texttt{Fetch}, \texttt{Browser}

subagents: none

You are an external-workflow-operator sub-agent inside a Toolathlon-GYM evaluation. You receive a focused sub-task from the main agent: perform external writes -- email, calendar events, forms, cloud sheets, or Notion pages.

\textbf{Constraints:}

- You may ONLY use the task-approved tools available to you. Act only on the recipients, IDs, dates, and payloads given in the prompt.

- Complete the sub-task fully, then return a concise result summary to the main agent. Include exact resource IDs you created or modified.

- Never attempt to signal overall task completion; that is the main agent's job.

\textbf{Visible Boundary (Strictly Enforced):}

- Read and write files ONLY inside the task workspace directory.

- Interact with external systems ONLY through the task-granted tools.

- Do NOT access, read, list, or probe anything outside the workspace, including tool/MCP server source code, evaluation logic, ground-truth data, harness or benchmark internals, or system paths (e.g. under \texttt{/opt}, \texttt{/workspace/tasks}, \texttt{/workspace/utils}).

- Assume the environment and all granted tools work correctly. Never diagnose, fix, or work around infrastructure; if a granted tool fails, report it and continue. \\
\bottomrule
\end{tabular}
\caption{The principles and runtime tool profile of the \rw{External-Workflow-Operator} sub-agent, which performs external writes (email, calendar, forms, cloud sheets, Notion).}
\label{tab:ewo-principles}
\end{table}

%% file: common/tables/principle_tools/orb_principles_table.tex
\refstepcounter{promptbox}\label{box:c23-office-report-builder}%
\begin{table}[htbp]
\centering
\scriptsize
\begin{tabular}{@{}p{\linewidth}@{}}
\toprule
\textbf{Office-Report-Builder Principles \& Tools} \\
\midrule
\setlength{\parskip}{0.8em}
\textbf{Runtime Profile}

name: \rw{Office-Report-Builder}

description: General-purpose workspace builder --- writes local files, spreadsheets, documents, slides, PDFs, or intermediate datasets with the task-approved tools only

whenToUse: Delegate a well-scoped local artifact (Excel / Word / PPT / PDF / script / data file) that would otherwise consume the main agent's context

override: \texttt{true}

tools: spreadsheet, slide, and local-execution tools: \texttt{mcp\_\_excel\_\_*} (\texttt{apply\_formula}, \texttt{copy\_range}/\texttt{copy\_worksheet}, \texttt{create\_chart}/\texttt{create\_pivot\_table}/\texttt{create\_table}/\texttt{create\_workbook}/\texttt{create\_worksheet}, \texttt{delete\_*}, \texttt{format\_range}, \texttt{get\_*}, \texttt{insert\_columns}/\texttt{insert\_rows}, \texttt{merge\_cells}/\texttt{unmerge\_cells}, \texttt{read\_data\_from\_excel}, \texttt{rename\_worksheet}, \texttt{validate\_*}, \texttt{write\_data\_to\_excel}); \texttt{mcp\_\_pptx\_\_*} (\texttt{add\_*}, \texttt{apply\_*}, \texttt{auto\_generate\_presentation}, \texttt{create\_*}, \texttt{extract\_*}, \texttt{format\_table\_cell}, \texttt{get\_*}, \texttt{list\_*}, \texttt{manage\_*}, \texttt{open\_presentation}, \texttt{optimize\_slide\_text}, \texttt{populate\_placeholder}, \texttt{save\_presentation}, \texttt{set\_core\_properties}, \texttt{switch\_presentation}, \texttt{update\_chart\_data}); \texttt{mcp\_\_local\_\_python\_execute}, \texttt{save\_overlong\_output}, \texttt{view\_overlong\_output}; \texttt{mcp\_\_terminal\_\_run\_command}, \texttt{show\_security\_rules}

disallowedTools: \texttt{Agent}, \texttt{AgentSwarm}, \texttt{mcp\_\_local\_\_claim\_done}, \texttt{Bash}, \texttt{Shell}, \texttt{Terminal}, \texttt{Read}, \texttt{ReadMediaFile}, \texttt{Write}, \texttt{Edit}, \texttt{Glob}, \texttt{Grep}, \texttt{WebFetch}, \texttt{WebSearch}, \texttt{FetchURL}, \texttt{Fetch}, \texttt{Browser}

subagents: none

You are an office-report-builder sub-agent inside a Toolathlon-GYM evaluation. You receive a focused sub-task from the main agent: produce workspace artifacts -- spreadsheets, documents, slides, PDFs, scripts, or intermediate data files.

\textbf{Constraints:}

- You may ONLY use the task-approved tools available to you. This includes filesystem read/write, Python execution, spreadsheet/document/slide creation, and any other tools granted for this sub-task. Do not send email or mutate calendars, forms, cloud sheets, or Notion pages.

- Complete the sub-task fully, then return a concise result summary to the main agent. Include exact file paths you created or modified.

- Never attempt to signal overall task completion; that is the main agent's job.

\textbf{Visible Boundary (Strictly Enforced):}

- Read and write files ONLY inside the task workspace directory.

- Interact with external systems ONLY through the task-granted tools.

- Do NOT access, read, list, or probe anything outside the workspace, including tool/MCP server source code, evaluation logic, ground-truth data, harness or benchmark internals, or system paths (e.g. under \texttt{/opt}, \texttt{/workspace/tasks}, \texttt{/workspace/utils}).

- Assume the environment and all granted tools work correctly. Never diagnose, fix, or work around infrastructure; if a granted tool fails, report it and continue. \\
\bottomrule
\end{tabular}
\caption{The principles and runtime tool profile of the \rw{Office-Report-Builder} sub-agent, which produces local workspace artifacts (spreadsheets, documents, slides, PDFs, scripts).}
\label{tab:orb-principles}
\end{table}

%% file: common/references.bib
@misc{kaplan2020scalinglawsneurallanguage,
  title         = {Scaling Laws for Neural Language Models},
  author        = {Jared Kaplan and Sam McCandlish and Tom Henighan and Tom B. Brown and Benjamin Chess and Rewon Child and Scott Gray and Alec Radford and Jeffrey Wu and Dario Amodei},
  year          = {2020},
  eprint        = {2001.08361},
  archiveprefix = {arXiv},
  primaryclass  = {cs.LG},
  url           = {https://arxiv.org/abs/2001.08361}
}

@inproceedings{retu,
  title     = {Rethinking Expert Trajectory Utilization in {LLM} Post-training for Mathematical Reasoning},
  author    = {Ding, Bowen  and
               Chen, Yuhan  and
               Lyu, Jiayang  and
               Yuan, Jiyao  and
               Zhu, Qi  and
               Tian, Shuangshuang  and
               Zhu, Dantong  and
               Wang, Futing  and
               Deng, Heyuan  and
               Mi, Fei  and
               Shang, Lifeng  and
               Lin, Tao},
  editor    = {Liakata, Maria  and
               Moreira, Viviane P.  and
               Zhang, Jiajun  and
               Jurgens, David},
  booktitle = {Proceedings of the 64th Annual Meeting of the {A}ssociation for {C}omputational {L}inguistics (Volume 1: Long Papers)},
  month     = jul,
  year      = {2026},
  address   = {San Diego, California, United States},
  publisher = {Association for Computational Linguistics},
  url       = {https://aclanthology.org/2026.acl-long.1528/},
  doi       = {10.18653/v1/2026.acl-long.1528},
  pages     = {33081--33106},
  isbn      = {979-8-89176-390-6}
}

@misc{glm5,
  title         = {GLM-5: from Vibe Coding to Agentic Engineering},
  author        = {GLM-5-Team and : and Aohan Zeng and Xin Lv and Zhenyu Hou and Zhengxiao Du and Qinkai Zheng and Bin Chen and Da Yin and Chendi Ge and Chenghua Huang and Chengxing Xie and Chenzheng Zhu and Congfeng Yin and Cunxiang Wang and Gengzheng Pan and Hao Zeng and Haoke Zhang and Haoran Wang and Huilong Chen and Jiajie Zhang and Jian Jiao and Jiaqi Guo and Jingsen Wang and Jingzhao Du and Jinzhu Wu and Kedong Wang and Lei Li and Lin Fan and Lucen Zhong and Mingdao Liu and Mingming Zhao and Pengfan Du and Qian Dong and Rui Lu and Shuang-Li and Shulin Cao and Song Liu and Ting Jiang and Xiaodong Chen and Xiaohan Zhang and Xuancheng Huang and Xuezhen Dong and Yabo Xu and Yao Wei and Yifan An and Yilin Niu and Yitong Zhu and Yuanhao Wen and Yukuo Cen and Yushi Bai and Zhongpei Qiao and Zihan Wang and Zikang Wang and Zilin Zhu and Ziqiang Liu and Zixuan Li and Bojie Wang and Bosi Wen and Can Huang and Changpeng Cai and Chao Yu and Chen Li and Chengwei Hu and Chenhui Zhang and Dan Zhang and Daoyan Lin and Dayong Yang and Di Wang and Ding Ai and Erle Zhu and Fangzhou Yi and Feiyu Chen and Guohong Wen and Hailong Sun and Haisha Zhao and Haiyi Hu and Hanchen Zhang and Hanrui Liu and Hanyu Zhang and Hao Peng and Hao Tai and Haobo Zhang and He Liu and Hongwei Wang and Hongxi Yan and Hongyu Ge and Huan Liu and Huanpeng Chu and Jia'ni Zhao and Jiachen Wang and Jiajing Zhao and Jiamin Ren and Jiapeng Wang and Jiaxin Zhang and Jiayi Gui and Jiayue Zhao and Jijie Li and Jing An and Jing Li and Jingwei Yuan and Jinhua Du and Jinxin Liu and Junkai Zhi and Junwen Duan and Kaiyue Zhou and Kangjian Wei and Ke Wang and Keyun Luo and Laiqiang Zhang and Leigang Sha and Liang Xu and Lindong Wu and Lintao Ding and Lu Chen and Minghao Li and Nianyi Lin and Pan Ta and Qiang Zou and Rongjun Song and Ruiqi Yang and Shangqing Tu and Shangtong Yang and Shaoxiang Wu and Shengyan Zhang and Shijie Li and Shuang Li and Shuyi Fan and Wei Qin and Wei Tian and Weining Zhang and Wenbo Yu and Wenjie Liang and Xiang Kuang and Xiangmeng Cheng and Xiangyang Li and Xiaoquan Yan and Xiaowei Hu and Xiaoying Ling and Xing Fan and Xingye Xia and Xinyuan Zhang and Xinze Zhang and Xirui Pan and Xu Zou and Xunkai Zhang and Yadi Liu and Yandong Wu and Yanfu Li and Yidong Wang and Yifan Zhu and Yijun Tan and Yilin Zhou and Yiming Pan and Ying Zhang and Yinpei Su and Yipeng Geng and Yong Yan and Yonglin Tan and Yuean Bi and Yuhan Shen and Yuhao Yang and Yujiang Li and Yunan Liu and Yunqing Wang and Yuntao Li and Yurong Wu and Yutao Zhang and Yuxi Duan and Yuxuan Zhang and Zezhen Liu and Zhengtao Jiang and Zhenhe Yan and Zheyu Zhang and Zhixiang Wei and Zhuo Chen and Zhuoer Feng and Zijun Yao and Ziwei Chai and Ziyuan Wang and Zuzhou Zhang and Bin Xu and Minlie Huang and Hongning Wang and Juanzi Li and Yuxiao Dong and Jie Tang},
  year          = {2026},
  eprint        = {2602.15763},
  archiveprefix = {arXiv},
  primaryclass  = {cs.LG},
  url           = {https://arxiv.org/abs/2602.15763}
}

@article{lost_in_the_middle,
  title     = {Lost in the Middle: How Language Models Use Long Contexts},
  author    = {Liu, Nelson F.  and
               Lin, Kevin  and
               Hewitt, John  and
               Paranjape, Ashwin  and
               Bevilacqua, Michele  and
               Petroni, Fabio  and
               Liang, Percy},
  journal   = {Transactions of the Association for Computational Linguistics},
  volume    = {12},
  year      = {2024},
  address   = {Cambridge, MA},
  publisher = {MIT Press},
  url       = {https://aclanthology.org/2024.tacl-1.9/},
  doi       = {10.1162/tacl_a_00638},
  pages     = {157--173}
}

@inproceedings{mitigating_context_interference,
  title     = {Mitigating Context Interference for Reliable and Efficient Search Agents},
  author    = {Xue, Boyang  and
               Wu, Bin  and
               Qiao, Shuofei  and
               Wang, Sheng  and
               Wang, Rui  and
               Du, Yiming  and
               Wang, Hongru  and
               Pan, Jeff Z.  and
               Yilmaz, Emine  and
               Wong, Kam-Fai  and
               Lipani, Aldo},
  editor    = {Liakata, Maria  and
               Moreira, Viviane P.  and
               Zhang, Jiajun  and
               Jurgens, David},
  booktitle = {Proceedings of the 64th Annual Meeting of the {A}ssociation for {C}omputational {L}inguistics (Volume 1: Long Papers)},
  month     = jul,
  year      = {2026},
  address   = {San Diego, California, United States},
  publisher = {Association for Computational Linguistics},
  url       = {https://aclanthology.org/2026.acl-long.160/},
  doi       = {10.18653/v1/2026.acl-long.160},
  pages     = {3541--3558},
  isbn      = {979-8-89176-390-6}
}

@inproceedings{toolathlon,
  author    = {Junlong Li and Wenshuo Zhao and Jian Zhao and Weihao Zeng and Haoze Wu and Xiaochen Wang and Rui Ge and Yuxuan Cao and Yuzhen Huang and Wei Liu and Junteng Liu and Zhaochen Su and Yiyang Guo and Fan Zhou and Lueyang Zhang and Juan Michelini and Xingyao Wang and Xiang Yue and Shuyan Zhou and Graham Neubig and Junxian He},
  title     = {The Tool Decathlon: Benchmarking Language Agents for Diverse, Realistic, and Long-Horizon Task Execution},
  booktitle = {International Conference on Learning Representations},
  year      = {2026}
}

@inproceedings{camel,
  author    = {Guohao Li and Hasan Hammoud and Hani Itani and Dmitrii Khizbullin and Bernard Ghanem},
  title     = {{CAMEL}: Communicative Agents for {"Mind"} Exploration of Large Language Model Society},
  booktitle = {Advances in Neural Information Processing Systems},
  volume    = {36},
  year      = {2023},
  doi       = {10.52202/075280-2264}
}

@inproceedings{multiagentbench,
  author    = {Kunlun Zhu and Hongyi Du and Zhaochen Hong and Xiaocheng Yang and Shuyi Guo and Zhe Wang and Zhenhailong Wang and Cheng Qian and Xiangru Tang and Heng Ji and Jiaxuan You},
  title     = {{MultiAgentBench}: Evaluating the Collaboration and Competition of {LLM} Agents},
  booktitle = {Proceedings of the 63rd Annual Meeting of the Association for Computational Linguistics (Volume 1: Long Papers)},
  pages     = {8580--8622},
  year      = {2025},
  publisher = {Association for Computational Linguistics},
  doi       = {10.18653/v1/2025.acl-long.421}
}

@article{clawarenateam,
  author  = {Kaiwen Xiong and Haonian Ji and Shi Qiu and Zeyu Zheng and Cihang Xie and Xinyu Ye and Huaxiu Yao},
  title   = {{ClawArena-Team}: Benchmarking Subagent Orchestration and Dynamic Workflows in Language-Model Agents},
  journal = {arXiv preprint arXiv:2606.31174},
  year    = {2026}
}

@misc{anthropic2024mcp,
  author       = {{Anthropic}},
  title        = {Introducing the Model Context Protocol},
  howpublished = {\url{https://www.anthropic.com/news/model-context-protocol}},
  year         = {2024}
}

@misc{claude-code,
  author       = {{Anthropic}},
  title        = {Claude Code},
  year         = {2026},
  howpublished = {Claude Code Documentation},
  url          = {https://code.claude.com/docs/en/overview},
  note         = {Accessed: 2026-09-16}
}

@article{xu2026theagentcompany,
  title   = {Theagentcompany: benchmarking llm agents on consequential real world tasks},
  author  = {Xu, Frank Fangzheng and Song, Yufan and Li, Boxuan and Tang, Yuxuan and Jain, Kritanjali and Bao, Mengxue and Wang, Zora and Zhou, Xuhui and Guo, Zhitong and Cao, Murong and others},
  journal = {Advances in Neural Information Processing Systems},
  volume  = {38},
  year    = {2026}
}

@misc{toolathlon-gym,
  author = {Puzhen Zhang and Weijie Bai and Wendong Fan and Guohao Li},
  title  = {{Toolathlon-GYM: Large-Scale Long-Horizon Environments for Tool-Use Agents}},
  year   = {2026},
  url    = {https://github.com/eigent-ai/toolathlon_gym}
}

@misc{codex,
  author       = {{OpenAI}},
  title        = {Codex},
  year         = {2026},
  howpublished = {GitHub repository},
  url          = {https://github.com/openai/codex},
  note         = {Accessed: 2026-09-16}
}

@misc{cursor,
  author       = {{Cursor}},
  title        = {Cursor},
  year         = {2026},
  howpublished = {Cursor Documentation},
  url          = {https://cursor.com/en-US/docs},
  note         = {Accessed: 2026-09-16}
}

@misc{cursor_subagents,
  author = {{Cursor}},
  title  = {Subagents},
  year   = {2026},
  url    = {https://cursor.com/docs/subagents}
}

@misc{trae,
  author       = {{TRAE Team}},
  title        = {{TRAE}: AI Coding Agent and Development Environment},
  year         = {2026},
  howpublished = {\url{https://www.trae.ai/ide}}
}

@misc{moonshot_kimi_code_agent_tools,
  author = {{Moonshot AI}},
  title  = {Built-in Tools: Agent and {AgentSwarm}},
  year   = {2026},
  url    = {https://moonshotai.github.io/kimi-code/en/reference/tools.html}
}

@misc{openai2026gpt56,
  author       = {{OpenAI}},
  title        = {{GPT-5.6: Frontier Intelligence That Scales with Your Ambition}},
  year         = {2026},
  month        = jul,
  howpublished = {\url{https://openai.com/index/gpt-5-6/}}
}

@misc{anthropic2026sonnet5,
  author = {{Anthropic}},
  title  = {Claude Sonnet 5},
  year   = {2026},
  url    = {https://www.anthropic.com/news/claude-sonnet-5}
}

@misc{qwen38,
  title  = {Qwen3.8-Max: A New Bar for Coding and Cowork},
  url    = {https://qwen.ai/blog?id=qwen3.8},
  author = {{Qwen Team}},
  month  = {August},
  year   = {2026}
}

@misc{k3,
  title         = {Kimi K3: Open Frontier Intelligence},
  author        = {Kimi Team and Tongtong Bai and Yifan Bai and Yiping Bao and M. C. and Jianfeng Cai and Xinyuan Cai and Peizhou Cao and Yuxuan Cao and Ziwei Chai and Y. Charles and H. S. Che and Guanduo Chen and Guangyu Chen and Guanzheng Chen and Huarong Chen and Jia Chen and Jianlong Chen and Jun Chen and Kexin Chen and Peng Chen and Ruijue Chen and Wentao Chen and Xin Chen and Yang Chen and Yanru Chen and Yifei Chen and Yingjiang Chen and Yuankun Chen and Yujie Chen and Yutian Chen and Zhirong Chen and Dazhi Cheng and Yean Cheng and Jialei Cui and Jingbing Cui and Anqi Dai and Jiaqi Deng and Hao Ding and Rui Ding and Shaofeng Ding and Mengfan Dong and Mengnan Dong and Yuhao Dong and Yuxin Dong and Angang Du and Chenzhuang Du and Dikang Du and Jusen Du and Yulun Du and Yu Fan and Jing Feng and Qiulin Feng and Yichen Feng and Kelin Fu and Qiang Fu and Fuxuan Gao and Hongcheng Gao and Jingyue Gao and Tong Gao and Weijia Gao and Shangyi Geng and Jie Gong and Linhu Gong and Shengao Gong and Xiaochen Gong and Qizheng Gu and Yicheng Gu and Shuhao Guan and Haiqing Guo and Shiqi Guo and Xiang Guo and Zhengyan Guo and Beixi Hao and Wenxin Hao and Xiaoru Hao and Dailan He and Haotian He and Lehan He and Qi He and Weiran He and Xinran He and Xinyi He and Yibo He and Yunjia He and Chao Hong and Tiange Hong and Hao Hu and Jiaxi Hu and Ruikun Hu and Weiming Hu and Yangyang Hu and Zhenxing Hu and Liang Hua and Jinbin Huang and Ke Huang and Ruiyuan Huang and Siying Huang and Weixiao Huang and Yan Huang and Zhengjie Huang and Zhiqi Huang and Yulong Hui and Chaobo Jia and Yutong Jiang and Zhejun Jiang and Zuoyou Jiang and Wenyi Jin and Xinyi Jin and Yu Jing and Huanjun Kong and Guokun Lai and Aidi Li and Cheng Li and Chengyuan Li and Cong Li and Fang Li and Guanyu Li and Haoyang Li and Jia Li and Junxiong Li and Lei Li and Letian Li and Lincan Li and Weihong Li and Wentao Li and Xintong Li and Yang Li and Yishen Li and Yiwei Li and Yuxiao Li and Zhaowei Li and Zhaoxi Li and Zheming Li and Zhengxiao Li and Zhiyuan Li and Jiawei Lin and Xiaohan Lin and Yibo Lin and Zichao Lin and Ziyan Lin and Bill Liu and Boxiao Liu and Chuan Liu and Liang Liu and Shaowei Liu and Shudong Liu and Shuran Liu and Tianwei Liu and Weizhou Liu and Yangyang Liu and Yanming Liu and Yibo Liu and Yipeng Liu and Zhengying Liu and Zhiheng Liu and Enzhe Lu and Haoyu Lu and Linqiang Lu and Tingzhan Lu and Zhiyuan Lu and Aotian Luo and G. Luo and Junyu Luo and Yifan Luo and B. Lyu and Wenzhou Lyu and Shaoguang Mao and Yuan Mei and Xin Men and Minqing Ni and Yixuan Niu and Siyuan Pan and Shujun Peng and Zhangyang Qi and Ruoyu Qin and ZeChao Qin and Zeyu Qin and Haiquan Qiu and Jianxin Qiu and Jiezhong Qiu and Bowen Qu and Yuhao Qu and Zeyu Shang and Youbo Shao and Han Shen and Jincheng Shi and Juanfeng Shi and Lidong Shi and Shengyuan Shi and Wingchun Siu and Pengwei Song and Xiaoxi Song and Jianlin Su and Yunfeng Su and Zhaochen Su and Lin Sui and Jingsong Sun and Junyao Sun and Shaoning Sun and Shuzhe Sun and Tongyu Sun and Yujun Sun and Yunpeng Tai and Chuning Tang and Heyi Tang and Sirui Tang and Zecheng Tang and Chaoran Tian and Rongpeng Tian and Yu Tian and Wei Tu and Chensi Wang and Chuang Wang and Chunjie Wang and Dinglu Wang and Feng Wang and Hailong Wang and Haiming Wang and Hao Wang and Hao Wang and Huaqing Wang and Hui Wang and Jiayi Wang and Jinglong Wang and Jinhong Wang and Jiuzheng Wang and Linian Wang and Shaobo Wang and Shenzhi Wang and Shuyi Wang and Si Wang and Siyuan Wang and Tianfu Wang and Wenjue Wang and Xingran Wang and Xinmei Wang and Xinyuan Wang and Xusheng Wang and Yalin Wang and Yangkun Wang and Yao Wang and Yaoyu Wang and Yejie Wang and Yiqin Wang and Yucheng Wang and Yuzhi Wang and Zhaoji Wang and Zhaowei Wang and Zhengtao Wang and Zhenhao Wang and Zhongsheng Wang and Zifan Wang and Chu Wei and Ming Wei and Shouxin Wei and Zichen Wen and Fan Wu and Haoning Wu and Rucong Wu and Wenhao Wu and Xiaoxue Wu and Yingcong Wu and Yongqi Wu and Yuxin Wu and Zijian Wu and Xinglang Xian and Chenxuan Xiang and Yuye Xiang and Bocheng Xiao and Chenjun Xiao and Xin Xiao and Jin Xie and Xiaotong Xie and Yifeng Xie and Zhe Xie and Bowei Xing and Yiming Xiong and Baosheng Xu and Boyu Xu and Jiale Xu and Jianfan Xu and Jing Xu and Jinjing Xu and L. H. Xu and Qingtao Xu and Shuyao Xu and Suting Xu and Tiantian Xu and Tianxiang Xu and Weixin Xu and Xinran Xu and Yangchuan Xu and Ye Xu and Yueni Xu and Ziyao Xu and Haonan Xue and Junjie Yan and Yaoyao Yan and Fan Yang and Guangyao Yang and Hao Yang and Junwei Yang and Ruoyu Yang and Wenjie Yang and Xiaofei Yang and Xinyu Yang and Yi Yang and Yiling Yang and Ying Yang and Yuchen Yang and Zhen Yang and Zhilin Yang and Zian Yang and Zuhao Yang and Haotian Yao and Dan Ye and Haoran Ye and Wenjie Ye and Zhanbo Ye and Bohong Yin and Haoxiang Yin and Xietong Yin and Chengzhen Yu and Haozhen Yu and Longhui Yu and Shengnan Yu and Shuying Yu and Tianxiang Yu and Enming Yuan and Mengjie Yuan and Tongtian Yue and Wei Yue and Yang Yue and Dunyuan Zha and Haobing Zhan and B. H. Zhang and Dehao Zhang and Fei Zhang and Hao Zhang and Haoyuan Zhang and Huanyu Zhang and Jiapei Zhang and Jiaxuan Zhang and Jin Zhang and Kaiyi Zhang and Miaozhen Zhang and Puqi Zhang and Qinglei Zhang and Rong Zhang and Rui Zhang and Shaoshuai Zhang and Shiyi Zhang and Xiaobin Zhang and Xiaoyun Zhang and Y. Zhang and Yangkun Zhang and Ye Zhang and Yichi Zhang and Yikun Zhang and Yizhi Zhang and Yongting Zhang and Yu Zhang and Yutao Zhang and Yutong Zhang and Zheng Zhang and Zijing Zhang and Bin Zhao and Chenguang Zhao and Feifan Zhao and Jinglun Zhao and Jinxiang Zhao and Shuai Zhao and Wenshuo Zhao and Xiangyu Zhao and Xuanle Zhao and Yikai Zhao and Zijia Zhao and Haozhi Zheng and Huabin Zheng and Ruihan Zheng and Shaojie Zheng and Tengyang Zheng and Haofeng Zhong and Lei Zhong and Longguang Zhong and M. Zhou and Qiankang Zhou and Runjie Zhou and Ruozhang Zhou and Xinyu Zhou and Yiqiao Zhou and Zaida Zhou and Jinguo Zhu and Liya Zhu and Xinhao Zhu and Yangjunfeng Zhu and Yuxuan Zhu and Zhen Zhu and Chen Zhuang and Weiyu Zhuang and Xinxing Zu},
  year          = {2026},
  eprint        = {2607.24653},
  archiveprefix = {arXiv},
  primaryclass  = {cs.CL},
  url           = {https://arxiv.org/abs/2607.24653}
}

@misc{zai2026glm53,
  author = {{Z.ai}},
  title  = {{GLM-5.3}: Frontier Coding with Emergent Cyber Capabilities},
  year   = {2026},
  url    = {https://z.ai/blog/glm-5.3}
}

@misc{zai2026glm53-falsh,
  author = {{Z.ai}},
  title  = {{GLM-5.3-Flash}: Frontier Intelligence, Flash Cost},
  year   = {2026},
  url    = {https://z.ai/blog/glm-5.3-flash}
}

@misc{minimax2026m3,
  author       = {{MiniMax}},
  title        = {{MiniMax M3}: Frontier Coding, 1M Context, Native Multimodality --- All in One Model},
  year         = {2026},
  month        = jun,
  howpublished = {\url{https://www.minimax.io/blog/minimax-m3}},
  note         = {Accessed: 2026-09-13}
}

@misc{deepseekai2026deepseekv4,
  title  = {DeepSeek-V4: Towards Highly Efficient Million-Token Context Intelligence},
  author = {DeepSeek-AI},
  year   = {2026}
}

@misc{kimiteam2026kimik25visualagentic,
  title         = {Kimi K2.5: Visual Agentic Intelligence},
  author        = {Kimi-Team and Tongtong Bai and Yifan Bai and Yiping Bao and S. H. Cai and Yuan Cao and Ziwei Chai and Y. Charles and H. S. Che and Cheng Chen and Guanduo Chen and Huarong Chen and Jia Chen and Jianlong Chen and Jun Chen and Kefan Chen and Liang Chen and Ruijue Chen and Xinhao Chen and Yanru Chen and Yanxu Chen and Yicun Chen and Yimin Chen and Yingjiang Chen and Yuankun Chen and Yujie Chen and Yutian Chen and Zhirong Chen and Ziwei Chen and Dazhi Cheng and Yean Cheng and Minghan Chu and Jialei Cui and Jiaqi Deng and Muxi Diao and Hao Ding and Mengfan Dong and Mengnan Dong and Yuxin Dong and Yuhao Dong and Angang Du and Chenzhuang Du and Dikang Du and Lingxiao Du and Yulun Du and Yu Fan and Shengjun Fang and Qiulin Feng and Yichen Feng and Garimugai Fu and Kelin Fu and Hongcheng Gao and Tong Gao and Yuyao Ge and Shangyi Geng and Chengyang Gong and Xiaochen Gong and Zhuoma Gongque and Qizheng Gu and Xinran Gu and Yicheng Gu and Longyu Guan and Shuhao Guan and Yuanying Guo and Xiaoru Hao and Dailan He and Tianhong He and Weiran He and Wenyang He and Yibo He and Yunjia He and Chao Hong and Hao Hu and Jiaxi Hu and Yangyang Hu and Zhenxing Hu and Ke Huang and Ruiyuan Huang and Weixiao Huang and Zhiqi Huang and Chaobo Jia and Tao Jiang and Zhejun Jiang and Xinyi Jin and Yu Jing and Guokun Lai and Aidi Li and C. Li and Cheng Li and Fang Li and Guanghe Li and Guanyu Li and Haitao Li and Haoyang Li and Jia Li and Jingwei Li and Junxiong Li and Lincan Li and Mo Li and Weihong Li and Wentao Li and Xinhang Li and Xinhao Li and Yang Li and Yanhao Li and Yiwei Li and Yuxiao Li and Zhaowei Li and Zhaoxi Li and Zheming Li and Weilong Liao and Jiawei Lin and Xiaohan Lin and Yibo Lin and Zhishan Lin and Zichao Lin and Cheng Liu and Chenyu Liu and Hongzhang Liu and Liang Liu and Shaowei Liu and Shudong Liu and Shuran Liu and Tianwei Liu and Tianyu Liu and Weizhou Liu and Xiangyan Liu and Yangyang Liu and Yanming Liu and Yibo Liu and Yuanxin Liu and Zhengying Liu and Zhongnuo Liu and Enzhe Lu and Haoyu Lu and Zhiyuan Lu and G. Luo and Junyu Luo and Tongxu Luo and Yashuo Luo and Long Ma and Shaoguang Mao and Yuan Mei and Xin Men and Fanqing Meng and Zhiyong Meng and Yibo Miao and Minqing Ni and Kun Ouyang and Siyuan Pan and Bo Pang and Yuchao Qian and Ruoyu Qin and Zeyu Qin and Jiezhong Qiu and Bowen Qu and Zeyu Shang and Youbo Shao and Tianxiao Shen and Zhennan Shen and Juanfeng Shi and Lidong Shi and Shengyuan Shi and Feifan Song and Pengwei Song and Tianhui Song and Xiaoxi Song and Hongjin Su and Jianlin Su and Zhaochen Su and Lin Sui and Jinsong Sun and Junyao Sun and Tongyu Sun and Flood Sung and Yunpeng Tai and Chuning Tang and Heyi Tang and Xiaojuan Tang and Zhengyang Tang and Jiawen Tao and Shiyuan Teng and Chaoran Tian and Pengfei Tian and Bowen Wang and Chensi Wang and Chuang Wang and Congcong Wang and Dingkun Wang and Dinglu Wang and Dongliang Wang and Feng Wang and Hailong Wang and Haiming Wang and Hao Wang and Hengzhi Wang and Huaqing Wang and Hui Wang and Jiahao Wang and Jinhong Wang and Jiuzheng Wang and Kaixin Wang and Linian Wang and Qibin Wang and Shengjie Wang and Shuyi Wang and Si Wang and Wei Wang and Xiaochen Wang and Xinyuan Wang and Yao Wang and Yejie Wang and Yipu Wang and Yiqin Wang and Yucheng Wang and Yuzhi Wang and Zhaoji Wang and Zhaowei Wang and Zhengtao Wang and Zhexu Wang and Zifan Wang and Zihan Wang and Zizhe Wang and Chu Wei and Ming Wei and Chuan Wen and Zichen Wen and Chengjie Wu and Haoning Wu and Junyan Wu and Rucong Wu and Wenhao Wu and Yuefeng Wu and Yuhao Wu and Yuxin Wu and Zijian Wu and Chenjun Xiao and Jin Xie and Xiaotong Xie and Yuchong Xie and Bowei Xing and Boyu Xu and Jianfan Xu and Jing Xu and Jinjing Xu and L. H. Xu and Lin Xu and Suting Xu and Weixin Xu and Xinbo Xu and Xinran Xu and Yangchuan Xu and Yichang Xu and Yuemeng Xu and Zelai Xu and Ziyao Xu and Junjie Yan and Yuzi Yan and Guangyao Yang and Hao Yang and Junwei Yang and Kai Yang and Ningyuan Yang and Xiaofei Yang and Xinlong Yang and Xinyu Yang and Ying Yang and Yi Yang and Yi Yang and Zhen Yang and Zhilin Yang and Zonghan Yang and Haotian Yao and Dan Ye and Haoran Ye and Wenjie Ye and Zhuorui Ye and Peng Yebo and Bohong Yin and Chengzhen Yu and Longhui Yu and Tao Yu and Tianxiang Yu and Enming Yuan and Mengjie Yuan and Xiaokun Yuan and Yang Yue and Weihao Zeng and Dunyuan Zha and Haobing Zhan and Dehao Zhang and Hao Zhang and Jin Zhang and Puqi Zhang and Qiao Zhang and Rui Zhang and Xiaobin Zhang and Xiaoyun Zhang and Y. Zhang and Yadong Zhang and Yangkun Zhang and Yichi Zhang and Yizhi Zhang and Yongting Zhang and Yu Zhang and Yushun Zhang and Yutao Zhang and Yutong Zhang and Zheng Zhang and Chenguang Zhao and Feifan Zhao and Jinxiang Zhao and Shuai Zhao and Xiangyu Zhao and Xuanle Zhao and Yikai Zhao and Zijia Zhao and Huabin Zheng and Ruihan Zheng and Shaojie Zheng and Tengyang Zheng and Junfeng Zhong and Longguang Zhong and Weiming Zhong and M. Zhou and Runjie Zhou and Xinyu Zhou and Zaida Zhou and Jinguo Zhu and Liya Zhu and Xinhao Zhu and Yuxuan Zhu and Zhen Zhu and Jingze Zhuang and Weiyu Zhuang and Ying Zou and Xinxing Zu},
  year          = {2026},
  eprint        = {2602.02276},
  archiveprefix = {arXiv},
  primaryclass  = {cs.CL},
  url           = {https://arxiv.org/abs/2602.02276}
}

@misc{ds_v41_flash,
  author       = {{DeepSeek-AI}},
  title        = {DeepSeek-V4.1-Flash Technical Report},
  year         = {2026},
  howpublished = {Technical Report},
  url          = {https://huggingface.co/deepseek-ai/DeepSeek-V4.1-Flash/blob/main/DeepSeek_V41_Tech_Report.pdf}
}

@misc{mcpworld,
  title         = {MCPWorld: A Unified Benchmarking Testbed for API, GUI, and Hybrid Computer Use Agents},
  author        = {Yunhe Yan and Shihe Wang and Jiajun Du and Yexuan Yang and Yuxuan Shan and Qichen Qiu and Xianqing Jia and Xinge Wang and Xin Yuan and Xu Han and Mao Qin and Yinxiao Chen and Chen Peng and Shangguang Wang and Mengwei Xu},
  year          = {2025},
  eprint        = {2506.07672},
  archiveprefix = {arXiv},
  primaryclass  = {cs.AI},
  url           = {https://arxiv.org/abs/2506.07672}
}

@misc{mcpeval,
  title         = {MCPEval: Automatic MCP-based Deep Evaluation for AI Agent Models},
  author        = {Zhiwei Liu and Jielin Qiu and Shiyu Wang and Jianguo Zhang and Zuxin Liu and Roshan Ram and Haolin Chen and Weiran Yao and Shelby Heinecke and Silvio Savarese and Huan Wang and Caiming Xiong},
  year          = {2025},
  eprint        = {2507.12806},
  archiveprefix = {arXiv},
  primaryclass  = {cs.AI},
  url           = {https://arxiv.org/abs/2507.12806}
}

@misc{livemcpbench,
  title         = {LiveMCPBench: Can Agents Navigate an Ocean of MCP Tools?},
  author        = {Guozhao Mo and Wenliang Zhong and Jiawei Chen and Qianhao Yuan and Xuanang Chen and Yaojie Lu and Hongyu Lin and Ben He and Xianpei Han and Le Sun},
  year          = {2026},
  eprint        = {2508.01780},
  archiveprefix = {arXiv},
  primaryclass  = {cs.AI},
  url           = {https://arxiv.org/abs/2508.01780}
}

@misc{mcp_radar,
  title         = {MCP-RADAR: A Multi-Dimensional Benchmark for Evaluating Tool Use Capabilities in Large Language Models},
  author        = {Xuanqi Gao and Siyi Xie and Juan Zhai and Shiqing Ma and Chao Shen},
  year          = {2025},
  eprint        = {2505.16700},
  archiveprefix = {arXiv},
  primaryclass  = {cs.AI},
  url           = {https://arxiv.org/abs/2505.16700}
}

@misc{mcpuniverse,
  title         = {MCP-Universe: Benchmarking Large Language Models with Real-World Model Context Protocol Servers},
  author        = {Ziyang Luo and Zhiqi Shen and Wenzhuo Yang and Zirui Zhao and Prathyusha Jwalapuram and Amrita Saha and Doyen Sahoo and Silvio Savarese and Caiming Xiong and Junnan Li},
  year          = {2025},
  eprint        = {2508.14704},
  archiveprefix = {arXiv},
  primaryclass  = {cs.AI},
  url           = {https://arxiv.org/abs/2508.14704}
}

@misc{livemcp101,
  title         = {LiveMCP-101: Stress Testing and Diagnosing MCP-enabled Agents on Challenging Queries},
  author        = {Ming Yin and Dinghan Shen and Silei Xu and Sixun Dong and Mian Zhang and Yebowen Hu and Shujian Liu and Jianbing Han and Simin Ma and Song Wang and Sathish Reddy Indurthi and Xun Wang and Yiran Chen and Kaiqiang Song},
  year          = {2026},
  eprint        = {2508.15760},
  archiveprefix = {arXiv},
  primaryclass  = {cs.CL},
  url           = {https://arxiv.org/abs/2508.15760}
}

@misc{mcpagentbench,
  title         = {MCP-AgentBench: Evaluating Real-World Language Agent Performance with MCP-Mediated Tools},
  author        = {Zikang Guo and Benfeng Xu and Chiwei Zhu and Wentao Hong and Xiaorui Wang and Zhendong Mao},
  year          = {2025},
  eprint        = {2509.09734},
  archiveprefix = {arXiv},
  primaryclass  = {cs.CL},
  url           = {https://arxiv.org/abs/2509.09734}
}

@misc{mcp_atlas,
  title         = {MCP-Atlas: A Large-Scale Benchmark for Tool-Use Competency with Real MCP Servers},
  author        = {Chaithanya Bandi and Razvan-Gabriel Dumitru and Ben Hertzberg and Divyansh Agarwal and Geobio Boo and Tejas Polakam and Sami Hassaan and Jeff Da and HiJae Kim and Vipul Gupta and Manasi Sharma and Andrew Park and Martin Dimakis and Ernesto Gabriel Hernandez Montoya and Dan Rambado and Ivan Salazar and Rafael Cruz and MohammadHossein Rezaei and Chetan Rane and Ben Levin and Daniel Yue Zhang and Brad Kenstler and Bing Liu},
  year          = {2026},
  eprint        = {2602.00933},
  archiveprefix = {arXiv},
  primaryclass  = {cs.SE},
  url           = {https://arxiv.org/abs/2602.00933}
}

@misc{mcpmark,
  title         = {MCPMark: A Benchmark for Stress-Testing Realistic and Comprehensive MCP Use},
  author        = {Zijian Wu and Xiangyan Liu and Xinyuan Zhang and Lingjun Chen and Fanqing Meng and Lingxiao Du and Yiran Zhao and Fanshi Zhang and Yaoqi Ye and Jiawei Wang and Zirui Wang and Jinjie Ni and Yufan Yang and Arvin Xu and Michael Qizhe Shieh},
  year          = {2025},
  eprint        = {2509.24002},
  archiveprefix = {arXiv},
  primaryclass  = {cs.CL},
  url           = {https://arxiv.org/abs/2509.24002}
}

@misc{du2023improvingfactualityreasoninglanguage,
  title         = {Improving Factuality and Reasoning in Language Models through Multiagent Debate},
  author        = {Yilun Du and Shuang Li and Antonio Torralba and Joshua B. Tenenbaum and Igor Mordatch},
  year          = {2023},
  eprint        = {2305.14325},
  archiveprefix = {arXiv},
  primaryclass  = {cs.CL},
  url           = {https://arxiv.org/abs/2305.14325}
}

@misc{societyofmind,
  title         = {Generative Agents: Interactive Simulacra of Human Behavior},
  author        = {Joon Sung Park and Joseph C. O'Brien and Carrie J. Cai and Meredith Ringel Morris and Percy Liang and Michael S. Bernstein},
  year          = {2023},
  eprint        = {2304.03442},
  archiveprefix = {arXiv},
  primaryclass  = {cs.HC},
  url           = {https://arxiv.org/abs/2304.03442}
}

@misc{when_do_mas_help,
  title         = {When Do Multi-Agent Systems Help? An Information Bottleneck Perspective},
  author        = {Wendi Yu and Lianhao Zhou and Xiangjue Dong and Sai Sudarshan Barath and Declan Staunton and Byung-Jun Yoon and Xiaoning Qian and James Caverlee and Shuiwang Ji},
  year          = {2026},
  eprint        = {2607.16133},
  archiveprefix = {arXiv},
  primaryclass  = {cs.LG},
  url           = {https://arxiv.org/abs/2607.16133}
}

@inproceedings{llm_based_multi_agents,
  title     = {Large Language Model Based Multi-agents: A Survey of Progress and Challenges},
  author    = {Guo, Taicheng and Chen, Xiuying and Wang, Yaqi and Chang, Ruidi and Pei, Shichao and Chawla, Nitesh V. and Wiest, Olaf and Zhang, Xiangliang},
  booktitle = {Proceedings of the Thirty-Third International Joint Conference on
               Artificial Intelligence, {IJCAI-24}},
  publisher = {International Joint Conferences on Artificial Intelligence Organization},
  editor    = {Kate Larson},
  pages     = {8048--8057},
  year      = {2024},
  month     = {8},
  note      = {Survey Track},
  doi       = {10.24963/ijcai.2024/890},
  url       = {https://doi.org/10.24963/ijcai.2024/890}
}

@misc{masterman2024landscapeemergingaiagent,
  title         = {The Landscape of Emerging AI Agent Architectures for Reasoning, Planning, and Tool Calling: A Survey},
  author        = {Tula Masterman and Sandi Besen and Mason Sawtell and Alex Chao},
  year          = {2024},
  eprint        = {2404.11584},
  archiveprefix = {arXiv},
  primaryclass  = {cs.AI},
  url           = {https://arxiv.org/abs/2404.11584}
}

@misc{dora_explorer,
  title         = {DORA Explorer: Improving the Exploration Ability of LLMs Without Training},
  author        = {Priya Gurjar and Md Farhan Ishmam and Kenneth Marino},
  year          = {2026},
  eprint        = {2604.17244},
  archiveprefix = {arXiv},
  primaryclass  = {cs.CL},
  url           = {https://arxiv.org/abs/2604.17244}
}

@misc{fu2026agentshelpcontrolledprotocolaligned,
  title         = {Do More Agents Help? Controlled and Protocol-Aligned Evaluation of LLM Agent Workflows},
  author        = {Yuhang Fu and Ruishan Fang and Jiaqi Shao and Huiyu Zheng and Zhengtao Zhu and Bing Luo and Tao Lin},
  year          = {2026},
  eprint        = {2606.05670},
  archiveprefix = {arXiv},
  primaryclass  = {cs.AI},
  url           = {https://arxiv.org/abs/2606.05670}
}

@misc{deepseek_api_pricing,
  author       = {{DeepSeek-AI}},
  title        = {DeepSeek API Pricing},
  howpublished = {DeepSeek API Documentation},
  url          = {https://platform.deepseek.com/usage},
  year          = {2026},
}

@misc{qwen38max_api_pricing,
  author       = {{Alibaba Cloud}},
  title        = {Qwen3.8-Max Pricing},
  howpublished = {Qianwen AI Model Documentation},
  url          = {https://www.qianwenai.com/models/qwen3.8-max},
  year          = {2026},
}

@misc{zai_api_pricing,
  author       = {{Z.AI}},
  title        = {Z.AI API Pricing},
  howpublished = {Z.AI Developer Documentation},
  url          = {https://docs.z.ai/guides/overview/pricing},
  year          = {2026},
}

@misc{kimi_k3_api_pricing,
  author       = {{Moonshot AI}},
  title        = {Kimi K3 API Pricing},
  howpublished = {Kimi API Documentation},
  url          = {https://platform.kimi.com/docs/pricing/chat-k3.md},
  year          = {2026},
}

@misc{minimax_m3_api_pricing,
  author       = {{MiniMax}},
  title        = {MiniMax API Pay-as-you-go Pricing},
  howpublished = {MiniMax API Documentation},
  url          = {https://platform.minimax.com/docs/guides/pricing-paygo},
  year          = {2026},
}

@misc{claude_api_pricing,
  author       = {{Anthropic}},
  title        = {Claude API Pricing},
  howpublished = {Anthropic API Documentation},
  url          = {https://platform.claude.com/docs/en/about-claude/pricing},
  year          = {2026},
}

@misc{gpt56sol_api_pricing,
  author       = {{OpenAI}},
  title        = {GPT-5.6 Sol API Pricing},
  howpublished = {OpenAI API Documentation},
  url          = {https://developers.openai.com/api/docs/models/gpt-5.6-sol},
  year          = {2026},
}
